\documentclass[]{xiaomiev}

\usepackage[toc,page,header]{appendix}
\usepackage{minitoc}
\usepackage{solarized-light}
\usepackage{booktabs}
\usepackage{multirow}
\usepackage{graphicx}
\usepackage{array}
\usepackage{siunitx,array}
\usepackage[table]{xcolor}
\usepackage{makecell,array,booktabs}
\usepackage{makecell}
\usepackage{framed}

\usepackage{bbding}
\usepackage{hyperref}
\usepackage{amsmath} 
\usepackage{placeins}
\usepackage[colorinlistoftodos]{todonotes}
\usepackage{longtable}
\usepackage{hhline}
\usepackage{fancyvrb}
\usepackage{graphicx}
\usepackage[table]{xcolor}
\usepackage{booktabs}
\usepackage{float}
\usepackage{fvextra}
\usepackage{CJKutf8}
\usepackage{multicol}
\usepackage{float}
\usepackage{placeins}
\usepackage{cleveref}
\usepackage{tablefootnote}
\usepackage{threeparttable}
\usepackage{tabularx}
\usepackage{mdframed}
\usepackage{subcaption}
\usepackage[usestackEOL]{stackengine}
\usepackage[numbers]{natbib}
\newcommand{\commentout}[1]{}
\renewcommand{\paragraph}[1]{\noindent\textbf{#1.}\hspace*{1em}}
\usepackage{enumitem}
\usepackage{hyperref}
\setlist[itemize]{leftmargin=15pt}
\usepackage{siunitx,array}
\RequirePackage{xspace}
\makeatletter
\DeclareRobustCommand\onedot{\futurelet\@let@token\@onedot}
\def\@onedot{\ifx\@let@token.\else.\null\fi\xspace}

\makeatother
\newcommand{\XFM}{MiMo-Embodied}

\title{MiMo-Embodied: X-Embodied Foundation Model \\ Technical Report}

\author{Xiaomi Embodied Intelligence Team}
\vspace{-11pt}

\contribution{See \hyperref[sec:contributions]{Contributions and Acknowledgments} section for a full author list.}

\abstract{

We open-source \XFM{}, the first cross-embodied foundation model to successfully integrate and achieve state-of-the-art performance in both Autonomous Driving and Embodied AI. \XFM{} sets new records across \textbf{17} embodied AI benchmarks in \textbf{Task Planning}, \textbf{Affordance Prediction} and \textbf{Spatial Understanding}, while also excelling in \textbf{12} autonomous driving benchmarks across \textbf{Environmental Perception}, \textbf{Status Prediction}, and \textbf{Driving Planning}. Across these tasks, \XFM{} significantly outperforms existing open-source, closed-source, and specialized baselines. Our results indicate that through multi-stage learning, curated data construction, and CoT/RL fine-tuning, these two domains exhibit strong positive transfer and \textbf{mutually reinforce} one another. We provide a detailed analysis of our model design and training methodologies to facilitate further research. Code and models are available at \url{https://github.com/XiaomiMiMo/MiMo-Embodied}. }

\begin{document}
\maketitle
\vspace{-4pt}

\begin{figure}[!h]
    \centering
    \vspace{-10pt}
\includegraphics[width=\linewidth]{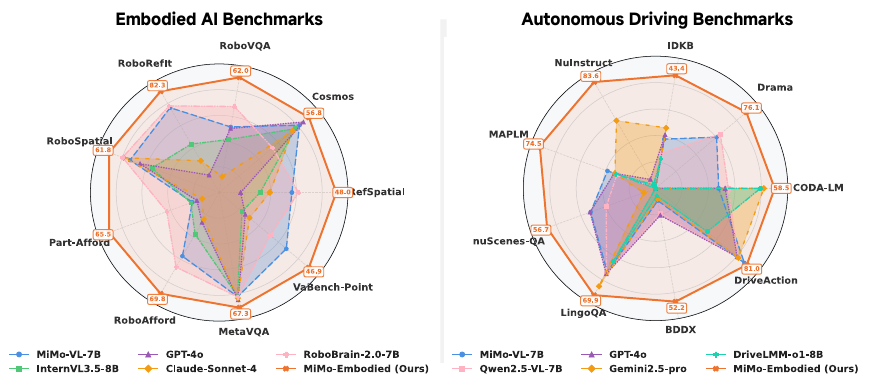}
    \caption{\textbf{Performance Comparison in Autonomous Driving and Embodied AI Benchmarks}. \XFM{} achieves state-of-the-art performance on both benchmarks, surpassing previous open-source, closed-source, and specialized VLMs, highlighting its superior capabilities in various autonomous driving and embodied AI tasks.}
    \label{fig:model}
\end{figure}

\newpage
\tableofcontents
\newpage

\section{Introduction}
Vision-Language Models (VLMs)~\cite{gpt4o, team2025gemini, bai2025qwen2, wang2025internvl3, seed15, chen2025nanovla} demonstrate significant promise in advancing multimodal perception, understanding, and reasoning capabilities. 
Recently, some specialized embodied VLMs have emerged in domains such as autonomous driving and embodied AI, highlighting their interactions with the physical world~\cite{ji2025robobrain,baairobobrainteam2025robobrain20technicalreport,luo2025vebrain,hao2025roboafford++,team2025gemini}.
Specifically, specialized VLMs for embodied AI, such as RoboBrain~\cite{ji2025robobrain,baairobobrainteam2025robobrain20technicalreport} and VeBrain~\cite{luo2025vebrain}, emphasize individual capabilities like task planning and spatial understanding, delivering critical information for robots during physical interactions. Similarly, specialized VLMs in autonomous driving, such as RoboTron-Drive~\cite{huang2025robotron} and DriveLMM-o1~\cite{ishaq2025drivelmm}, focus on enhancing specific aspects of driving performance, including environmental perception, status prediction,  and driving planning, providing necessary information support for autonomous driving systems.
While these specialized embodied VLMs have made significant progress, they are constrained by their narrowly defined application scenarios. 
The focus of embodied AI on indoor tasks and autonomous driving on outdoor roads creates a significant domain gap, which hinders the cross-domain generalization of capabilities.
We summarize the challenges faced by specialized embodied VLMs as follows: \textbf{(1) Lack of Unified Embodied VLMs}. Existing VLMs focus on a single domain and lack a unified VLM that bridges autonomous driving and embodied AI. This fragmentation hinders the generalization of spatial understanding and reasoning across diverse indoor and outdoor scenarios, ultimately limiting the model's capability to interact effectively with the physical world in dynamic environments.
\textbf{(2) Absence of Comprehensive Cross-Embodiment Capability Evaluation}. Existing VLMs only assess partial capabilities in either autonomous driving or embodied AI, lacking a comprehensive cross-embodiment capability evaluation to assess the overall performance of specialized embodied VLMs.

\begin{figure}[!t]
    \centering
    \includegraphics[width=1.0\linewidth]{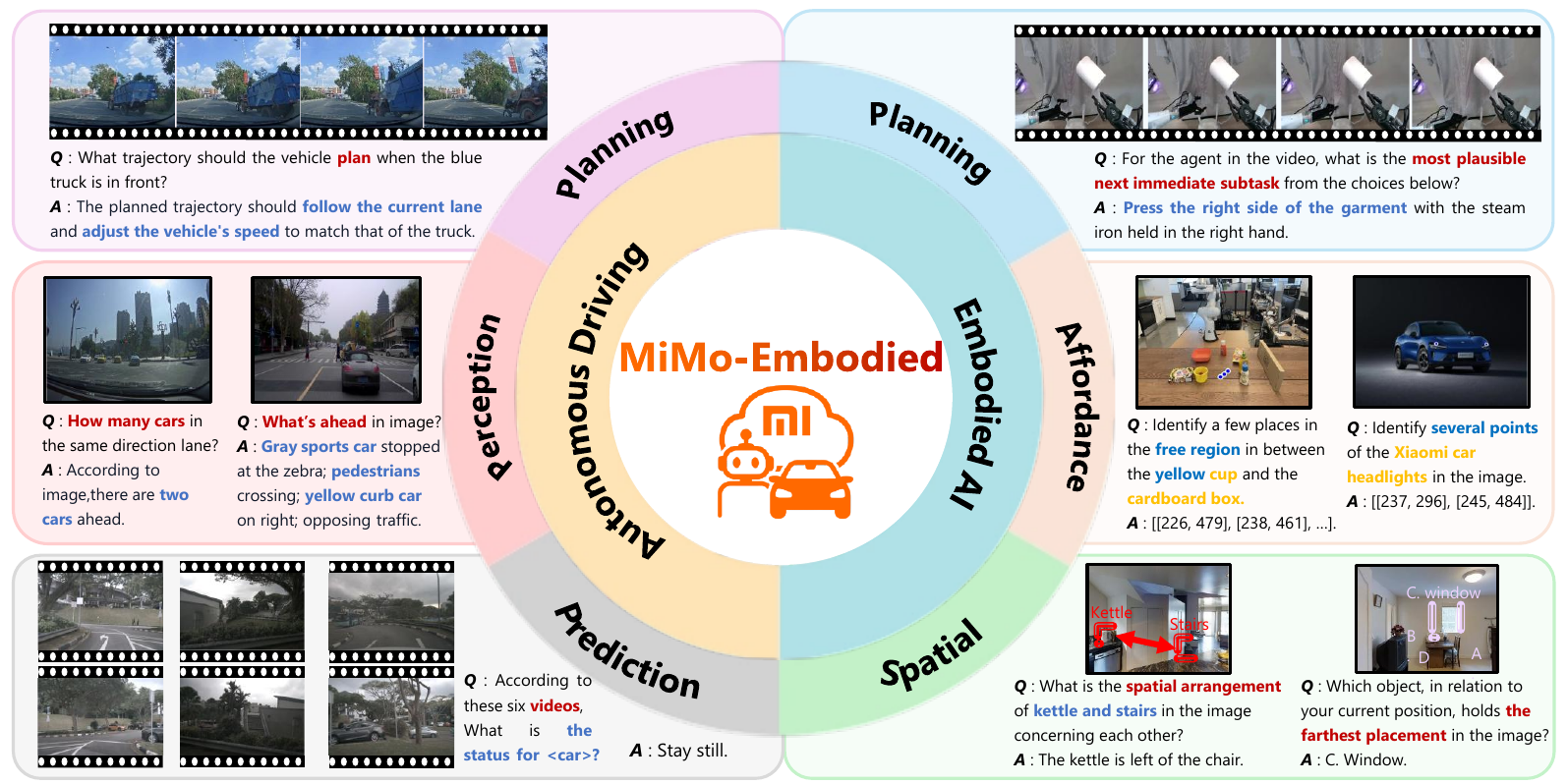}
    \caption{\textbf{Overview of \XFM{} Capabilities.}
  \XFM{} supports both Autonomous Driving and Embodied AI tasks, featuring 12 benchmarks in Autonomous Driving that cover Environmental Perception, Status Prediction and Driving Planning, along with 17 benchmarks in Embodied AI tasks focusing on Affordance Prediction, Task Planning, and Spatial Understanding.
    }
    \label{fig2}
    \vspace{-1em}
\end{figure}

To address these challenges, we present \XFM{}, a unified VLM that merges the tasks of autonomous driving and embodied AI into a single model. 
To our knowledge, \XFM{} is the first open-source VLM to integrate these critical areas, significantly enhancing understanding and reasoning in dynamic physical environments.
To evaluate \XFM{}, we conduct a series of comprehensive cross-embodiment capability benchmarks to assess the overall performance of specialized embodied VLMs. As shown in \Cref{fig2}, our \XFM{} model focuses on key capabilities in both autonomous driving and embodied AI. 
For autonomous driving, we evaluate three core capabilities: (1) \textbf{Environmental Perception}: understanding traffic scenes through semantic and geometric analysis to scene comprehension, region-level interpretation and potential hazards detection; (2) \textbf{Status Prediction}: forecasting the behaviors of agents and multi-agent interactions to facilitate proactive decision-making; and (3) \textbf{Driving Planning}: generating safe driving maneuvers with explainable justifications based on traffic logic, ensuring compliance with road rules while optimizing safety and efficiency.
For embodied AI, we evaluate three core capabilities: (1) \textbf{Affordance Prediction}: inferring actionable interaction possibilities from visual scenes to enable effective interactions; (2) \textbf{Task Planning}: translating abstract instructions into executable action sequences for accomplishing specific goals; and (3) \textbf{Spatial Understanding}: reasoning about spatial relationships, including directions, distances, and layouts, to facilitate navigation and interaction within the environment.
Finally, \XFM{} not only advances the state of specialized embodied VLMs but also sets a new standard for integrating diverse competencies, paving the way for more intelligent and adaptable systems in complex real-world applications.

To equip \XFM{} with the capability to handle both embodied AI and autonomous driving tasks, we constructed a diverse dataset of spanning general visual-language understanding, embodied tasks, and autonomous driving scenarios.
This dataset provides the multimodal foundation for \XFM{}'s core capabilities in perception, prediction, and planning for both autonomous driving and embodied AI.
Based on the constructed dataset, we develop a progressive four-stage training strategy, which is critical to achieving state-of-the-art performance. The training stages comprises:
\textit{}{Stage 1: General and Embodied Knowledge Learning} establish core affordance understanding, high-level task planning, and spatial reasoning abilities based on general visual knowledge.
\textit{Stage 2: Autonomous Driving Knowledge Learning} integrates cross-domain understanding through mixed supervision from both autonomous driving and embodied data.
\textit{Stage 3: Chain-of-Thought Fine-tuning} enhances complex reasoning by incorporating generated rationales.
\textit{Stage 4: Reinforcement Learning Fine-tuning} further refines task-specific performance via GRPO~\cite{guo2025deepseek} optimization.
Extensive evaluations demonstrate \XFM{}'s strength on 17 benchmarks in embodied AI, and leading performance across 12 benchmarks in autonomous driving, which outperforms current open-source and closed-source general VLMs, as well as specialized models.

This report offers a comprehensive overview of the development of \XFM{}, including its model architecture, pretraining dataset, training strategy, benchmark evaluation results, and application cases.
The remaining sections of this report are organized as follows: \textbf{\Cref{sec:arch}} details the model architecture, establishing the foundation for our unified framework.
\textbf{\Cref{sec:pretrain}} outlines the data curation and construction process across three categories: general multimodal understanding, embodied AI, and autonomous driving.
\textbf{\Cref{sec:strategy}} elaborates on the four-stage training strategy, which is crucial to achieving our state-of-the-art performance.
\textbf{\Cref{sec:evaluation}} presents comprehensive quantitative and qualitative results, validating the superior performance of our model across various domains.

\section{Architecture}
\label{sec:arch}

\begin{figure*}[t]
    \centering
    \includegraphics[width=1.0\linewidth]{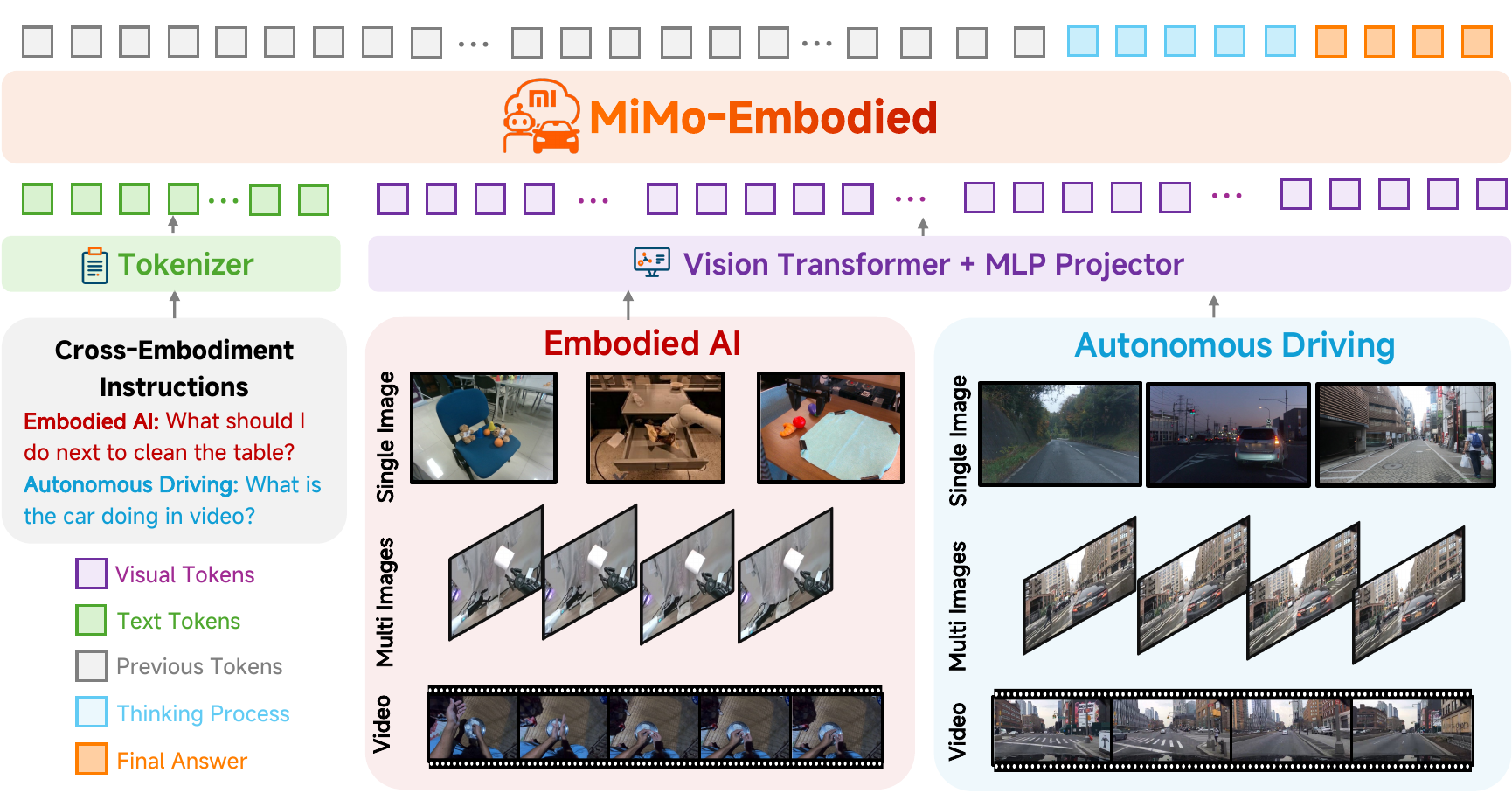}
    \caption{\textbf{Model architecture of \XFM{}.}
    The \XFM{} model architecture is designed for embodied AI and autonomous driving tasks, effectively processing single images, multiple images, and videos, and consists of three main components: (1) a Vision Transformer for encoding visual inputs; (2) a projector that maps visual encodings to a latent space aligned with a LLM; and (3) the LLM itself for textual understanding and reasoning.
    }
    \label{fig3}
\end{figure*}

\subsection{Core Components of \XFM{}}
\label{sec2.1}

The \XFM{} architecture consists of three main components: (1) a Vision Transformer (ViT) for encoding visual inputs; (2) a projector that transforms these visual encodings into a latent space aligned with the Large Language Model (LLM); and (3) the LLM, which is responsible for text understanding and reasoning. To effectively process high-resolution inputs, we adopt the vision encoder from MiMo-VL~\cite{coreteam2025mimovltechnicalreport}. Similarly, the LLM backbone and the projector are initialized using their respective pre-trained weights from MiMo-VL~\cite{coreteam2025mimovltechnicalreport}, inheriting its established vision-language alignment and powerful reasoning capabilities. The projector is implemented using a multi-layer perceptron (MLP), which maps visual tokens to the LLM's input space. The overall architecture of \XFM{} is shown in Figure~\ref{fig3}.

\subsection{Visual Input Processing}
\label{sec2.2}

During the visual input processing stage, the ViT is essential for encoding various types of visual inputs, including single images, multiple images, and videos. This component employs a self-attention mechanism to extract significant features from the input data, allowing the model to discern complex patterns and relationships. The visual tokens generated through this process are formatted for seamless integration with the subsequent stages of the architecture. This encoding is vital for accurately representing the visual context, facilitating its interpretation by the LLM and enabling richer interactions and more coherent reasoning.

\subsection{Latent Space Projection and Output Generation}
\label{sec2.3}
After encoding the visual input, the projector maps these visual tokens into a latent space compatible with the LLM. The projection stage utilizes a MLP to transform the high-dimensional visual representation, making it compatible with the LLM while preserving its essential characteristics. Once in the latent space, the LLM initiates a ``thinking'' process, interpreting the projected data and generating coherent and contextually relevant responses. 
By seamlessly integrating the visual and textual domains, \XFM{} enhances the potential for diverse multimodal reasoning tasks and applications.

\section{Training Dataset}
\label{sec:pretrain}
\begin{figure*}[t]
    \centering
    \includegraphics[width=1.0\linewidth]{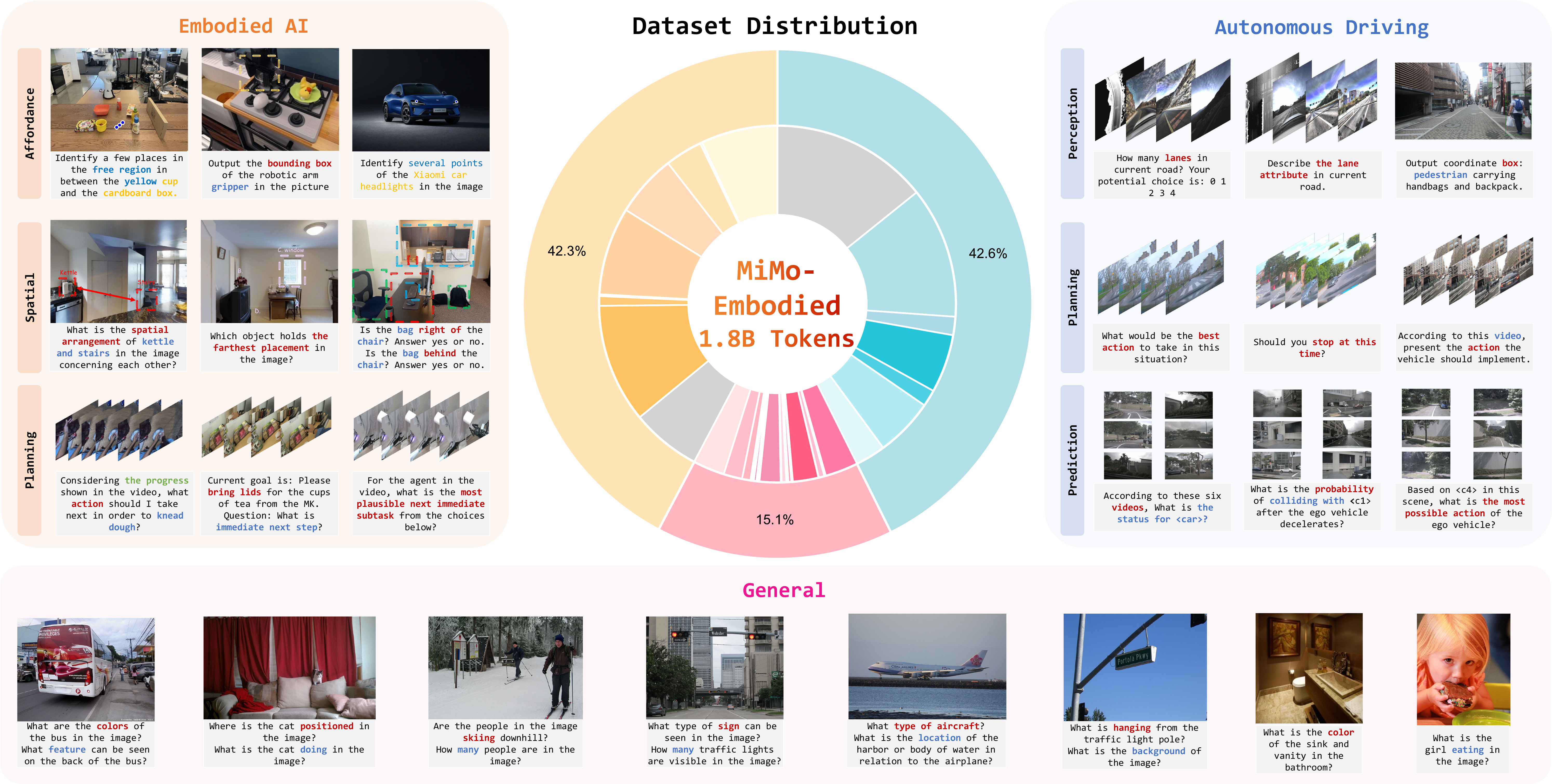}
    \caption{\textbf{Overview of the Training Data used by \XFM{}.}
   Our model comprises three core components of training datasets: the General Dataset establishes foundational capabilities, the Embodied AI Dataset enhances capabilities in affordance, planning, and spatial perception, and the Autonomous Driving Dataset focuses on improving capabilities in perception, prediction, and planning for autonomous driving.}
    \label{fig4}
\end{figure*}

In this section, we outline the composition and curation of the multimodal training dataset utilized for developing \XFM{}, specifically targeting foundational capabilities for embodied AI and autonomous driving. As shown in Figure~\ref{fig4}, our training data encompasses general dataset, along with specialized datasets for embodied AI and autonomous driving scenarios. Each subset is meticulously sourced and processed to ensure diversity, scalability, and alignment with the demands of real-world applications.

\subsection{General Dataset}
To build a versatile and powerful foundation for both general and specialized multimodal tasks, we leverage the training corpus of MiMo-VL~\cite{coreteam2025mimovltechnicalreport}. This dataset integrates diverse data types including high-resolution images, videos, long text, long documents, and synthetic reasoning data, ensuring broad coverage of perceptual, reasoning, and interactive capabilities. These data can be categorized into the following categories: (1) \textit{Visual Grounding} data cultivates fine-grained object-level localization and cross-modal alignment across diverse scenes, enabling precise region-reference understanding and object attribute recognition. (2) \textit{Document and Chart Comprehension} resources develop structural understanding of complex textual layouts, tables, diagrams, and information graphics, facilitating robust OCR capability and logical information extraction from formatted content. (3) \textit{Video Understanding} materials enable temporal reasoning through dense event captioning and dynamic scene analysis, supporting both short-term action recognition and long-range temporal dependency modeling. (4) \textit{Multimodal Reasoning} components build complex logical inference capabilities spanning mathematical, scientific, and symbolic domains, incorporating both perceptual reasoning and long-chain logical deduction tasks. This strategically balanced composition ensures synergistic development of both general vision-language understanding and specialized task proficiency, providing a solid foundation for diverse real-world multimodal scenarios.

\subsection{Embodied AI Dataset}
The embodied AI dataset are categorized into three parts based on the target capabilities as follows: affordance prediction, high-level task planning, and spatial understanding.

\subsubsection{Affordance Prediction}
To establish a comprehensive understanding of affordance, we collect data from the following sources: PixMo-Points~\cite{deitke2024molmo}, RoboAfford~\cite{tang2025roboafford}, and RoboRefIt~\cite{lu2023vl}.

\textbf{PixMo-Points} PixMo-Points~\cite{deitke2024molmo} provides a large-scale collection of diverse web images specifically designed to enhance the model’s ability to perform fine-grained localization through natural language queries. It supports two core capabilities: (1) \textit{Object Counting by Pointing}, where the model enumerates instances by sequentially pointing to each occurrence, and (2) \textit{Visual Explanation via Pointing}, where the model grounds its answers by referring to relevant regions in the image. By training on PixMo-Points, the model learns to associate linguistic expressions with precise spatial locations, improving both referential clarity and interactive capabilities.

\textbf{RoboAfford} RoboAfford~\cite{tang2025roboafford} is a unified affordance-centric dataset systematically structured to address both object-level and scene-level interaction reasoning. It covers the following five task categories: (1) \textit{Visual Captioning}: generating contextual descriptions of scenes and objects; (2) \textit{Visual Grounding}: establishing correspondences between language descriptions and object locations using bounding boxes; (3) \textit{Pointing}: identifying objects based on category and attribute descriptions; (4) \textit{Object Affordance Grounding and Pointing}: localizing functional parts of objects such as handles or graspable areas using both bounding boxes and points; and (5) \textit{Spatial Affordance Localization}: identifying suitable placement regions in free space or on supporting surfaces to facilitate object manipulation. This multi-task setup enables the model to reason about not only \textit{what} actions an object affords, but also \textit{where} those actions can be executed.

\textbf{RoboRefIt} Beyond object pointing and functional grounding, referential understanding in spatially complex settings is essential for embodied interaction. To this end, we incorporate RoboRefIt~\cite{lu2023vl}, a dataset featuring referring expressions collected from 187 cluttered indoor scenes. RoboRefIt emphasizes challenging visual grounding scenarios with multiple instances of the same object category, requiring the model to resolve referential ambiguity through attributes and spatial relations. Its annotations include both 2D bounding boxes and segmentation masks, supporting tasks such as referring expression comprehension (REC) and segmentation (RES). By integrating RoboRefIt, the model strengthens its ability to interpret language directives in context-rich, multi-object environments.

\subsubsection{High-level Task Planning}
To advance the capabilities of multimodal models in embodied reasoning and long-horizon task planning, we construct a comprehensive task planning dataset by processing and integrating three key sources: Cosmos-Reason1~\cite{azzolini2025cosmos}, EgoPlan-IT~\cite{chen2023egoplan}, and RoboVQA~\cite{sermanet2024robovqa}.

\textbf{Cosmos-Reason1} The Cosmos-Reason1~\cite{azzolini2025cosmos} component forms the foundation of planning dataset, providing a large-scale collection focused on grounding language in physical reality. We specifically leverage its data derived from BridgeData V2~\cite{walke2023bridgedata}, RoboVQA~\cite{sermanet2024robovqa}, AgiBot~\cite{bu2025agibot}, and HoloAssist~\cite{wang2023holoassist}, which together cover a wide spectrum of embodiments, including robot arms, humanoid robots, and human activities. The data is structured into two complementary annotation types: (1) Understanding Annotations: These provide dense, structured captions for video clips, detailing objects, their attributes, states, and the actions being performed. This serves as the foundational visual context.
(2) Reasoning Annotations: This subset includes complex questions accompanied by long-chain reasoning traces generated by DeepSeek-R1~\cite{guo2025deepseek}, which require multi-step physical and embodied reasoning beyond simple caption understanding.
We filter out short videos and use longer videos for training. This dataset equips our model with a robust understanding of intuitive physics and cross-embodiment reasoning, enabling the model to perform task completion verification, action affordance judgment, and next-action prediction for a variety of agents.

\textbf{EgoPlan-IT} To equip the model with human-like planning from a first-person perspective, we incorporate a high-quality subset of EgoPlan-IT~\cite{chen2023egoplan}. This dataset is built upon large-scale egocentric video datasets (Epic-Kitchens~\cite{damen2018scaling}, Ego4D~\cite{grauman2022ego4d}), mirroring how humans perceive and interact with the world. Each sample is structured as a multiple-choice question that encapsulates a real-world planning problem. The input includes a current observation image with a task goal described in open-form language, and a video clip showing the historical progress of the task. The model is required to predict the correct next action plan from a set of candidate options, which include the ground-truth action and several carefully selected negative answers derived from the same task goal.

\textbf{RoboVQA} To enhance long-horizon reasoning, we integrate data from RoboVQA~\cite{sermanet2024robovqa} including ten types of QA pairs that are automatically generated from long-horizon activity sequences. This component is vital for training models on a broad range of visual question-answering tasks that are fundamental to robotics.

\subsubsection{Spatial Understanding}
To equip the model with spatial reasoning capabilities, we integrated a comprehensive collection of datasets specifically focused on 3D visual grounding, spatial reasoning, and embodied scene understanding. These datasets provide rich supervision for understanding object relationships in 3D space, interpreting scene layouts, and responding to complex spatial queries, thereby forming a foundational pillar for tasks requiring spatial awareness. The data are collected from the following datasets:

\textbf{SQA3D and Self-curated Data} We utilize SQA3D~\cite{ma2022sqa3d} and a curated 3D dataset for 3D Question Answering (3D-QA), a task requiring holistic 3D scene understanding from textual questions. To advance this paradigm, we introduce 3D bounding box localization. Our data combines questions with calibrated images and camera parameters, supporting diverse challenges from object attributes to inter-object spatial reasoning. Furthermore, we generate large-scale 3D grounding data from existing datasets, where each sample includes an RGB image with a spatial query, requiring the model to regress 3D boxes in a camera coordinate system. This diverse collection, spanning various scenes and object scales, provides crucial supervision for monocular 3D understanding and domain generalization.

\textbf{VLM-3R} We utilize spatial reasoning and navigation QA pairs from the VLM-3R~\cite{fan2025vlm} dataset, which extends the VSI-Bench~\cite{yang2025thinking} framework with large-scale automated generation of spatio-temporal question–answer pairs. The data encompass seven core spatial reasoning tasks and embodied route planning, each sample combining egocentric video inputs with questions about 3D object relations and navigational actions.

\textbf{RefSpatial} To enhance spatial reasoning capabilities for robotic referring tasks, we select samples from RefSpatial~\cite{zhou2025roborefer} dataset. This collection focuses on spatial understanding tasks including object localization, spatial relation comprehension, and free-space identification for placement. The dataset features multi-modal RGB-D inputs paired with textual instructions requiring precise 2D coordinate predictions. Each sample contains multi-turn conversations that progressively address spatial constraints, supporting both single-step perception and multi-step reasoning. The data format integrates visual grounding with structured responses, enabling the model to learn fine-grained spatial relationships essential for real-world robotic interaction.

\textbf{EmbSpatial} To enhance spatial reasoning capabilities in embodied scenarios, we employ data from EmbSpatial-SFT~\cite{du2024embspatial}, which comprise spatial relationship identification and object localization tasks. These samples enables our model to develop robust spatial understanding from an egocentric perspective, which is essential for embodied task execution.

\subsection{Autonomous Driving Dataset}
The autonomous driving dataset is categorized into three parts based on the core functional modules of the system as follows: environmental perception, status prediction, and driving planning.

\subsubsection{Environmental Perception}
Environmental perception serves as the fundamental cornerstone for autonomous driving systems to comprehend surrounding environments, laying the groundwork for subsequent processes. 
To fulfill this objective, the relevant data not only encompass both panoramic overviews and local specifics but also demonstrate multi-level capabilities, including information comprehension and target localization.

\textbf{General Scene Understanding} As a core component, its primary function is to holistically describe the overall context of the driving scenario and identify key environmental and traffic elements within it.
This task aims to build macro-level environmental cognition, distinguishing scenario types and identifying large-scale traffic components.
Specifically, CODA-LM~\cite{chen2025automated} includes a general perception subtask that focuses on understanding key road entities in driving scenarios, covering the appearance and location of these entities, as well as the reasons why they influence the driving behavior of the ego vehicle.
A scenery dataset proposed in prior research~\cite{marcu2024lingoqa} incorporates annotations across roughly 15 different categories.
These categories cover driver actions, justifications, attention, observations of vehicles, observations of pedestrians, road elements, and environmental details.
Building on nuScenes, DriveLM~\cite{sima2024drivelm} generates identify questions related to objects and traffic elements.
Similarly, related research~\cite{wang2024omnidrive} generates QA pairs that cover a comprehensive range of perception and understanding tasks for driving scenarios, encompassing core dimensions to fully address the understanding of scenario descriptions and key attention elements: it includes scene description, identifying close objects, identifying threatening traffic elements, object countings, recognition of object color and relative position, and OCR-type tasks. 
nuScenes-QA~\cite{qian2024nuscenes} generates scene graphs using existing 3D detection annotations and manually designs question templates to support the generation of general perception data.
MME-RealWorld~\cite{zhang2024mme} emphasizes key tasks including object identification, object counting, and driver attention understanding, while IDKB~\cite{lu2025can} specializes in encoding knowledge related to traffic laws and regulations, providing a regulatory foundation for interpreting scenario elements and their implications for driving behavior. 
MAPLM~\cite{cao2024maplm} focuses on road-centric analysis tasks, covering road scene understanding, point cloud quality analysis, road intersection recognition, lane counting, and road and lane description.
Collectively, these datasets contribute to a multi-faceted understanding of driving scenarios.

\textbf{Regional Object Understanding}
As a object-level perception task, its core objective is to enable fine-grained QA interactions for specific target objects within driving scenarios, focusing on detailed recognition, attribute analysis, and behavioral impact interpretation of localized objects. 
This involves not only detailed recognition of these objects but also in-depth attribute analysis and interpretation of their behavioral impacts. 
Specifically, CODA-LM~\cite{chen2025automated} is designed to understand corner-case objects in scenarios where specific bounding boxes are provided. 
Beyond simply describing the objects within these predefined bounding boxes, it delves into elaborating on the specific mechanisms through which these often rare or high-risk objects influence the driving behavior of the ego vehicle. DriveLM~\cite{sima2024drivelm} concentrates on two key aspects of localized traffic elements: it supports both traffic elements visual description, which captures static attributes such as shape, color, and texture, and traffic elements motion State, which analyzes dynamic characteristics like speed, direction, and acceleration.
DriveAction~\cite{hao2025driveaction} enriches regional object understanding with 14 vision and language tasks. 
These tasks cover a wide range of aspects, including road structure details, traffic sign attributes, and both dynamic and static traffic elements. 
MME-RealWorld~\cite{zhang2024mme} focus on critical dynamic and attribute tasks. These include pedestrians and vehicles motion attribute identification and traffic signal attribute identification.
nuScenes-QA~\cite{qian2024nuscenes}enhances object-centric QA by supporting three core tasks: object recognition, which confirms object categories; object status, which assesses states such as stopped; and objects comparison, for example, judging which vehicle is closer to the ego car. 
Complementing these, IDKB~\cite{lu2025can} provides knowledge support specific to the autonomous driving domain by encoding detailed information about road signs, traffic signals, and lane markings. 
This knowledge base serves as a foundation for interpreting the semantic meaning of regional objects and how these objects regulate driving decisions. 
Collectively, these datasets form a comprehensive framework for Regional Object Understanding, enabling precise, targeted analysis of localized objects in driving scenarios that also takes context into account.

\textbf{Regional Object Localization}
A object-level positioning task, its core objective is to achieve accurate target detection for critical objects that influence driving decisions in driving scenarios, rather than detecting all objects without distinction. 
It focuses on locking the spatial positions of critical objects and outputting positioning information applicable to subsequent decision-making, thereby providing fundamental support for key links of autonomous driving systems. 
This task not only requires accurately identifying the categories of critical objects from complex driving environment but also demands clarifying their specific positions.
Specifically, related work~\cite{malla2023drama} is a representative dataset in this task field. 
Its core function is to identify critical objects in driving scenarios and output corresponding 2D coordinates. Unlike conventional datasets that seek comprehensiveness in object detection, it focuses on objects that directly impact driving decisions, such as sudden obstacles, important traffic signs, and pedestrians crossing the road.

\subsubsection{Status Prediction}

\textbf{Intent Prediction}
A core task in autonomous driving perception and planning, its primary goal is to forecast the future driving behaviors of surrounding traffic elements in dynamic driving scenarios. 
By analyzing historical motion data, current states, and contextual information of traffic elements, this task enables autonomous driving systems to proactively anticipate potential action trends—such as whether a preceding vehicle will turn, a pedestrian will cross the road, or an adjacent vehicle will change lanes—thereby laying a critical foundation for safe path planning, collision avoidance, and adaptive speed adjustment. 
This task not only requires capturing short-term motion patterns of traffic elements but also demands integrating environmental context to improve the accuracy and robustness of behavior predictions, effectively reducing the uncertainty of autonomous driving systems in complex interaction scenarios.
Specifically, DriveLM~\cite{sima2024drivelm} serves as a dataset for advancing intent prediction, with a dual focus on two critical aspects of traffic element behavior analysis. 
First, it supports traffic elements motion prediction, which models the future motion of traffic elements based on multi-frame sequential data. 
Second, it emphasizes object interaction between traffic elements, capturing the interdependencies between different traffic participants. 
Complementing DriveLM, MME-Realworld~\cite{zhang2024mme} focuses on enhancing the practicality of intent prediction in real-world driving environments, with a dedicated emphasis on traffic elements intention prediction. 
These datasets form a complementary framework for intent prediction in autonomous driving. 

\subsubsection{Driving Planning}

\textbf{Action Decision}
The task's primary goal is to predict the ego vehicle’s meta actions in dynamic driving scenarios. 
Meta actions refer to high-level decision-making directions that guide the ego vehicle’s subsequent operations.
To determine the most appropriate high-level action strategy, this task requires the autonomous driving system to analyze information. 
For this task, two key requirements must be met: the predicted meta actions need to align with real-world traffic rules, and multiple objectives such as safety priority and driving efficiency must be balanced.
Specifically, DriveLM~\cite{sima2024drivelm} focus on two critical dimensions of ego vehicle meta action prediction. 
First, it supports the classification of safe or unsafe action by labeling whether the ego vehicle’s potential meta actions conform to safety standards. 
Second, it emphasizes the definition of goal action, which associates meta actions with specific driving goals. 
MME-Realworld~\cite{zhang2024mme} focuses on enhancing the adaptability of action decision in real-world scenarios, with a dedicated emphasis on ego intention prediction. This dataset is built on a large number of real driving cases that cover urban, suburban and highway scenarios.
In addition, IDKB~\cite{lu2025can} provides critical knowledge support for action decision by integrating information on vehicle control and driving techniques.
These datasets collectively form a complementary framework for action decision in autonomous driving, covering safety validation, goal alignment, real-scenario adaptation, and control knowledge integration.

\textbf{Driving Reasoning}
Different from action decision, driving reasoning is to predict the ego vehicle’s driving decisions while explicitly outputting the corresponding reasoning process. 
This task aims to build interpretable decision-making cognition, not only determining the optimal driving actions for the ego vehicle but also clarifying the logical basis behind decisions — such as how environmental factors, traffic rules, and safety priorities jointly guide the generation of specific decisions—thereby enhancing the transparency and trustworthiness of autonomous driving systems. 
Specifically, multiple datasets focus on the integration of decision suggestion and reasoning explanation for Driving Reasoning. 
CODA-LM~\cite{chen2025automated} supports the task of formulating driving advice, which requires models to first accurately perceive both the general context and regional details of the current driving environment, then provide the ego vehicle with optimal driving suggestions while implicitly embedding the reasoning logic in the advice.
Nuinstruct~\cite{ding2024holistic} complements this by focusing on high-level decision-making and safety reasoning.
A novel dataset~\cite{marcu2024lingoqa} is built from recorded driving sessions featuring notable events that trigger changes in the car’s behavior—including decelerations, accelerations, lane changes, narrow gap passes, and turns. 
Each event is paired with two key pieces of information: descriptions of the car’s current action and the corresponding justification for that action. 
Related work~\cite{kim2018textual} extends this to video-based scenarios: given a driving video clip, it requires models to output the ego vehicle’s specific action and the corresponding reasoning.
DriveLM~\cite{sima2024drivelm} provides comprehensive support for driving reasoning through its planning and reasoning task. 
IDKB~\cite{lu2025can} strengthens the reasoning foundation by specializing in encoding knowledge related to driver responsibility and defensive driving.
Collectively, these datasets form a multi-faceted framework for driving reasoning in autonomous driving, and this framework covers decision-advice reasoning, event-justification matching, and knowledge-guided logical analysis.

\section{Training Strategy}
\label{sec:strategy}
\XFM{} develops comprehensive multimodal capabilities through a progressive four-stage training strategy. 
Building on MiMo-VL\footnote{In this work, ``MiMo-VL'' 
refers to the 7B-SFT-2508 checkpoint, which serves as the base model for \XFM{}. 
The original MiMo-VL technical report reports results for the earlier 7B-SFT-2505 
checkpoint under the same name.} ~\cite{coreteam2025mimovltechnicalreport}, we systematically introduce specialized supervision in embodied AI and autonomous driving, culminating in advanced reasoning capabilities through chain-of-thought fine-tuning and reinforcement learning. This strategy facilitates the model to build upon previously acquired capabilities, enabling robust performance across embodied interaction and autonomous driving domains. The training configuration for each stage is detailed in Table~\ref{table1}.

\subsection{Stage 1: Embodied AI Supervised Fine-tuning}
The initial stage establishes core vision-language understanding and embodied reasoning capabilities by integrating general data with specialized embodied AI datasets. We fine-tune our model on the data sourced from MiMo-VL's training corpus~\cite{coreteam2025mimovltechnicalreport} and our curated embodied datasets covering affordance prediction, high-level task planning, and spatial understanding. This combination enables the model to develop robust spatial reasoning while maintaining broad visual recognition capabilities. The training emphasizes multi-scale understanding from fine-grained object part localization to scene-level spatial relationships. Through this stage, the model acquires essential capabilities for interpreting instructions in physically grounded contexts and reasoning about object interactions in real-world environments.

\subsection{Stage 2: Autonomous Driving Supervised Fine-tuning }
Building upon the embodied AI foundation established in Stage 1, the second stage specializes the model for autonomous driving through intensive driving-specific samples. This phase focuses on developing critical capabilities for understanding dynamic environments, including multi-view spatial reasoning, temporal consistency across video sequences, and complex traffic scenario analysis. The training incorporates diverse driving conditions for environment comprehension, status prediction, and driving planning.
Special attention is given to safety-critical perception tasks, such as identifying hazardous objects, predicting the intentions of traffic participants, and understanding complex road geometries. The model learns to process multi-camera inputs simultaneously, maintain object permanence across frames, and reason about the kinematic relationships among vehicles, pedestrians, and infrastructure elements. This specialized training enables robust performance in autonomous driving scenarios, where precise scene understanding and spatial-temporal reasoning are essential for decision-making.

\begin{table*}[!t]
    \centering
    \vspace{0.2cm}
    \setlength{\tabcolsep}{12pt}
    \renewcommand{\arraystretch}{1.2}
    \resizebox{0.98\textwidth}{!}{%
    \begin{tabular}{@{}l|c|c|c|c}
        \toprule
        \textbf{Stages} & \textbf{Stage 1} & \textbf{Stage 2} & \textbf{Stage 3} & \textbf{Stage 4} \\ 
        \midrule
        \textbf{Dataset} & \makecell[c]{General Data \\ $+$ \\ \textbf{Embodied AI}} & \makecell[c]{Previous \\ $+$ \\ \textbf{Autonomous Driving}} & \makecell[c]{Previous \\ $+$ \\ \textbf{CoT Data}} & \textbf{RL Data} \\
        \midrule
        \textbf{Batch Size} & 512 & 512 & 512 & 32 \\
        \textbf{Learning Rate} & 2$\times 10^{-6}$ & 2 $\times 10^{-6}$ & 2 $\times 10^{-6}$ & 1 $\times 10^{-6}$ \\
        \textbf{Optimizer} & AdamW & AdamW & AdamW & AdamW \\
        \textbf{Weight Decay} & 0.05 & 0.05 & 0.05 & 0.0 \\
        \textbf{LR Schedule} & Cosine & Cosine & Cosine & Cosine \\
        \textbf{Max Sequence Length} & 32768 & 32768 & 32768 & 32768 \\
        \textbf{Trainable Components} & All & All & All & All \\
        \bottomrule
    \end{tabular}
    }
    \caption{Detailed configuration for each training stage of \XFM{}.}
    \label{table1}
    \vspace{1mm}
\end{table*}

\subsection{Stage 3: Chain-of-Thought Reasoning Supervised Fine-tuning}
The third stage enhances the model's reasoning transparency and logical coherence through chain-of-thought (CoT) methodology with generated reasoning samples. We sample a subset from our training data for CoT data generation by employing a structured approach that breaks down multifaceted problems into sequential reasoning steps. Each training sample includes explicit reasoning chains that demonstrate how to analyze situational context, generate candidate solutions, evaluate alternatives, and justify final decisions. For embodied domains, this includes reasoning about object affordances and spatial constraints; for autonomous driving, it encompasses risk assessment, trajectory evaluation, and behavior justification. This stage significantly improves the model's ability to handle multi-step problems that require integrating perceptual information with domain knowledge and logical inference, producing interpretable reasoning processes that enhance trustworthiness in safety-critical applications.

\subsection{Stage 4: Reinforcement Learning Fine-Tuning}
In the final stage, we employ reinforcement learning (RL) fine-tuning to further enhance the model's precision and reliability. 
Building upon the visual perception and reasoning capabilities established in previous stages, this phase optimizes both correctness and response quality on the carefully curated samples spanning diverse multimodal tasks. 
We implement the Group Relative Policy Optimization (GRPO)~\cite{guo2025deepseek} algorithm, which samples multiple responses for each query and computes advantages through group-wise normalization.
The training specifically addresses corner cases and failure modes identified in previous stages, with particular focus on enhancing performance in visual reasoning, spatial grounding, and complex instruction following. 
To enable the multi-task mixed training, we design different reward signals for tasks with deterministic solutions. For multi-choice visual reasoning, the reward is evaluated by exact answer matching; For spatial grounding and pointing, it is assessed via IoU between predicted and ground-truth boxes or point in mask. The format compliance is verified through strict template checks. 
Through this targeted, rule-driven optimization, the model systematically learns to produce more precise and reliable outputs.

\section{Evaluation}
\label{sec:evaluation}

In this section, we evaluate the performance of \XFM{} through both quantitative and qualitative analyses. The quantitative comparisons involve objective assessments against a diverse range of established academic and industry benchmarks for embodied AI and autonomous driving, enabling direct empirical comparisons with leading models. Complementarily, the qualitative evaluation showcases \XFM{}'s practical efficacy in real-world tasks, highlighting its deployment in complex robotic and autonomous driving scenarios and providing tangible evidence of its ability to translate learned capabilities into effective performance. Next, we will explore the experimental results and analyses in detail.

\subsection{Quantitative Comparisons on Benchmarks}

\subsubsection{Embodied AI Benchmarks}
To evaluate the embodied capabilities of \XFM{}, we conduct a comprehensive evaluation across three core domains: affordance prediction, task planning, and spatial understanding. 
The results in Table~\ref{table:aff_plan} and Table~\ref{table:Spatial_table} show that \XFM{} delivers competitive results, showing particular strength in affordance prediction and spatial understanding compared to both general-purpose multimodal models and specialized embodied models.

\textbf{Affordance Prediction Capability}
We evaluate the model's ability to infer actionable interaction possibilities from visual scenes using five specialized benchmarks, each targeting a distinct aspect of affordance understanding. 
Roborefit~\cite{lu2023vl} assesses object reference capabilities in robotic manipulation scenarios, specifically testing the identification of target with multiple visually similar objects that can only be distinguished through relational reasoning. 
Where2Place~\cite{yuan2024robopoint} evaluates the model's ability to localize semantically appropriate and physically feasible free space for object placement based on the spatial relationships described in instructions. 
The pointing subset of VABench~\cite{yuan2025seeing} (VABench-Point) evaluates a model's precision in grounding natural language commands to specific coordinate locations for robotic manipulation, requiring predicted points to fall within human-annotated regions of target objects or free space.
Part-Afford~\cite{myers2015affordance} examines fine-grained object part-level affordance recognition across diverse object categories. 
RoboAfford-Eval~\cite{tang2025roboafford} provides a comprehensive benchmark covering both object and spatial affordance types in robotic manipulation contexts.

As shown in Table~\ref{table:aff_plan}, \XFM{} achieves SOTA performance across all affordance prediction benchmarks. Specially, our model outperforms other embodied VLMs by a large margin on the VABench-Point~\cite{yuan2025seeing}, Part-Afford~\cite{myers2015affordance}, and RoboAfford-Eval~\cite{tang2025roboafford} benchmarks, demonstrating strong capabilities in fine-grained affordance reasoning. The results highlight \XFM{}'s effectiveness in interpreting natural language commands for physical interactions, establishing a solid foundation for real-world embodied manipulation tasks.

\begin{table}[!t]
\centering

\setlength{\tabcolsep}{3pt} 

\renewcommand{\arraystretch}{1.4}
\resizebox{\linewidth}{!}{%

\begin{tabular}{lc|*{5}{S[table-format=2.2, table-space-text-post={*}]} | *{3}{S[table-format=2.2, table-space-text-post={*}]}S[table-format=2.2, table-space-text-post={*}]}
\toprule
\multicolumn{2}{c}{\textbf{Model Info}} & \multicolumn{5}{c}{\textbf{Affordance}} & \multicolumn{3}{c}{\textbf{Planning}} & \multicolumn{1}{c}{} \tabularnewline
\cmidrule(lr){1-2}\cmidrule(lr){3-7}\cmidrule(lr){8-10}
\multicolumn{1}{c}{\textbf{Names}} & \multicolumn{1}{c}{\textbf{Params}} &
\multicolumn{1}{c}{\textbf{RoboRefIt}} & \multicolumn{1}{c}{\textbf{Where2Place}} & \multicolumn{1}{c}{\textbf{VABench-Point}} & \multicolumn{1}{c}{\textbf{Part-Afford}} & \multicolumn{1}{c}{\textbf{RoboAfford-Eval}} &
\multicolumn{1}{c}{\textbf{EgoPlan2}} & \multicolumn{1}{c}{\textbf{RoboVQA}} & \multicolumn{1}{c}{\textbf{Cosmos}} & \multicolumn{1}{c}{} \tabularnewline 
\midrule
\rowcolor[HTML]{FFE0CC}
\multicolumn{11}{c}{\textit{Open-Source Models}} \tabularnewline 
\midrule
\addlinespace[2pt] 

MiMo-VL~\cite{coreteam2025mimovltechnicalreport} & 7B & 68.92{*} & 29.60{*} & 35.13{*} & 15.98{*} & 43.88{*} & 34.14{*} & 35.27{*} & 50.91{*} & \tabularnewline
Qwen2.5-VL~\cite{bai2025qwen2} & 7B & {\underline{80.42}}{*} & 42.00{*} & 24.50{*} & 42.65{*} & 16.10 & 39.67{*} & {\underline{57.17}}{*} & 53.70 & \tabularnewline
InternVL3.5~\cite{wang2025internvl3} & 8B & 39.38{*} & 16.49{*} & 11.93{*} & 16.92{*} & 28.72{*} & 42.92 & 28.55{*} & 48.24{*} & \tabularnewline

\midrule
\rowcolor[HTML]{FFE0CC}
\multicolumn{11}{c}{\textit{Closed-Source (Proprietary) Models}} \tabularnewline
\midrule
\addlinespace[2pt] 

GPT-4o~\cite{gpt4o} & {--} & 14.15{*} & 20.41 & 13.67{*} & 13.25{*} & 20.50 & 41.79 & 34.50 & 53.30 & \tabularnewline
Claude-Sonnet-4~\cite{claude4} & {--} & 25.75{*} & 25.63 & 15.83{*} & 10.20{*} & 18.40{*} & 41.26 & 8.51 & 46.53{*} & \tabularnewline
Gemini2.5-Pro~\cite{gemini25pro} & {--} & 38.44{*} & 42.38 & 27.92{*} & 25.53{*} & 23.40{*} & 42.85 & 33.90 & 48.64 & \tabularnewline
Qwen-VL-Max~\cite{bai2023qwenvlversatilevisionlanguagemodel} & {--} & 70.31{*} & 18.92{*} & {\underline{41.50}}{*} & {\underline{65.35}}{*} & 37.92{*} & {\textbf{44.68}}{*} & 54.37{*} & {\textbf{66.36}}{*} & \tabularnewline
\midrule
\rowcolor[HTML]{FFE0CC}
\multicolumn{11}{c}{\textit{Embodied Models}} \tabularnewline
\midrule
\addlinespace[2pt]

Cosmos-Reason1~\cite{azzolini2025cosmos} & 7B & 47.25{*} & 11.40 & 5.96{*} & 5.10{*} & 12.62{*} & 26.87 & {--} & {\underline{61.80}} & \tabularnewline
VeBrain~\cite{luo2025vebrain} & 8B & 32.15{*} & 12.27 & 1.67{*} & 32.10{*} & 2.08{*} & 27.30 & {--} & {--} & \tabularnewline
Magma~\cite{magma} & 8B & 4.95 & 9.93 & {--} & {--} & {--} & 4.09 & {--} & {--} & \tabularnewline
RoboBrain-1.0~\cite{ji2025robobrain} & 7B & {--} & 53.04 & {--} & {--} & {--} & {--} & {--} & {--} & \tabularnewline
RoboBrain-2.0~\cite{baairobobrainteam2025robobrain20technicalreport} & 7B & 70.40{*} & {\underline{63.59}{\ }{\ }} & 26.67{*} & 31.20{*} & {\underline{51.46}}{*} & 33.23 & 46.32{*} & 33.82{*} & \tabularnewline
\rowcolor[HTML]{FFE0CC}\textbf{\XFM{} (Ours)} & 7B & {\textbf{82.30}{\ }{\ }} & {\textbf{63.60}{\ }{\ }} & {\textbf{46.93}{\ }{\ }} & {\textbf{65.50}{\ }{\ }} & {\textbf{69.81}{\ }{\ }} & {\underline{43.00}{\ }{\ }} & {\textbf{61.99}{\ }{\ }} & 56.80 & \tabularnewline
\bottomrule
\end{tabular}
}
\vspace{1ex}
\setlength{\abovecaptionskip}{-0.1pt}
\caption{\textbf{Comparison of \XFM{} with other models on affordance and planning benchmarks.}
We evaluate the model against various open-source, closed-source, and specialized
embodied VLMs to show a comprehensive performance overview.
Results marked with * are obtained using our evaluation framework.
The best results among the listed models are \textbf{bolded} and the second-best is \underline{underlined}.}
\label{table:aff_plan}
\vspace{-1.0em}
\end{table}

\begin{table}[t]
\centering
\setlength{\tabcolsep}{3pt} 
\renewcommand{\arraystretch}{1.25}
\resizebox{\linewidth}{!}{%
\begin{tabular}{lc|*{9}{S[table-format=2.2, table-space-text-post={*}]}}
\toprule
\multicolumn{2}{c}{\textbf{Model Info}} & \multicolumn{9}{c}{\textbf{Spatial}} \tabularnewline
\cmidrule(lr){1-2}\cmidrule(lr){3-11}
\multicolumn{1}{c}{\textbf{Names}} & \multicolumn{1}{c}{\textbf{Params}} &
\multicolumn{1}{c}{\textbf{CV-Bench}} & \multicolumn{1}{c}{\textbf{ERQA}} & \multicolumn{1}{c}{\textbf{EmbSpatial}} &
\multicolumn{1}{c}{\textbf{SAT}} & \multicolumn{1}{c}{\textbf{RoboSpatial}} &
\multicolumn{1}{c}{\textbf{RefSpatial}} & \multicolumn{1}{c}{\textbf{CRPE}} & \multicolumn{1}{c}{\textbf{MetaVQA}} & \multicolumn{1}{c}{\textbf{VSI-Bench}} \tabularnewline
\midrule
\rowcolor[HTML]{FFE0CC}
\multicolumn{11}{c}{\textit{Open-Source Models}} \tabularnewline
\midrule
MiMo-VL~\cite{coreteam2025mimovltechnicalreport} & 7B & 81.69{*} & 41.50{*} & 71.29{*} & 59.33{*} & 50.05{*} & 29.79{*} & 72.33{*} & 60.89{*} & 42.10{*} \tabularnewline
Qwen2.5-VL~\cite{bai2025qwen2} & 7B & 75.40 & 38.80 & 70.25 & 52.00 & 49.33 & 38.00{*} & 76.40 & 58.63{*} & 35.90 \tabularnewline
InternVL3.5~\cite{wang2025internvl3} & 8B & 81.46{*} & 41.00 & 70.26{*} & 59.33{*} & 37.77{*} & 16.80{*} & 75.10 & 60.52{*} & {\textbf{56.30}{\ }{\ }} \tabularnewline
\midrule
\rowcolor[HTML]{FFE0CC}
\multicolumn{11}{c}{\textit{Closed-Source (Proprietary) Models}} \tabularnewline
\midrule
GPT-4o~\cite{gpt4o} & {--} & 78.63 & 32.48 & 71.92 & 66.67 & 44.42 & 8.78 & {\underline{76.60}{\ }{\ }} & 62.80 & 43.60 \tabularnewline
Claude-Sonnet-4~\cite{claude4} & {--} & 78.12 & 45.75 & 73.18{*} & 75.33 & 53.72{*} & 20.78{*} & 73.20{*} & 61.70{*} & 47.02 \tabularnewline
Gemini2.5-Pro~\cite{gemini25pro} & {--} & 84.59 & {\textbf{51.02}{\ }{\ }} & {\textbf{78.74}{\ }{\ }} & {\textbf{79.33}{\ }{\ }} & {\underline{59.87}{\ }{\ }} & {\underline{38.16}{\ }{\ }} & 72.17{*} & {\textbf{69.47}}{*} & 47.81 \tabularnewline
Qwen-VL-Max~\cite{bai2023qwenvlversatilevisionlanguagemodel} & {--} & 79.42{*} & 40.91 & 71.53{*} & 56.67{*} & 46.92{*} & {--} & 68.68{*} & 61.52{*} & 41.80{*} \tabularnewline
\midrule
\rowcolor[HTML]{FFE0CC}
\multicolumn{11}{c}{\textit{Embodied Models}} \tabularnewline
\midrule
Cosmos-Reason1~\cite{azzolini2025cosmos} & 7B & 68.57 & 39.09 & 65.22 & 60.67 & 38.81 & 5.44 & 69.42{*} & 65.81{*} & 25.64 \tabularnewline
VeBrain~\cite{luo2025vebrain} & 8B & 79.68 & 37.29 & 70.52 & 58.00 & 42.48 & 0.30 & 68.65{*} & 55.38{*} & 26.30 \tabularnewline
Magma~\cite{magma} & 8B & 65.88 & 25.73 & 64.59 & 71.33 & 33.71 & 4.50 & {--} & {--} & 12.65 \tabularnewline
RoboBrain-1.0~\cite{ji2025robobrain} & 7B & 76.22 & 36.52 & 68.13 & 59.33 & 51.53 & 9.92 & {--} & {--} & 31.12 \tabularnewline
RoboBrain-2.0~\cite{baairobobrainteam2025robobrain20technicalreport} & 7B & {\underline{85.75}{\ }{\ }} & 30.31 & \underline{76.32}{\ }{\ } & 75.33 & 54.23 & 32.50 & 71.58{*} & 61.11{*} & 36.10 \tabularnewline
\rowcolor[HTML]{FFE0CC}\textbf{\XFM{} (Ours)} & 7B & {\textbf{88.82}{\ }{\ }} & {\underline{46.75}{\ }{\ }} & {76.24{\ }{\ }} & {\underline{78.67}{\ }{\ }} & {\textbf{61.76}{\ }{\ }} & {\textbf{48.00}{\ }{\ }} & {\textbf{77.15}{\ }{\ }} & {\underline{67.33}{\ }{\ }} & {\underline{48.49}{\ }{\ }} \tabularnewline
\bottomrule
\end{tabular}
}
\vspace{1ex}
\caption{\textbf{Comparison of \XFM{} with other models on spatial benchmarks.}
We evaluate the model against various open-source, closed-source, and specialized
embodied VLMs to show a comprehensive performance overview.
Results marked with * are obtained using our evaluation framework.
The best results among the listed models are \textbf{bolded} and the second-best is \underline{underlined}.}
\label{table:Spatial_table}
\end{table}

\textbf{Task Planning Capability}
Task planning capability reflects a model's proficiency in translating abstract instructions into executable action sequences, requiring long-horizon reasoning, causal inference, and compositional logic. We evaluate this capacity through three specialized benchmarks: EgoPlan2~\cite{qiu2024egoplan}, RoboVQA~\cite{sermanet2024robovqa}, and Cosmos-Reason1~\cite{azzolini2025cosmos}. EgoPlan2~\cite{qiu2024egoplan} assesses long-horizon planning from video, requiring a model to generate the correct action based on current observed image to reach a goal. RoboVQA~\cite{sermanet2024robovqa} specifically tests causal reasoning and future state prediction, where a model must answer questions about the outcome of robotic actions, thereby probing its understanding of action consequences. Cosmos-Reason1~\cite{azzolini2025cosmos} presents a multifaceted challenge in compositional reasoning and planning, evaluating the ability to generate complex, multi-step action plans for robotic manipulation tasks requiring both logical and physical understanding.

The comparison results in Table~\ref{table:aff_plan} demonstrates \XFM{}'s advanced task planning capabilities. It outperforms other models on RoboVQA\cite{sermanet2024robovqa}, showcasing its superior ability in causal inference and understanding goal-oriented outcomes. \XFM{} also achieves a highly competitive performance on the long-horizon planning benchmark EgoPlan2~\cite{qiu2024egoplan}. This proficiency in analyzing multi-step action sequences underscores \XFM{}'s effectiveness in long-horizon reasoning, establishing it as a capable foundation for autonomous agents operating in complex, dynamic environments.

\begin{table}[!t]
\centering

\setlength{\tabcolsep}{3pt} 

\renewcommand{\arraystretch}{1.4}
\resizebox{0.93\linewidth}{!}{%

\begin{tabular}{lc|*{4}{S[table-format=2.2, table-space-text-post={*}]} | *{2}{S[table-format=2.2, table-space-text-post={*}]}S[table-format=2.2, table-space-text-post={*}]}
\toprule
\multicolumn{2}{c}{\textbf{Model Info}} & \multicolumn{1}{c}{\textbf{CODA-LM}} & \multicolumn{1}{c}{\textbf{Drama}} &
\multicolumn{1}{c}{\textbf{MME-RealWorld}} & \multicolumn{1}{c}{\textbf{IDKB}} & \multicolumn{1}{c}{\textbf{OmniDrive}} & \multicolumn{1}{c}{\textbf{NuInstruct}} \tabularnewline
\cmidrule(lr){1-2}\cmidrule(lr){3-3}\cmidrule(lr){4-4}\cmidrule(lr){5-5}\cmidrule(lr){6-6}\cmidrule(lr){7-7}\cmidrule(lr){8-8}
\multicolumn{1}{c}{\textbf{Names}} & \multicolumn{1}{c}{\textbf{Params}} &
\multicolumn{1}{c}{\textbf{PER. \& PLA.}} & \multicolumn{1}{c}{\textbf{PER.}} &
\multicolumn{1}{c}{\textbf{PER. \& PRE. \& PLA.}} & \multicolumn{1}{c}{\textbf{PER. \& PLA.}} & \multicolumn{1}{c}{\textbf{PER. \& PLA.}} & \multicolumn{1}{c}{\textbf{PLA.}} \tabularnewline
\midrule
\rowcolor[HTML]{FFE0CC}
\multicolumn{8}{c}{\textit{Open-Source Models}} \tabularnewline
\midrule

MiMo-VL~\cite{coreteam2025mimovltechnicalreport} & 7B & 31.02 & 51.22 & 54.05 & 18.90 & 3.06 & 0.68\tabularnewline
InternVL3.5~\cite{wang2025internvl3} & 8B & 32.61 & 0.08 & 49.20 & 19.66 & 12.81 & 0.00 \tabularnewline
InternVL3.5~\cite{wang2025internvl3} & 38B & 28.14 & 0.00 & 54.95 & 18.96 & 16.43 & 0.24 \tabularnewline
Qwen2.5-VL~\cite{bai2025qwen2} & 7B & 35.75 & 54.32 & 58.60 & 13.44 & 10.07 & 0.43 \tabularnewline
Qwen2.5-VL~\cite{bai2025qwen2} & 72B & 35.80 & 0.00 & 50.78 & 18.41 & 7.32 & 5.67 \tabularnewline
\midrule
\rowcolor[HTML]{FFE0CC}
\multicolumn{8}{c}{\textit{Closed-Source (Proprietary) Models}} \tabularnewline
\midrule
GPT-4o~\cite{gpt4o} & {--} & 34.18 & 0.00 & 58.00 & 20.65 & 19.22 & 7.08\tabularnewline
Gemini2.5-Pro~\cite{gemini25pro} & {--} & 53.21 & 0.24 & \textbf{67.00{\ }{\ }} & {\underline{23.21}{\ }{\ }} & 10.87 & 53.20 \tabularnewline
Qwen-VL-Max~\cite{bai2023qwenvlversatilevisionlanguagemodel} & {--} & 37.72 & 0.16 & {\underline{61.65}{\ }{\ }} & 16.82 & 10.57 & 0.03 \tabularnewline
\midrule
\rowcolor[HTML]{FFE0CC}
\multicolumn{8}{c}{\textit{Autonomous Driving Models}} \tabularnewline
\midrule
Specialist Model$^\dagger$ & {--} & 45.46 & {\underline{68.40}{\ }{\ }} & {--} & {--} & 40.97 & 35.20  \tabularnewline
DriveLMM-o1~\cite{ishaq2025drivelmm} & 8B &  51.53 & 0.00 & 49.59 & 11.38 & 9.77 & 2.88 \tabularnewline
RoboTron-Drive~\cite{huang2025robotron} & 8B &  {\underline{58.10}}{*} & 0.00 & 41.30 & 8.32 &  \textbf{48.76}{*} & \underline{83.00}{*}\tabularnewline
\rowcolor[HTML]{FFE0CC}\textbf{\XFM{} (Ours)} & 7B & \textbf{58.55{\ }{\ }} & \textbf{76.14{\ }{\ }} & 60.25 & {\textbf{43.42}{\ }{\ }} & {\underline{45.21}{\ }} & \textbf{83.58{\ }{\ }}\tabularnewline

\bottomrule
\end{tabular}
}
\vspace{1ex}
\caption{\textbf{Comparison of \XFM{} with other models on four single-view image benchmarks and two multi-view video benchmarks in autonomous driving.}
We evaluate the model against various open-source, closed-source, and specialized  autonomous driving  VLMs to show a comprehensive performance overview.
Results marked with * are obtained using our evaluation framework.
${^\dagger}$Specialist models
correspond to the performance of different models~\cite{zhang2024minidrive,malla2023drama,wang2024omnidrive,ding2024holistic}.
The best results among the listed models are \textbf{bolded} and the second-best is \underline{underlined}. Here PER. denotes perception, PRE. denotes prediction, PLA. denotes planning.}
\label{table:ad_table1}
\vspace{-1em}
\end{table}

\textbf{Spatial Understanding Capability}
Spatial understanding capability measures the model's proficiency in parsing spatial relationships, direction, distance, size, count, configuration, and etc, which is essential for navigation, object localization, and scene comprehension. We evaluate this capacity across nine benchmarks organized into three categories: spatial relationship reasoning (EmbSpatial~\cite{du2024embspatial}, RoboSpatial~\cite{song2025robospatial}, SAT~\cite{ray2024sat}, VSI-Bench~\cite{yang2025thinking}, CRPE-relation~\cite{wang2024all}), spatial language grounding (RefSpatial-Bench~\cite{zhou2025roborefer}, ERQA~\cite{team2025gemini}, the VQA subset of MetaVQA~\cite{wang2025embodied}), and comprehensive spatial intelligence (CV-Bench~\cite{tong2024cambrian}).

We present the comparison results in Table~\ref{table:Spatial_table}. It can be seen that \XFM{} achieves state-of-the-art results on CV-Bench~\cite{tong2024cambrian} among comprehensive spatial intelligence tasks, and leads performance on RoboSpatial~\cite{song2025robospatial}, RefSpatial-Bench~\cite{zhou2025roborefer}, and the relation subset of CRPE~\cite{wang2024all} in spatial relationship reasoning and language grounding. 
Meanwhile, \XFM{} achieves highly competitive performance on EmbSpatial~\cite{du2024embspatial}, SAT~\cite{ray2024sat}, and the VQA subset of MetaVQA~\cite{wang2025embodied}. 
These results across diverse spatial reasoning tasks validate \XFM{}'s capacity for embodied reasoning in the physical world, demonstrating robust understanding of spatial relationships, object references, and contextual question answering.

\subsubsection{Autonomous Driving Benchmarks}

To comprehensively assess the autonomous driving capabilities of \XFM{}, we conduct a systematic evaluation across three critical dimensions: Perception Capability, Prediction Capability, and Planning Capability. 
Specifically, we evaluate the model’s performance on 12 benchmarks across 4 data types in terms of its ability to understand complex traffic scenes, predict the behaviors of dynamic road agents, and generate safe and efficient driving suggestions. 
As shown in Table~\ref{table:ad_table1} and Table~\ref{table:ad_table2}, \XFM{} delivers exceptional performance, outperforming general large multimodal models and specialized autonomous driving models.

\begin{table}[t]
\centering
\setlength{\tabcolsep}{2.2pt}
\renewcommand{\arraystretch}{1.25}
\resizebox{0.93\linewidth}{!}{%
\begin{tabular}{lc|*{3}{S[table-format=2.2, table-space-text-post={*}]}|*{3}{S[table-format=2.2, table-space-text-post={*}]}}
\toprule
\multicolumn{2}{c}{\textbf{Model Info}} & \multicolumn{1}{c}{\textbf{DriveLM}} & \multicolumn{1}{c}{\textbf{MAPLM}} & \multicolumn{1}{c}{\textbf{nuScenes-QA}} &
\multicolumn{1}{c}{\textbf{LingoQA}} & \multicolumn{1}{c}{\textbf{BDD-X}}  &
\multicolumn{1}{c}{\textbf{DriveAction}}
\tabularnewline
\cmidrule(lr){1-2}\cmidrule(lr){3-3}\cmidrule(lr){4-4}\cmidrule(lr){5-5}\cmidrule(lr){6-6}\cmidrule(lr){7-7}\cmidrule(lr){8-8}
\multicolumn{1}{c}{\textbf{Names}} & \multicolumn{1}{c}{\textbf{Params}} &
\multicolumn{1}{c}{\textbf{PER. \& PRE. \& PLA.}} & \multicolumn{1}{c}{\textbf{PER.}} & \multicolumn{1}{c}{\textbf{PER.}} &
\multicolumn{1}{c}{\textbf{PER. \& PLA.}} & \multicolumn{1}{c}{\textbf{PLA.}} &
\multicolumn{1}{c}{\textbf{PER. \& PLA.}}
\tabularnewline
\midrule
\rowcolor[HTML]{FFE0CC}
\multicolumn{8}{c}{\textit{Open-Source Models}} \tabularnewline
\midrule

MiMo-VL~\cite{coreteam2025mimovltechnicalreport} & 7B & 29.76 & 30.95 & 33.94 & 54.80 & 5.66 & \underline{78.89}{\ }{\ }  \tabularnewline
InternVL3.5~\cite{wang2025internvl3} & 8B & 29.74 & 14.24 & 17.24 & 46.70 & 6.97 & 78.10  \tabularnewline
InternVL3.5~\cite{wang2025internvl3} & 38B & 24.57 & 23.55 & 3.32 & 57.60 & 7.41 & 76.00 \tabularnewline
Qwen2.5-VL~\cite{bai2025qwen2} & 7B & 25.39 & 24.76 & 25.78 & 55.60 & 11.45 & 73.40 \tabularnewline
Qwen2.5-VL~\cite{bai2025qwen2} & 72B & 29.70 & 45.94 & 26.23 & 62.20 & 9.69 & 74.50 \tabularnewline
\midrule
\rowcolor[HTML]{FFE0CC}
\multicolumn{8}{c}{\textit{Closed-Source (Proprietary) Models}} \tabularnewline
\midrule
GPT-4o~\cite{gpt4o} & {--} & 41.21 & 26.64 & 34.26 & 56.00 & 12.38 & 72.52  \tabularnewline
Gemini2.5-Pro~\cite{gemini25pro} & {--} & 39.92 & 26.12 & 16.12 & 64.10 & 4.80 & 73.53 \tabularnewline
Qwen-VL-Max~\cite{bai2023qwenvlversatilevisionlanguagemodel} & {--} & 26.99 & 24.83 & 6.70 & 58.80 & 7.53 & 72.60  \tabularnewline
\midrule
\rowcolor[HTML]{FFE0CC}
\multicolumn{8}{c}{\textit{Autonomous Driving Models}} \tabularnewline
\midrule
Specialist Model$^\dagger$ & {--} & 57.00 & 71.76 & \underline{53.40}{\ }{\ } & 60.80 & \underline{48.61}{\ }{\ }  & {--}  \tabularnewline
DriveLMM-o1~\cite{ishaq2025drivelmm} & 8B & 37.05 & 25.51 & 16.76 & 47.80 & 3.43 & 45.89 \tabularnewline
RoboTron-Drive~\cite{huang2025robotron} & 8B &  \textbf{61.30}{*} & \underline{74.34}{*} & 21.15 & \underline{69.20}{*} & 12.80 & 58.87 \tabularnewline
\rowcolor[HTML]{FFE0CC}\textbf{\XFM{} (Ours)} & 7B & {\underline{57.85}{\ }{\ }} & \textbf{74.52}{\ } & \textbf{56.71}{\ }{\ } & \textbf{69.90}{\ }{\ } & \textbf{52.18}{\ }{\ } & \textbf{80.99}{\ }{\ } \tabularnewline
\bottomrule
\end{tabular}
} 
\vspace{1ex}

\caption{\textbf{Comparison of \XFM{} with other models on three multi-view image benchmarks and three single-view video benchmarks in autonomous driving.}
We evaluate the model against various open-source, closed-source, and specialized  autonomous driving  VLMs to show a comprehensive performance overview.
Results marked with * are obtained using our evaluation framework.
${^\dagger}$Specialist models
correspond to the performance of different models~\cite{zeng2025futuresightdrive,cao2024maplm,qian2024nuscenes,marcu2024lingoqa,jin2023adapt}.
The best results among the listed models are \textbf{bolded} and the second-best is \underline{underlined}. Here PER. denotes perception, PRE. denotes prediction, PLA. denotes planning.}

\label{table:ad_table2}
\vspace{-1em}
\end{table}

\textbf{Perception Capability}
Perception capability aims to evaluate the model’s ability to understand its surrounding environment and serves as the foundation for downstream planning tasks.
This capability requires the model to accurately extract both semantic and geometric information from sensor inputs, thereby constructing a comprehensive and structured representation of the driving scene. 
Different application scenarios impose varying requirements on the granularity of perception: panoramic perception emphasizes holistic scene understanding, including integrated analysis of foreground traffic participants, such as vehicles, pedestrians, and cyclists (\textit{e.g.}, language-guided scene understanding as evaluated in LingoQA~\cite{marcu2024lingoqa}), as well as high-level modeling of road topology, lane layouts, and traffic semantics (\textit{e.g.}, map-level understanding assessed by MAPLM~\cite{cao2024maplm}). 
In contrast, local perception focuses on fine-grained identification and reasoning about specific regions or objects, such as detailed analysis of rare or high-risk scenarios in CODA-LM~\cite{chen2025automated}, accurate grounding and recognition of objects referred to by natural language descriptions in MME-RealWorld~\cite{zhang2024mme}, and precise coordinate-level localization of critical objects within complex dynamic scenes as required by DRAMA ~\cite{malla2023drama}.
This multi-granular perception capability, spanning from scene comprehension to object-level reasoning, forms the foundation for robust understanding in real-world autonomous driving. 
The complete suite of perception benchmarks evaluated includes CODA-LM
~\cite{chen2025automated}, DriveAction~\cite{hao2025driveaction}, DRAMA~\cite{malla2023drama}, MME-RealWorld~\cite{zhang2024mme}, IDKB~\cite{lu2025can}, DriveLM~\cite{sima2024drivelm}, MAPLM~\cite{cao2024maplm}, nuScenes-QA~\cite{qian2024nuscenes}, LingoQA~\cite{marcu2024lingoqa}, and OmniDrive~\cite{wang2024omnidrive}.

\begin{figure*}[!t]
\centering
  \includegraphics[width=0.94\textwidth]{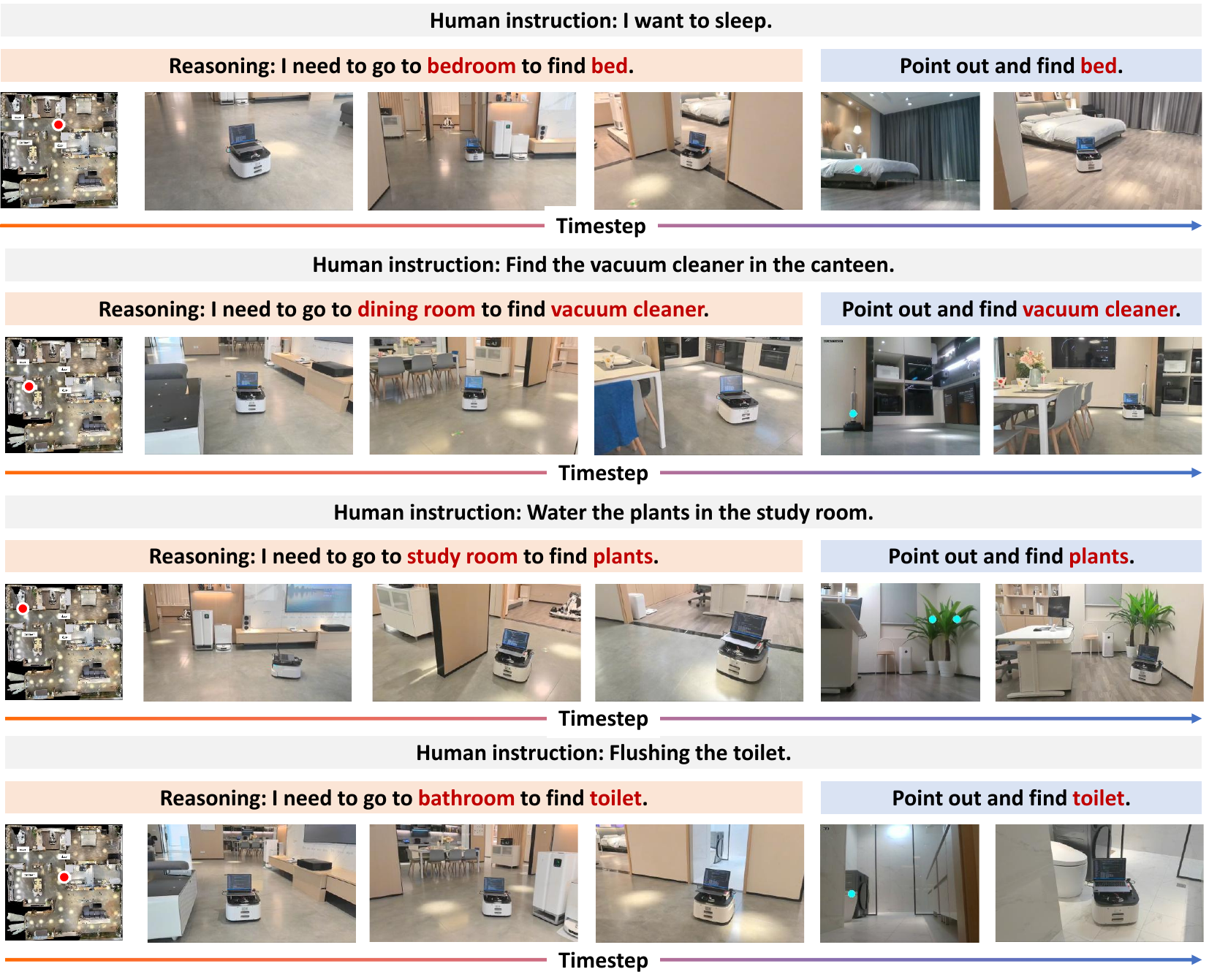}
  \vspace{-6pt}
  \caption{\textbf{Results of deploying \XFM{} to downstream embodied navigation tasks.} The target positions are indicated by cyan points.}
  \label{figure_navigation}
  \vspace{-1em}
\end{figure*}

As shown in Table~\ref{table:ad_table1} and Table~\ref{table:ad_table2}, \XFM{} achieves strong performance across all perception benchmarks, demonstrating state-of-the-art results in panoramic semantic understanding tasks while also exhibiting exceptional robustness in challenging local perception scenarios. 
These results provide compelling evidence that \XFM{} possesses multi-level, high-fidelity environmental perception capabilities, enabling it to effectively adapt to diverse real-world perception demands across varying granularities.

\begin{figure*}[!h]
\centering
  \includegraphics[width=0.96\textwidth]{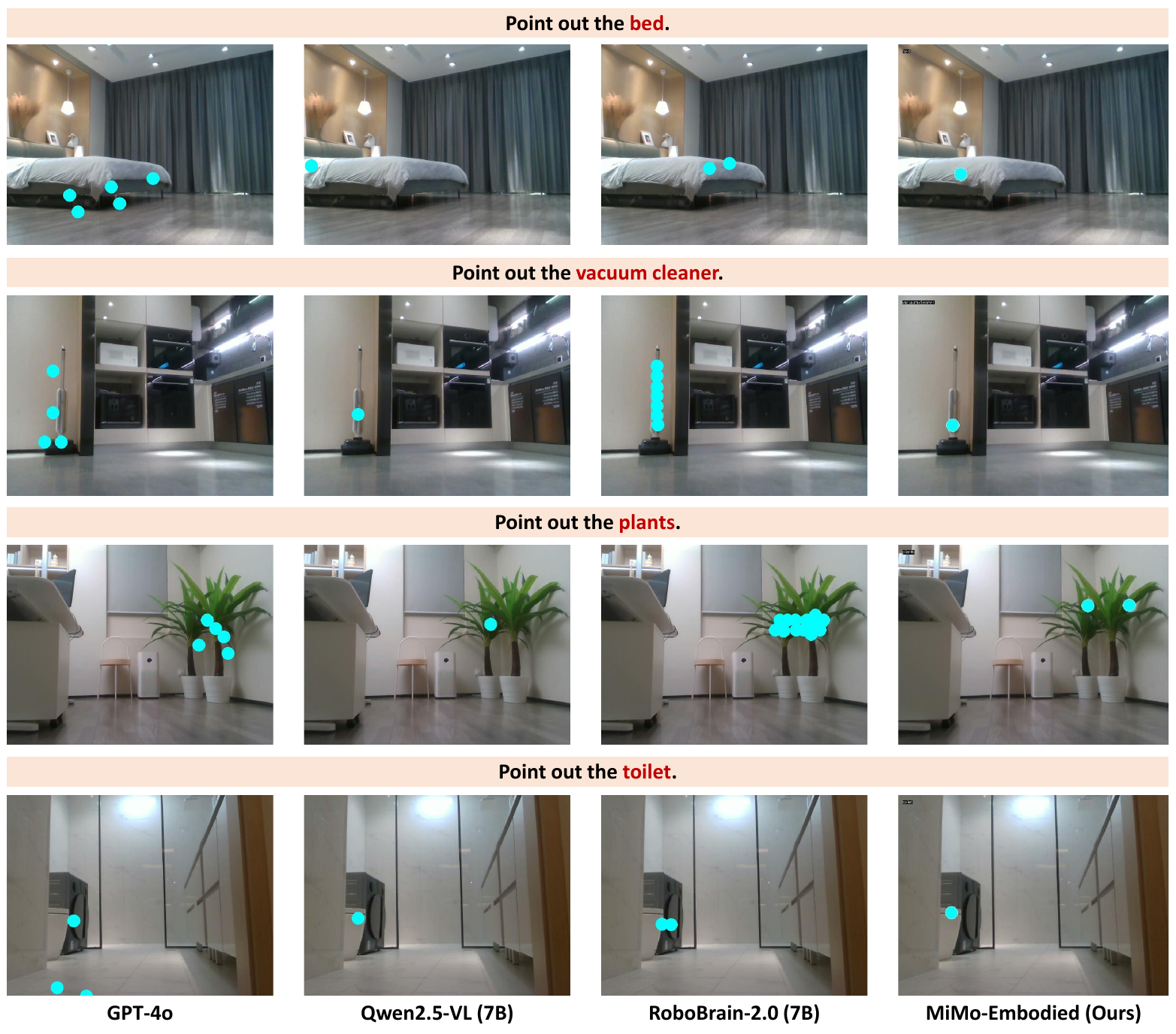}
  \vspace{-6pt}
  \caption{\textbf{Visualization of different models for target object localization in embodied navigation tasks.} The target positions are indicated by cyan points.}
  \label{figure_vis_navigation}
\end{figure*}

\textbf{Prediction Capability}
Prediction capability assesses the model’s capacity to reason about the future evolution of dynamic driving scenes and is fundamental to enabling safe, efficient, and proactive autonomous driving. 
It requires the model to go beyond momentary perception by integrating current and historical observations to accurately infer the latent intentions.
This future-oriented understanding directly informs critical downstream functions such as trajectory planning, collision avoidance, and interactive driving behaviors.
The prediction task encompasses multiple facets. 
At the individual level, it involves forecasting behavioral intent. 
Examples include whether a vehicle will change lanes or a pedestrian intends to cross, and this is typically evaluated in benchmarks such as MME-RealWorld~\cite{zhang2024mme}.
At the interaction level, it requires modeling dynamic relationships among agents, including behaviors like yielding, cooperative maneuvers, and implicit social conventions in multi-agent scenarios. 
These interaction-centric capabilities are explicitly assessed in tasks like DriveLM~\cite{sima2024drivelm} under the framework of object interaction between traffic elements.

Experimental results show that the model achieves strong performance on both the single image benchmark MME-RealWorld and the multi-view images benchmark DriveLM, accurately capturing individual behavioral intentions and effectively modeling complex interactions among multiple agents.

\textbf{Planning Capability}
Planning capability evaluates the model’s ability to synthesize perception and prediction outputs into safe, coherent, and context-aware driving actions.
It directly determine how an autonomous system navigates complex, real-world traffic scenarios.
Effective planning requires not only selecting appropriate low-level control commands including acceleration, braking, or turning but also reasoning about high-level driving strategies in alignment with traffic rules, social conventions, and dynamic scene context.
The first action decision focuses on generating precise, goal-directed driving maneuvers in reaction to the evolving traffic environment. 
Benchmarks such as DriveLM~\cite{sima2024drivelm}, MME-RealWorld~\cite{zhang2024mme} and IDKB~\cite{lu2025can} evaluate this aspect by assessing the model’s ability to select appropriate driving actions in response to the current scene state.
The second capability is driving reasoning, which provides explicit, interpretable justifications for those actions by grounding decisions in scene semantics and driving logic. 
This is assessed in benchmarks including LingoQA~\cite{marcu2024lingoqa}, CODA-LM~\cite{chen2025automated}, OmniDrive~\cite{wang2024omnidrive}, NuInstruct~\cite{ding2024holistic} and BDD-X~\cite{kim2018textual}, which require the model not only to choose the correct action but also to explain why it is appropriate, for instance, yielding to an oncoming vehicle that is expected to proceed first, or stopping due to a red traffic light ahead.

The empirical results, summarized in Table~\ref{table:ad_table1} and Table~\ref{table:ad_table2}, clearly demonstrate that \XFM{} achieves outstanding performance across all planning-oriented benchmarks.
This consistent superiority underscores the model’s strong capacity to not only generate accurate, context-appropriate driving decisions but also produce coherent and interpretable reasoning that aligns with real-world traffic logic and driving norms.

\begin{figure*}[!h]
\centering
  \includegraphics[width=0.98\textwidth]{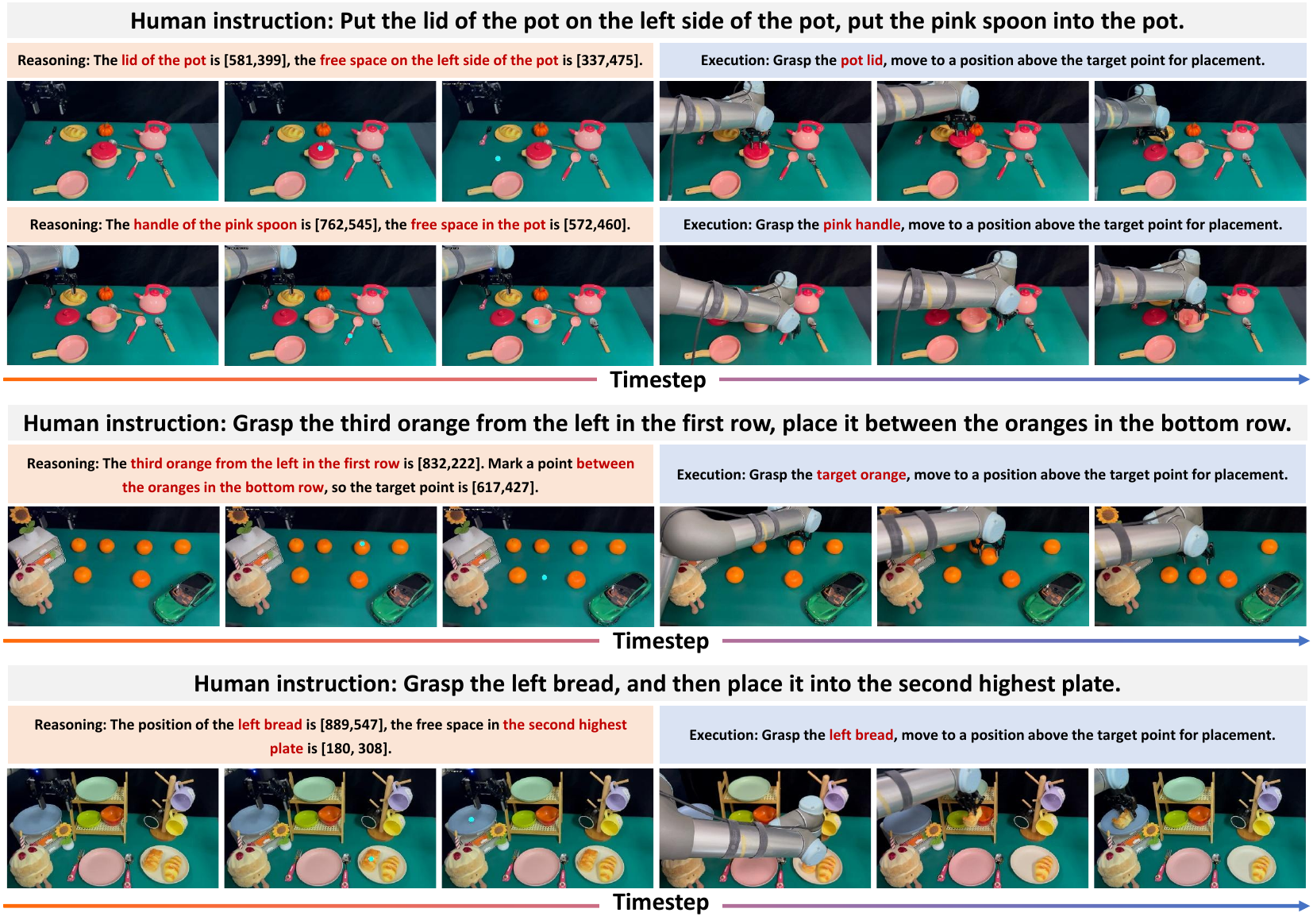}
  \caption{\textbf{Results of deploying \XFM{} to downstream embodied manipulation tasks.} The target positions are indicated by cyan points.}
  \label{figure_manipulation}
\end{figure*}

\begin{figure*}[!t]
\centering
  \includegraphics[width=0.98\textwidth]{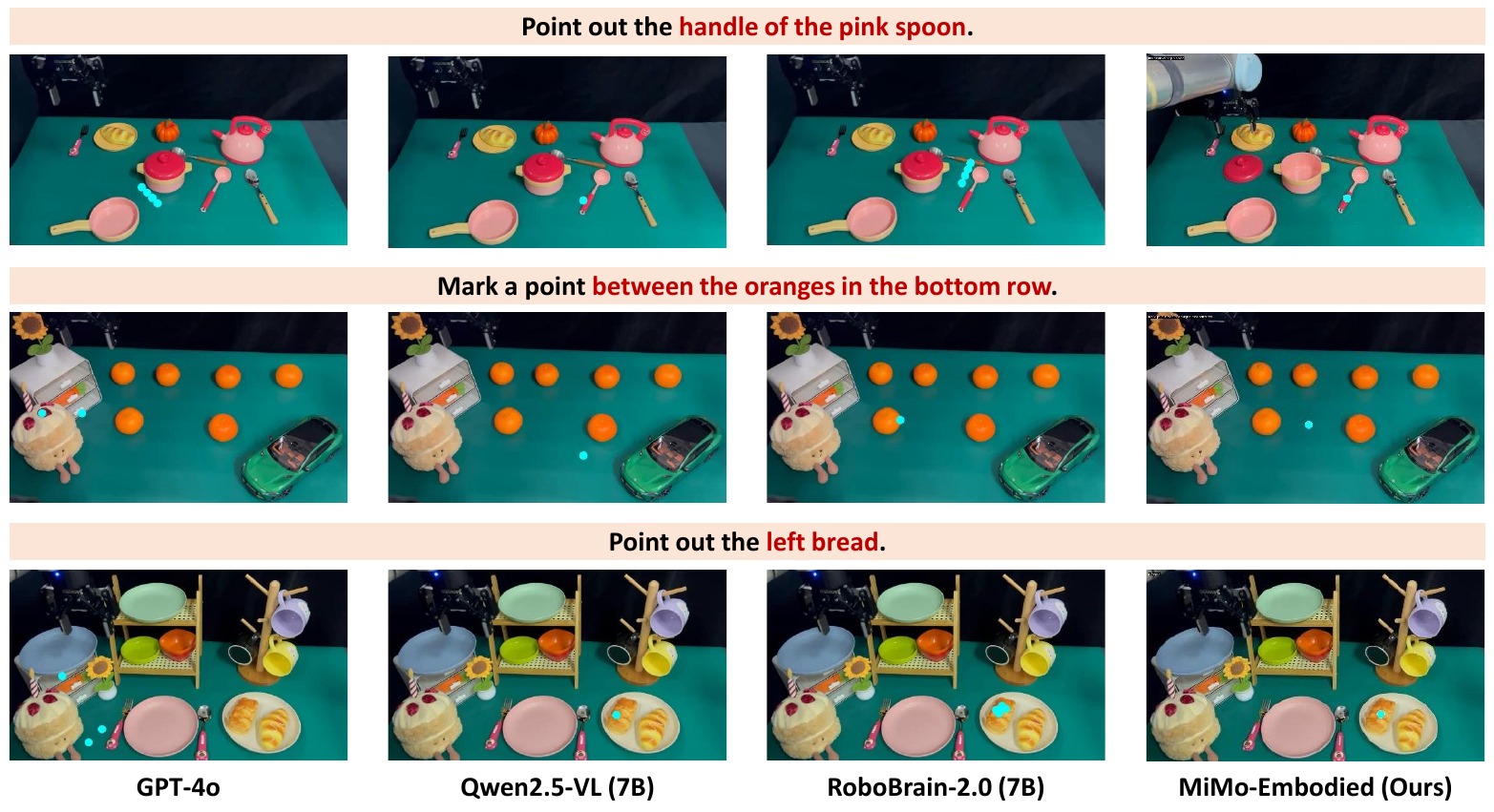}
  \vspace{-6pt}
  \caption{\textbf{Visualization of different models for affordance prediction in embodied manipulation tasks.} The target positions are indicated by cyan points.}
  \label{figure_vis_manipulation}
  \vspace{-1em}
\end{figure*}

\subsection{Qualitative Evaluation in Real-world Tasks}

\subsubsection{\XFM{} for Embodied Navigation \&  Manipulation}
To validate \XFM{}'s practical utility in complex interactive settings, we evaluate its performance on two fundamental downstream applications: embodied navigation and manipulation. 
The evaluation assesses the model's capacity to translate high-level instructions into precise spatial goals.
For navigation, this involves predicting keypoints on a map in global and egocentric views; for manipulation, it entails estimating functionally-grounded interaction points for object affordance.
Our analysis first examines \XFM{}'s capacity for planning and reasoning from long-horizon instructions to generate precise target positions for execution, and then compares these predictions against baseline models to highlight \XFM{}'s superior performance in spatial reasoning and affordance understanding.

\textbf{Embodied Navigation}
We evaluate \XFM{}'s capabilities in downstream embodied navigation tasks using the long-horizon navigation task proposed in $NavA^3$~\cite{zhang2025nava}. Given a high-level human instruction (\textit{e.g.}, ``I want to sleep'' or ``Water the plants in the study room''), the model first needs to infer the spatial region to be explored and the target object to be located. Then, the model marks points on a top-down map, indicating the area the robot needs to navigate to. Subsequently, it identifies and marks the target object in a first-person perspective image, guiding the robot to successfully navigate to the target. We designed four distinct scenarios to evaluate our \XFM{}'s spatial reasoning and object localization capabilities. As shown in Figure ~\ref{figure_navigation}, the visualization results demonstrate that \XFM{} performs well in four household navigation tasks: locating the bed in the bedroom, finding the vacuum cleaner in the dining room, identifying plants in the study room, and locating the toilet in the bathroom. Each task requires the model to decompose the high-level instruction into spatial navigation (region identification on the map) and object localization (precise localization from an egocentric perspective), thus showcasing the model's spatial understanding capabilities.

To intuitively demonstrates the performance, we present a comparative visualization of aforementioned navigation tasks across four multimodal models: GPT-4o~\cite{gpt4o}, Qwen2.5-VL~\cite{bai2025qwen2}, RoboBrain-2.0~\cite{baairobobrainteam2025robobrain20technicalreport}, and our proposed \XFM{}. The visual comparisons in Figure~\ref{figure_vis_navigation} highlight \XFM{}’s enhanced object localization abilities and consistent performance under diverse household scenes. In single object localization tasks, both GPT-4o~\cite{gpt4o} and RoboBrain2.0~\cite{baairobobrainteam2025robobrain20technicalreport} tend to produce multiple scattered or clustered points that may deviates from the object center, indicating limitations in spatial precision. Our \XFM{} model consistently achieves precise center localization, placing points directly at the core of target objects with remarkable accuracy. For plants localization, where the instruction implicitly requires recognizing multiple instances, \XFM{} demonstrates the ability to comprehend the plural nature of the query, successfully identifying and accurately positioning both plant pots in the scene. In contrast, other models show some deficiencies in handling such compositional reasoning, either missing one instance or providing ambiguous localizations. These results highlight \XFM{}'s superior understanding of spatial relationships and linguistic nuances, particularly in complex scenarios requiring both precise object localization and comprehension of implicit semantic cues. The demonstrated capabilities provide a solid foundation for downstream embodied navigation tasks where accurate target identification and spatial reasoning are crucial for successful task completion.

\textbf{Embodied Manipulation} We evaluate \XFM{}'s manipulation capability through a series of hierarchical pick-and-place tasks involving planning, affordance prediction, and spatial reasoning, as shown in Figure~\ref{figure_manipulation}. The model is required to interpret high-level instructions and decompose them into a sequence of precise actions, including spatial reasoning for target localization and object affordance understanding for successful grasping and placement. The first task corresponds to the first two rows in Figure~\ref{figure_manipulation} assess the model's ability in affordance prediction, such as distinguishing the lid of the pot and the handle of pink spoon for grasping and spatial locations for placement. The second task demonstrates the model's advanced counting and spatial reasoning. Given the instruction ``Grasp the third orange from the left in the first row, place it between the oranges in the bottom row,'' the model must identify the correct orange using ordinal counting and then determine the target placement by interpreting the spatial relationship ``between.'' Another task involves placing the left bread in the a specified plate, requiring estimation of height to identify the correct placement location given multiple plates. These results collectively highlight \XFM{}'s ability to integrate object affordance with sophisticated spatial and relational reasoning for embodied manipulation.

We visualize the affordance predictions of GPT-4o~\cite{gpt4o}, Qwen2.5-VL~\cite{bai2025qwen2}, RoboBrain-2.0~\cite{baairobobrainteam2025robobrain20technicalreport}, and our proposed \XFM{} in Figure~\ref{figure_vis_manipulation}. The evaluation focuses on three functionally-grounded tasks: identifying the graspable handle of a pink spoon, locating an intermediate placement position between the oranges in the bottom row, and selecting the leftmost bread for pickup. These tasks assess the models' capacity to interpret affordances and translate linguistic cues into spatially-grounded interaction points. The results reveal substantial differences in functional reasoning accuracy: both GPT-4o~\cite{gpt4o} and RoboBrain-2.0~\cite{baairobobrainteam2025robobrain20technicalreport} frequently generate points misaligned with functionally relevant regions, particularly struggling to identify graspable components and relational positions for object arrangement. While Qwen2.5-VL~\cite{bai2025qwen2} demonstrates outstanding performance in object affordance, it shows limitations in handling spatial affordance in the second task. Notably, \XFM{} consistently identifies functionally appropriate regions across all tasks. These findings underscore \XFM{}'s strong affordance and spatial reasoning capabilities, establishing a reliable foundation for embodied manipulation tasks.

\subsubsection{\XFM{} for Autonomous Driving}

To validate the efficacy of our training strategy, we evaluate our model on trajectory planning, a core capability for autonomous driving systems. Our assessment is designed to be both fair and comprehensive, utilizing a dual-pronged approach: we first establish performance on the challenging public NAVSIM benchmark~\cite{dauner2024navsim} for standardized comparison, and then test the model's capabilities on a large-scale, proprietary dataset rich with diverse, real-world driving scenarios.

\begin{table*}[!t]
  \small
  \centering
  \resizebox{\textwidth}{!}{ 
  \renewcommand{\arraystretch}{1.3} 
  \setlength{\tabcolsep}{8pt}       
  \rowcolors{2}{blue!5}{white}     
  \begin{tabular}{lccccccccc}
    \toprule
    \rowcolor[HTML]{FFE0CC}  
    \textbf{Model} & \textbf{Params} & \textbf{Input} & \textbf{Token} & \textbf{NC}$\uparrow$ & \textbf{DAC}$\uparrow$ & \textbf{TTC}$\uparrow$ & \textbf{Conf}$\uparrow$ & \textbf{EP}$\uparrow$ & \textbf{PDMS} $\uparrow$\\
    \midrule
    InternVL3~\cite{zhu2025internvl3} & 8B & C & 4096 & 97.4 & 93.7 & 93.2 & 100 & 81.2 & 86.0 \\
    \textbf{\XFM{}~(Ours)}       & \textbf{7B} & \textbf{C} & \textbf{796} & \textbf{97.9} & \textbf{94.3} & \textbf{93.8} & \textbf{100} & \textbf{81.7} & \textbf{86.5} \\
    \hline
    ReCogDrive-Large-IL~\cite{recogdrive} & 8B & C &  2304 & 98.1 & 94.5 & \textbf{94.2} & 100 & 80.9 & 86.5  \\
    \textbf{\XFM{}-IL (Ours)}       & \textbf{7B} & \textbf{C} &  \textbf{796}  & \textbf{98.2} & \textbf{94.7} & 94.1 & \textbf{100} & \textbf{82.0} & \textbf{87.4} \\
    \midrule
    ReCogDrive-Large-RL~\cite{recogdrive} & 8B & C & 2304 & 97.9 & 97.3 & 94.9 & 100 & \textbf{86.9} & 90.4 \\
    \textbf{\XFM{}+RL (Ours)}  & \textbf{7B} & \textbf{C} &  \textbf{796} & \textbf{98.3} & \textbf{98.1} & \textbf{95.5} & \textbf{100} & 86.3 & \textbf{91.0} \\
    \bottomrule
  \end{tabular}
  } 
  \caption{We evaluate \XFM{} on the NAVSIM~\cite{dauner2024navsim} planning dataset and present a comparison with the InternVL3~\cite{zhu2025internvl3}. IL (Imitation Learning) denotes the setting where the VLM forwarding results are utilized as conditions for a diffusion-based regression. RL (Reinforcement Learning) indicates the GRPO is further applied on top of IL stage. The best results are highlighted in \textbf{bold}.}
  \label{tab:planning_comparison}
\end{table*}

\begin{figure*}[!h]
\centering
  \includegraphics[width=0.96\textwidth]{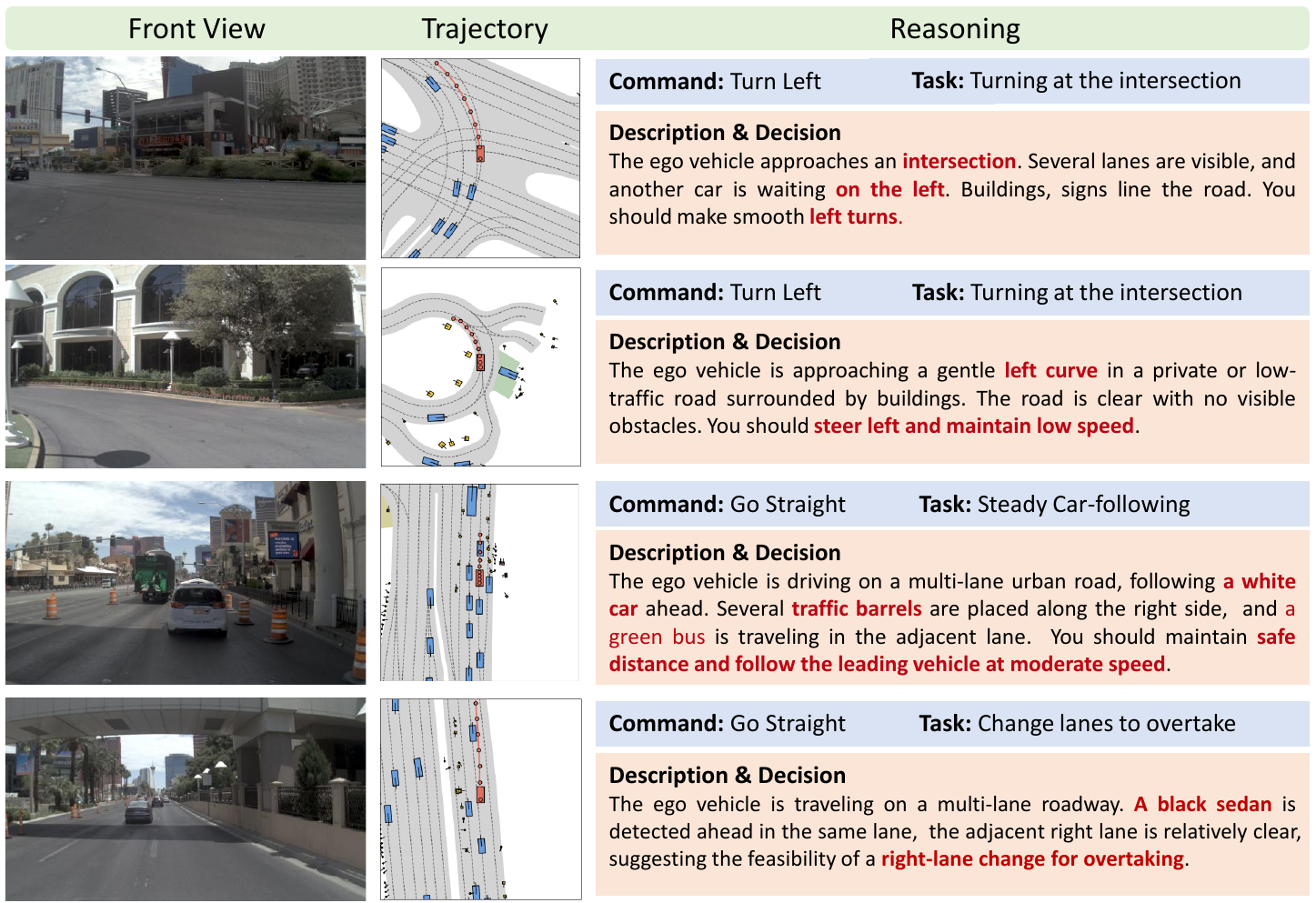}
  \caption{Qualitative results of trajectory planning by \XFM{} on the NAVSIM benchmark.}
  \label{figure_planning_navisim}
  \vspace{-1em}
\end{figure*}

\textbf{Trajectory Planning on Public Benchmark} 
NAVSIM~\cite{dauner2024navsim} is a public  autonomous driving planning dataset widely used to evaluate the performance of planners. Given one single front-view image, the ego vehicle's current state, and a navigation command (\textit{e.g.}, go straight, turn left or turn right), the model is required to analyze the environment, perform reasoning, and finally generate a 4-second future trajectory.

To assess the trajectory planning capability of \XFM{}, we provide quantitative comparisons in Table~\ref{tab:planning_comparison}, where \XFM{} is evaluated on the NAVSIM benchmark~\cite{dauner2024navsim} against InternVL3-8B~\cite{zhu2025internvl3} and its derivative ReCogDrive~\cite{recogdrive} planner of comparable scale. To ensure fair comparison with counterparts, we adopt the denoising policy from ReCogDrive~\cite{recogdrive} during the imitation IL stage, where noisy trajectories are alternately fused with VLM hidden states, ego states, and historical trajectories. During the RL stage, we employ DiffGRPO~\cite{recogdrive} to further enhance the model’s performance. Here, NC, DAC, TTC, Conf, EP denote No At-Fault Collision, Drivable Area Compliance, Time-to-Collision, Comfort and Ego Process, respectively. The PDMS (Predictive Driver Model Score) metric is the weighted combination of above sub-scores, with detailed computation procedures available in~\cite{dauner2024navsim}. All models take a single front-view image as input and are trained for the same number of epochs. \XFM{} consistently outperforms the competing models. Notably, unlike InternVL3~\cite{zhu2025internvl3} that discretizes a single-frame image into patches, \XFM{} employs 3D convolutions to significantly reduce the number of LLM tokens while preserving detailed spatial context.

Furthermore, we showcase a set of representative driving scenarios. As illustrated in Figure~\ref{figure_planning_navisim}, the qualitative results demonstrate that \XFM{} can handle diverse autonomous driving situations and accomplish challenging tasks, including intersection turning, U-turning on curved roads, car-following and lane-change overtaking. In each case, the model is expected to perceive the road context, integrate the ego status and navigation intent, and produce a coherent decision.

\begin{figure*}[!t]
\centering
  \includegraphics[width=0.94\textwidth]{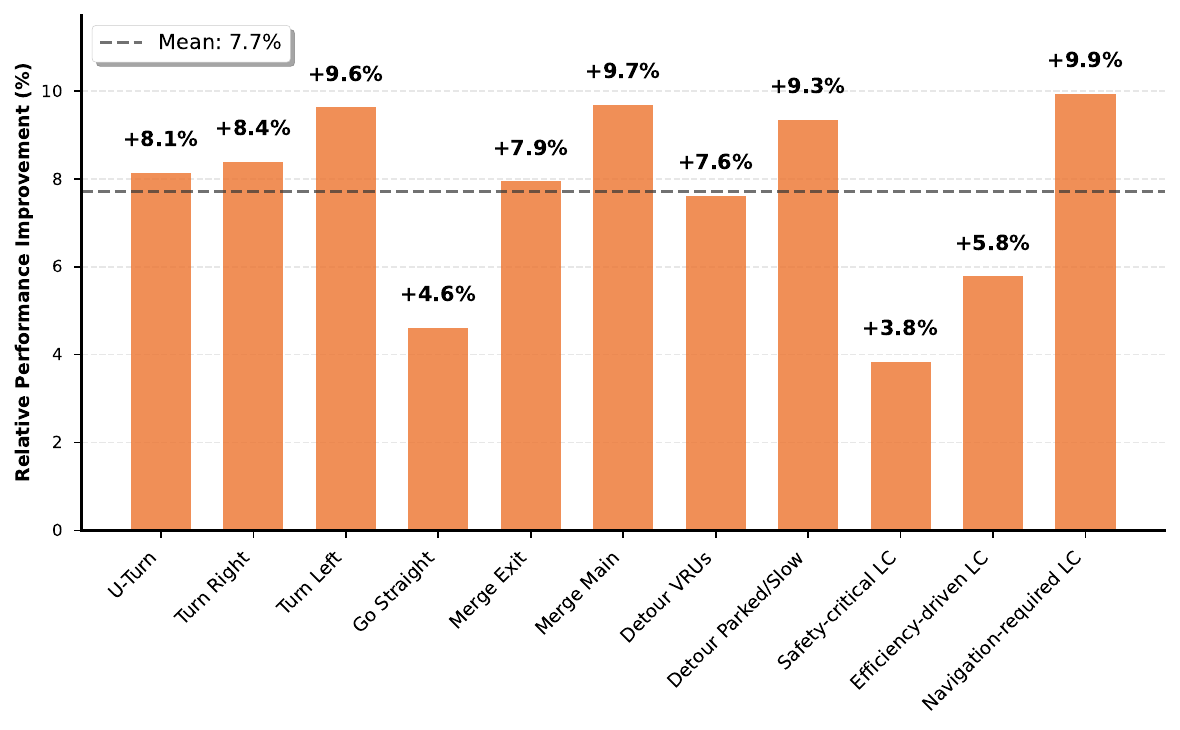}
  \caption{Relative performance improvement of \XFM{} 7B over baseline method Qwen2.5-VL 7B~\cite{bai2025qwen2} of trajectory planning on the proprietary dataset. Improvement is quantified as the percentage reduction in error metric, with higher bars indicating greater error reduction and better performance.}
  \label{figure_planning}
  \vspace{-1em}
\end{figure*}

\textbf{Trajectory Planning on Proprietary Data} 
To validate the efficacy of our training strategy in practical autonomous driving contexts, we evaluate the model's trajectory planning performance on a proprietary, large-scale dataset. This dataset, gathered from diverse geographic locations under a wide spectrum of traffic conditions, features a comprehensive set of driving scenarios. Beyond standard maneuvers such as straight-line driving and turns, it is enriched with challenging, safety-critical situations that involving detouring around vulnerable road users (VRUs), overtaking or bypassing slow and parked vehicles, and other complex interactions.
The demonstration trajectories are sourced from expert human drivers, selected for their exemplary safety records and ability to perform smooth, comfortable maneuvers. These drivers exhibit strict adherence to traffic regulations and proficient defensive driving skills, ensuring the dataset provides high-quality, near-optimal driving exemplars for imitation learning. The entire evaluation set is meticulously annotated with precise trajectories and scenario labels.

In our experimental setup, we predict ego vehicle's 3-second trajectory based on five consecutive front-view camera frames sampled at 2 Hz, with predictions represented in the ego-vehicle's coordinate.
To ensure a fair and rigorous comparison, our proposed \XFM{} and the Qwen2.5-VL~\cite{bai2025qwen2} baseline are evaluated under identical settings, including the input representation, full fine-tuning training strategy and the optimization procedure. We assess performance in an open-loop evaluation protocol, using the L2 distance~\cite{hu2022st} over the 3-second prediction horizon as the primary metric. As illustrated in Figure~\ref{figure_planning}, which depicts the relative performance over the baseline, \XFM{} consistently outperforms the baseline across all evaluated categories. Notably, the performance gains are most pronounced in complex, interactive maneuvers such as turns, nudges around obstacles, and lane changes. These scenarios are inherently challenging as they require anticipating the behavior of other road users and precise vehicle control. This substantial improvement, particularly in high-complexity scenarios, provides strong evidence that our embodied training paradigm equips the model with a superior capacity for reasoning in complex driving situations. It also translates to better generalization across unseen road conditions with universal driving logic. Consequently, the model generates trajectories that are more accurate and more aligned with expert human driving behavior.

\subsection{Ablation Study}

To verify the effectiveness of our proposed multi-stage training strategy, we conducted systematic ablation experiments, with the results presented in Table~\ref{tab:ablation_study}. The experimental results show that the model trained using only embodied data (MiMo-VL w/ Embodied) achieves strong performance in both domains, whereas the model trained solely on autonomous driving data (MiMo-VL w/ AD) excels in its own domain but suffers from a significant performance drop on embodied tasks. This indicates that simple single-task training is insufficient for cross-domain generalization. Under the experimental condition of ensuring approximately the same number of loss tokens, directly training embodied and autonomous driving tasks together (MiMo-VL w/ Embodied$+$AD) achieves improvements on embodied tasks, but the performance of autonomous driving declines slightly. In contrast, our proposed \XFM{} employs a multi-stage training strategy, effectively mitigating cross-domain task interference through progressive course learning. It achieves an average performance of 62.4\% on embodied tasks (4\% improvement compared to Embodied$+$AD) and a best performance of 63.3\% on autonomous driving tasks (8.1\% improvement compared to Embodied$+$AD). This fully demonstrates that our multi-stage training strategy can achieve synergistic improvement of embodied intelligence and autonomous driving capabilities without sacrificing the performance of a single task, providing an effective training paradigm for building a unified embodied foundation model.
\begin{table*}[!t]
  \small
  \centering
  \resizebox{\textwidth}{!}{ 
  \renewcommand{\arraystretch}{1.3} 
  \setlength{\tabcolsep}{6pt}       
  \begin{tabular}{lcccccccc}
    \toprule
    \rowcolor[HTML]{FFE0CC}  
    \textbf{Model} & \textbf{Embodied} & \textbf{AD} & \textbf{Multi-Stage} & \textbf{Affordance}$\uparrow$ & \textbf{Spatial}$\uparrow$ & \textbf{Plan}$\uparrow$ & \textbf{Embodied Avg.}$\uparrow$ & \textbf{Autonomous Driving}$\uparrow$ \\
    \midrule
    MiMo-VL (Baseline) & \XSolidBrush & \XSolidBrush & \XSolidBrush & 38.7 & 55.3 & 40.1& 47.7& 32.7\\
    MiMo-VL w/ Embodied & \Checkmark & \XSolidBrush & \XSolidBrush & 58.9 & 61.0 & 44.6& 57.4& 32.3\\
    MiMo-VL w/ AD & \XSolidBrush & \Checkmark & \XSolidBrush & 26.3 & 56.3 & 39.3& 44.5& 57.4\\
    MiMo-VL w/ Embodied$+$AD & \Checkmark & \Checkmark & \XSolidBrush & 59.6 & 62.0 & 50.5& 59.2& 55.2 \\
    \midrule
    \rowcolor[HTML]{FFE0CC}\textbf{\XFM{} (Ours)} & \textbf{\Checkmark} & \textbf{\Checkmark} & \textbf{\Checkmark} & \textbf{65.6} & \textbf{66.0} & \textbf{53.9}& \textbf{63.7}& \textbf{63.4}\\
    \bottomrule
  \end{tabular}
  } 
  \caption{Ablation study of different model configurations. The \Checkmark indicates the feature is enabled, while the \XSolidBrush indicates it is disabled. Multi-stage denotes our proposed multi-stage training strategy. The best results are highlighted in \textbf{bold}.}
  \label{tab:ablation_study}
\end{table*}
\section{Conclusion and Next Steps}
\label{sec:conclusion}
This report introduces \XFM{}, a pioneering cross-embodied vision-language model that achieves state-of-the-art performance in both autonomous driving and embodied AI tasks. 
As the first open-source VLM integrating these two critical domains, \XFM{} significantly enhances understanding and reasoning in dynamic physical environments.
Extensive evaluations across 29 benchmarks show that \XFM{} achieves
superior performance in both embodied and autonomous driving tasks, significantly outperforming existing open-source and closed-source general VLMs, as well as specialized VLMs for a single domain.
The report provides a comprehensive overview of the design, data construction, training methodologies, and practical applications of \XFM{}, aiming to inspire future research in the field.

Building on the capabilities of our \XFM{} model, we will explore Embodied AI Vision-Language Action (VLA) Models to enhance interaction in complex environments, enabling more intuitive task execution through natural language understanding. Additionally, we aim to develop Autonomous Driving VLA Models that allow these systems to interpret real-time traffic conditions and respond effectively to dynamic driving scenarios. Furthermore, we will investigate the integration of multi-modal learning, such as utilizing 3D point clouds, to improve long-term planning and decision-making capabilities, as well as advance human-robot interaction. These areas are crucial for enhancing overall performance and driving innovation in both embodied AI and autonomous driving technologies.

\clearpage

\bibliographystyle{plainnat}
\bibliography{main}

\clearpage
\section{Contributions and Acknowledgments}
\label{sec:contributions}
\setlength{\parskip}{0pt} 
\setlength{\itemsep}{0pt} 
\setlength{\parsep}{0pt}  
\begin{multicols}{-2}
\raggedcolumns

\subsubsection*{Core Contributors}
    \begin{itemize}
        \item Xiaoshuai Hao
        \item Lei Zhou
        \item Zhijian Huang
        \item Zhiwen Hou
        \item Yingbo Tang
        \item Lingfeng Zhang
        \item Guang Li
        \item Zheng Lu 
        \item Shuhuai Ren
        \item Fuli Luo
        \item Hangjun Ye
        \item Long Chen$^\dagger$

    \end{itemize}

    \columnbreak 
    \subsubsection*{Contributors}
    \begin{itemize}
        \item Xianhui Meng
        \item Yuchen Zhang
        \item Jing Wu
        \item Jinghui Lu
        \item Chenxu Dang
        \item Jiayi Guan
        \item Jianhua Wu
        \item Zhiyi Hou
        \item Hanbing Li
        \item Shumeng Xia
        \item Mingliang Zhou
        \item Yinan Zheng
        \item Zihao Yue
        \item Shuhao Gu
        \item Hao Tian
        \item Yuannan Shen
        \item Jianwei Cui
        \item Wen Zhang
        \item Shaoqing Xu
        \item Bing Wang
        \item Haiyang Sun
        \item Zeyu Zhu
        \item Yuncheng Jiang
        \item Zibin Guo
        \item Chuhong Gong
        \item Chaofan Zhang
        \item Wenbo Ding
        \item Kun Ma
        \item Guang Chen
        \item Rui Cai
        \item Diyun Xiang
        \item Heng Qu

    \end{itemize}
\end{multicols}

{\renewcommand{\thefootnote}{\fnsymbol{footnote}}\footnotetext[2]{Project Leader}}

\subsubsection*{Acknowledgments}
We would like to sincerely thank for the tremendous support from the broader team, including those not listed above: Naiyan Wang, Yongkang Li, Chitian Sun, Lin Liu, Feiyang Jia, Jie Wang, Haochen Tian, Yihang Qiu, Junli Wang, Yinfeng Gao. 

\clearpage
\beginappendix

\section{Appendix}

\subsection{General Visual Understanding}

To verify that \XFM{}'s specialized training does not come at the cost of general proficiency, we evaluate it across a range of general visual understanding benchmarks including MMMU-Pro (Standard and Vision)~\cite{yue2025mmmu}, Mantis~\cite{jiang2024mantis}, AI2D~\cite{kembhavi2016diagram}, V*~\cite{wu2024v}, VLMs are Blind~\cite{rahmanzadehgervi2024vision}, PixmoCount~\cite{deitke2024molmo}, and CountBench~\cite{paiss2023teaching}. The evaluation metrics for all benchmarks are based on accuracy. We compare \XFM{} with its base model, MiMo-VL~\cite{coreteam2025mimovltechnicalreport}, as well as top-tier proprietary and open-source models including Qwen2.5-VL~\cite{bai2025qwen2}, InternVL3~\cite{zhu2025internvl3}, GPT-4o~\cite{gpt4o}, and Claude 3.7 Sonnet~\cite{anthropic20253}).

\begin{table}[!h]
\centering
\setlength{\tabcolsep}{3pt} 
\renewcommand{\arraystretch}{1.25}
\resizebox{\linewidth}{!}{%
\begin{tabular}{lc|*{8}{c}}
\toprule
\multicolumn{2}{c}{\textbf{Model Info}} & \multicolumn{8}{c}{\textbf{General Visual Understanding}} \tabularnewline
\cmidrule(lr){1-2}\cmidrule(lr){3-10}
\multicolumn{1}{c}{\textbf{Names}} & \multicolumn{1}{c}{\textbf{Params}} &
\multicolumn{1}{c}{\textbf{\makecell{MMMU-\\Pro\textsubscript{standard}}}} & \multicolumn{1}{c}{\textbf{\makecell{MMMU-\\Pro\textsubscript{vision}}}} & \multicolumn{1}{c}{\textbf{{\ }{\ }Mantis{\ }{\ }}} &
\multicolumn{1}{c}{\textbf{{\ }{\ }{\ }AI2D{\ }{\ }{\ }}} & \multicolumn{1}{c}{\textbf{{\ }{\ }{\ }{\ }V*{\ }{\ }{\ }{\ }}} & \multicolumn{1}{c}{\textbf{\makecell{VLMs \\are Blind}}} & \multicolumn{1}{c}{\textbf{PixmoCount}} & \multicolumn{1}{c}{\textbf{CountBench}} \tabularnewline
\midrule
\rowcolor[HTML]{FFE0CC}
\multicolumn{10}{c}{\textit{Open-Source Models}} \tabularnewline
\midrule
MiMo-VL~\cite{coreteam2025mimovltechnicalreport} & 7B & 42.37 & 35.95 & \textbf{81.57} & 81.83 & \underline{81.68} & \textbf{74.91} & \underline{73.35} & 85.34 \tabularnewline
Qwen2.5-VL~\cite{bai2025qwen2} & 7B & 34.70 & 29.40 & 74.70 & 83.90 & 73.80 & 37.40 & 60.70 & 74.10 \tabularnewline
InternVL3~\cite{zhu2025internvl3} & 8B & 45.60 & 37.80 & 72.80 & \textbf{85.20} & 72.80 & 36.80 & 62.00 & 80.00 \tabularnewline
\midrule
\rowcolor[HTML]{FFE0CC}
\multicolumn{10}{c}{\textit{Closed-Source (Proprietary) Models}} \tabularnewline
\midrule
GPT-4o~\cite{gpt4o} & {--} & 42.50 & 36.10 & 75.60 & 82.60 & 73.90 & 49.80 & 54.40 & 85.70 \tabularnewline
Claude 3.7 Sonnet~\cite{anthropic20253} & {--} & \textbf{56.50} & \underline{45.80} & 75.10 & 81.40 & {--} & 72.10 & 53.50 & \textbf{90.20} \tabularnewline
\midrule
\rowcolor[HTML]{FFE0CC}
\textbf{\XFM{} (Ours)} & 7B & {\underline{52.08}} & {\textbf{45.84}} & \underline{81.10} & \underline{84.20} & {\textbf{82.72}} & \underline{72.32} & {\textbf{77.50}} & \underline{87.37} \tabularnewline
\bottomrule
\end{tabular}
}
\vspace{1ex}
\caption{\textbf{Comparison of \XFM{} with other models on general visual understanding benchmarks.} 
Results for Qwen2.5-VL~\cite{bai2025qwen2}, InternVL3~\cite{zhu2025internvl3}, GPT-4o~\cite{gpt4o}, and Claude 3.7 Sonnet~\cite{anthropic20253} are sourced from the MiMo-VL~\cite{coreteam2025mimovltechnicalreport} technical report for a consistent comparison. The best results among the listed models are \textbf{bolded} and the second-best is \underline{underlined}.}
\label{tab:general_comparison}
\end{table}

The comprehensive results in Table~\ref{tab:general_comparison} reveal that after multi-stage embodied and autonomous driving fine-tuning, \XFM{}'s general capabilities remain intact. \XFM{} preserves the proficiency of its base model MiMo-VL~\cite{coreteam2025mimovltechnicalreport} while achieving notable improvements across some benchmarks. On complex reasoning tasks, \XFM{} achieves a 9.71 and 9.88 points absolute gain on MMMU-Pro\textsubscript{standard} and MMMU-Pro\textsubscript{vision} benchmarks~\cite{yue2025mmmu}, respectively. For visual question answering, \XFM{} delivers competitive results with 81.10\% on Mantis~\cite{jiang2024mantis} and 84.20\% on AI2D~\cite{kembhavi2016diagram}. It also achieves the best performance on V*~\cite{wu2024v} benchmark. On general counting benchmarks, \XFM{}{} excels MiMo-VL~\cite{coreteam2025mimovltechnicalreport} with an improvement of 4.15 points on PixmoCount~\cite{deitke2024molmo} and 2.03 points on CountBench~\cite{paiss2023teaching}.

The improvements on general visual understanding benchmarks indicate that the specialized training for \XFM{} in embodied AI and autonomous driving yields benefits not only for target domains but also for fundamental visual comprehension. For instance, the gains in complex reasoning and counting likely stem from enhanced fine-grained object recognition and structural understanding, while spatial benchmarks benefit from improved dynamic relationship modeling and 3D spatial reasoning. This confirms that \XFM{}'s specialized training not only preserves but amplifies its general vision-language proficiency.

\subsection{Embodied Visualization Examples}
\label{subsec:appendix_embodied}

\begin{figure}[H]
    \subsubsection{Spatial Understanding}
    \label{sssec:spatial_understanding}
    \centering
    \includegraphics[width=1.0\linewidth,height=21cm]{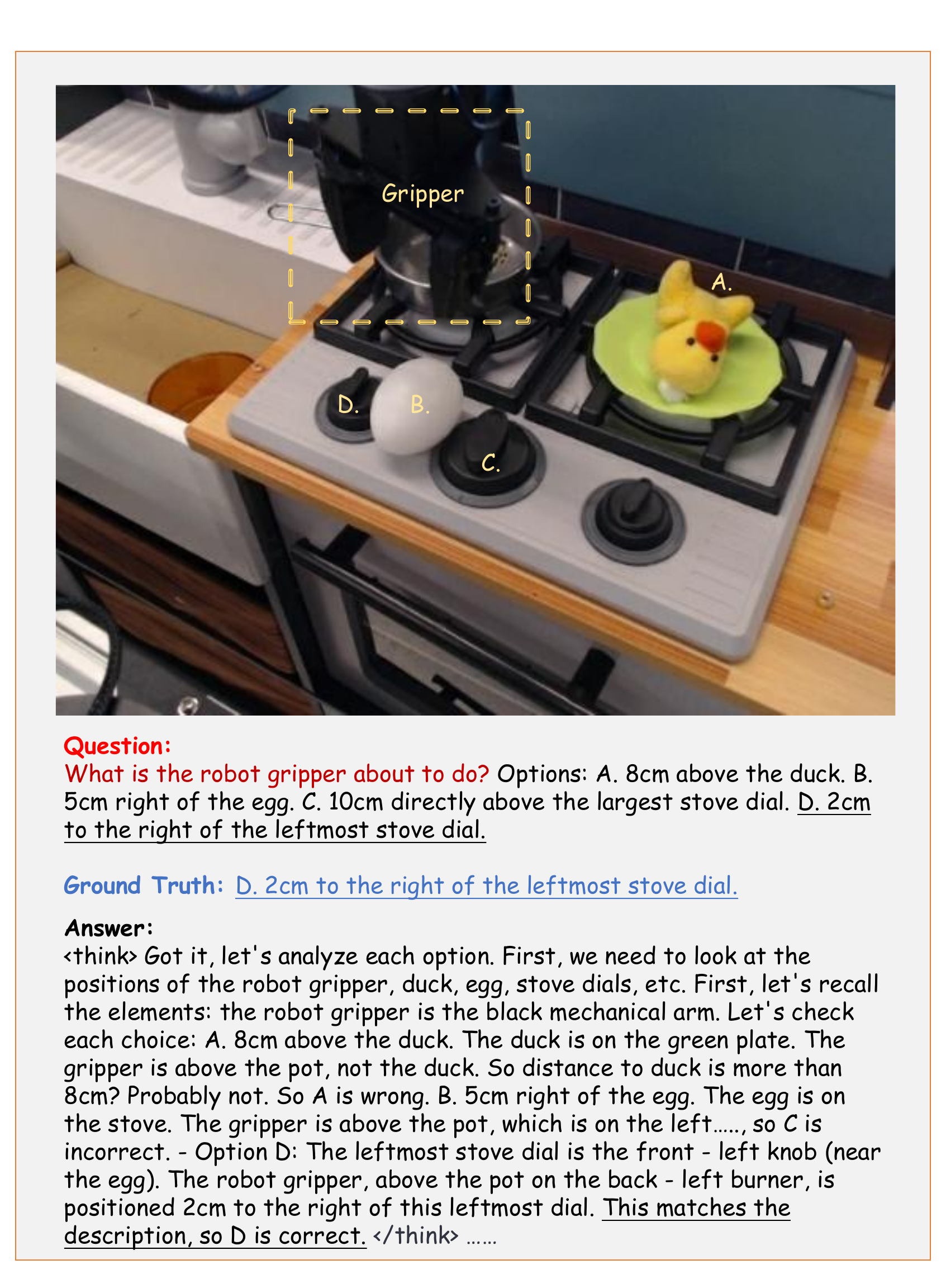}
    \caption{Embodied spatial understanding example 1.}
    \label{fig:spatial-1}
\end{figure}

\begin{figure}[H]
    \centering
    \includegraphics[width=1.0\linewidth]{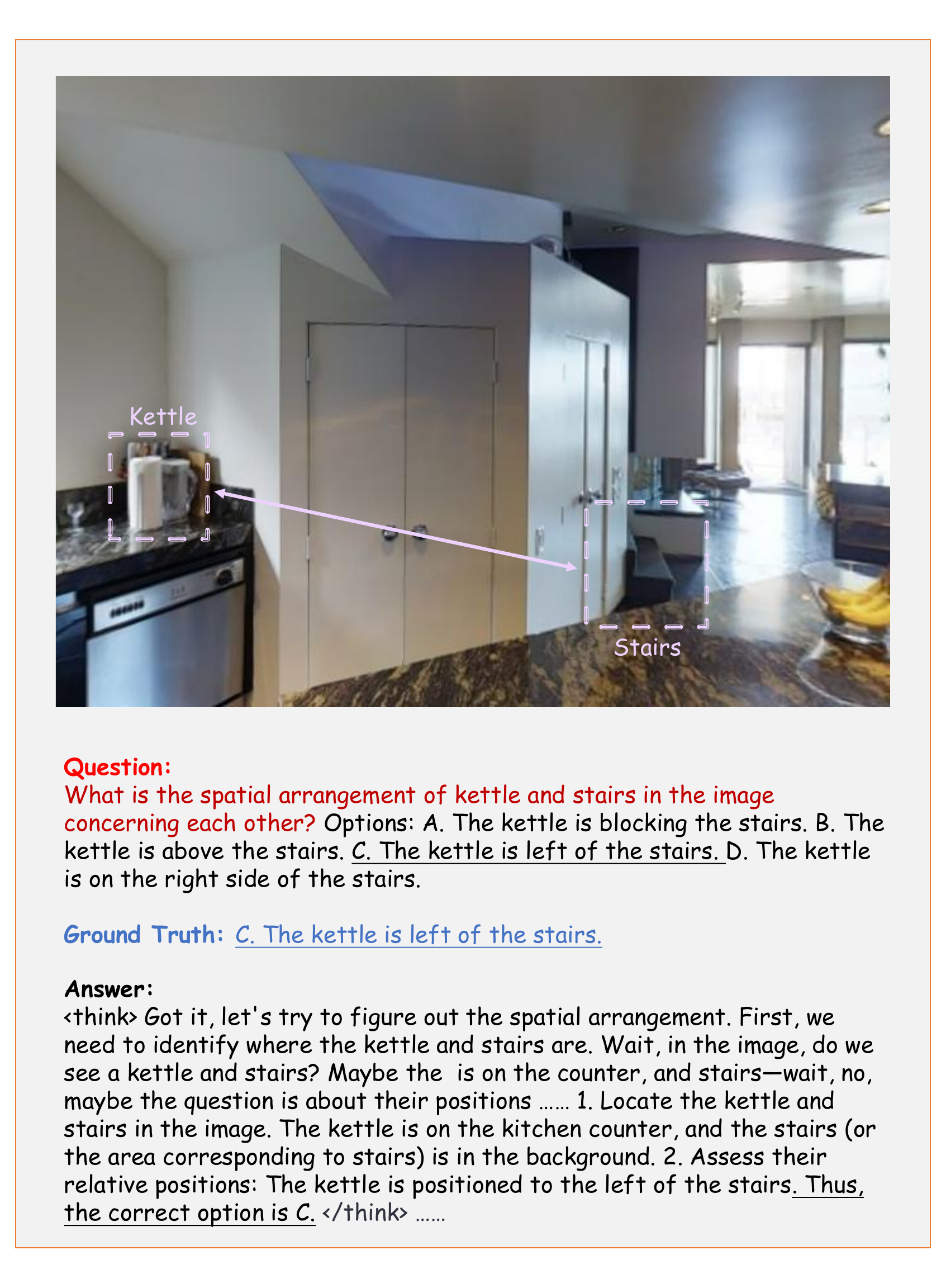}
    \caption{Embodied spatial understanding example 2.}
    \label{fig:spatial-2}
\end{figure}

\begin{figure}[H]
    \centering
    \includegraphics[width=1.0\linewidth]{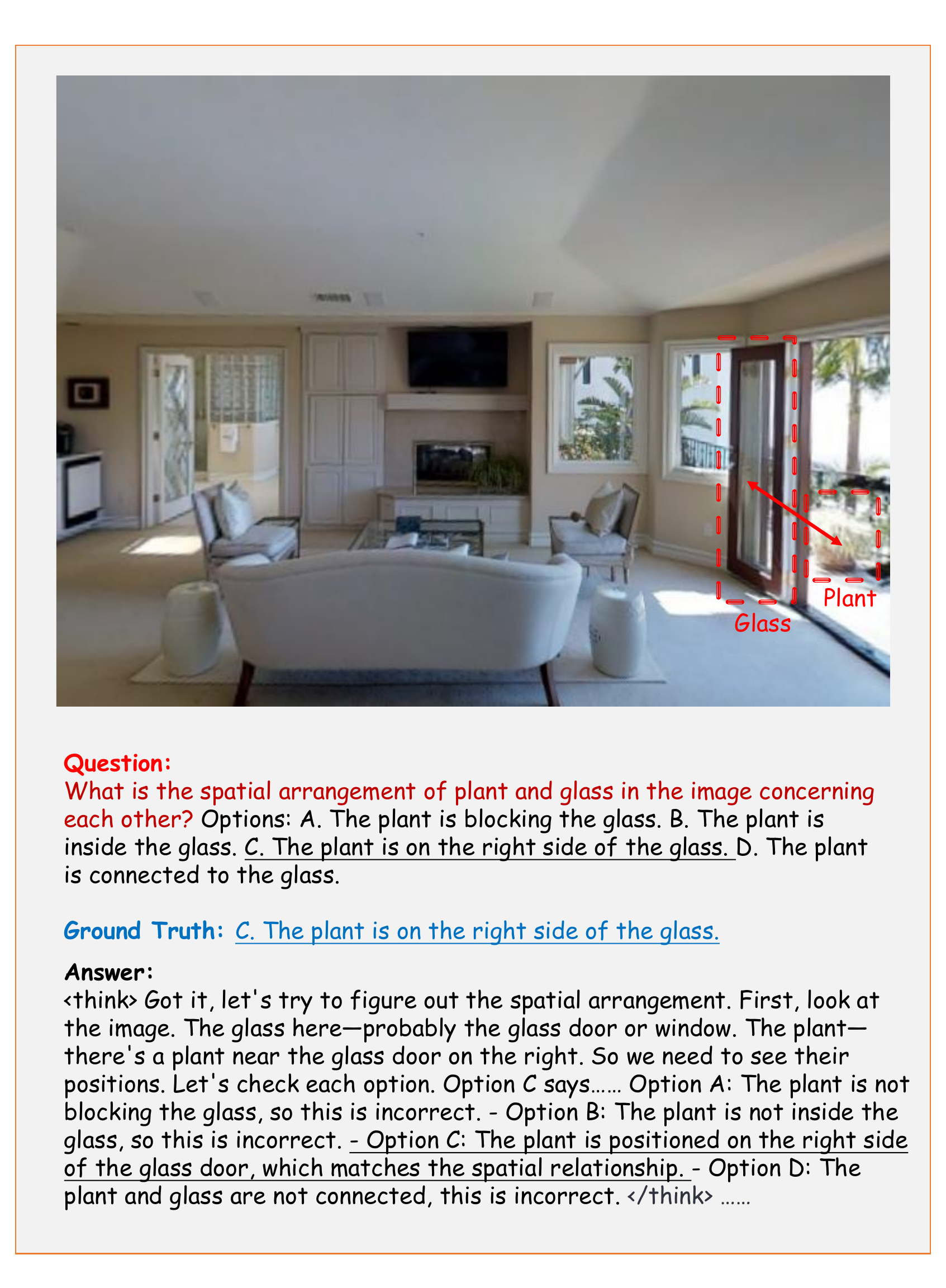}
    \caption{Embodied spatial understanding example 3.}
    \label{fig:spatial-3}
\end{figure}

\begin{figure}[H]
    \centering
    \includegraphics[width=1.0\linewidth]{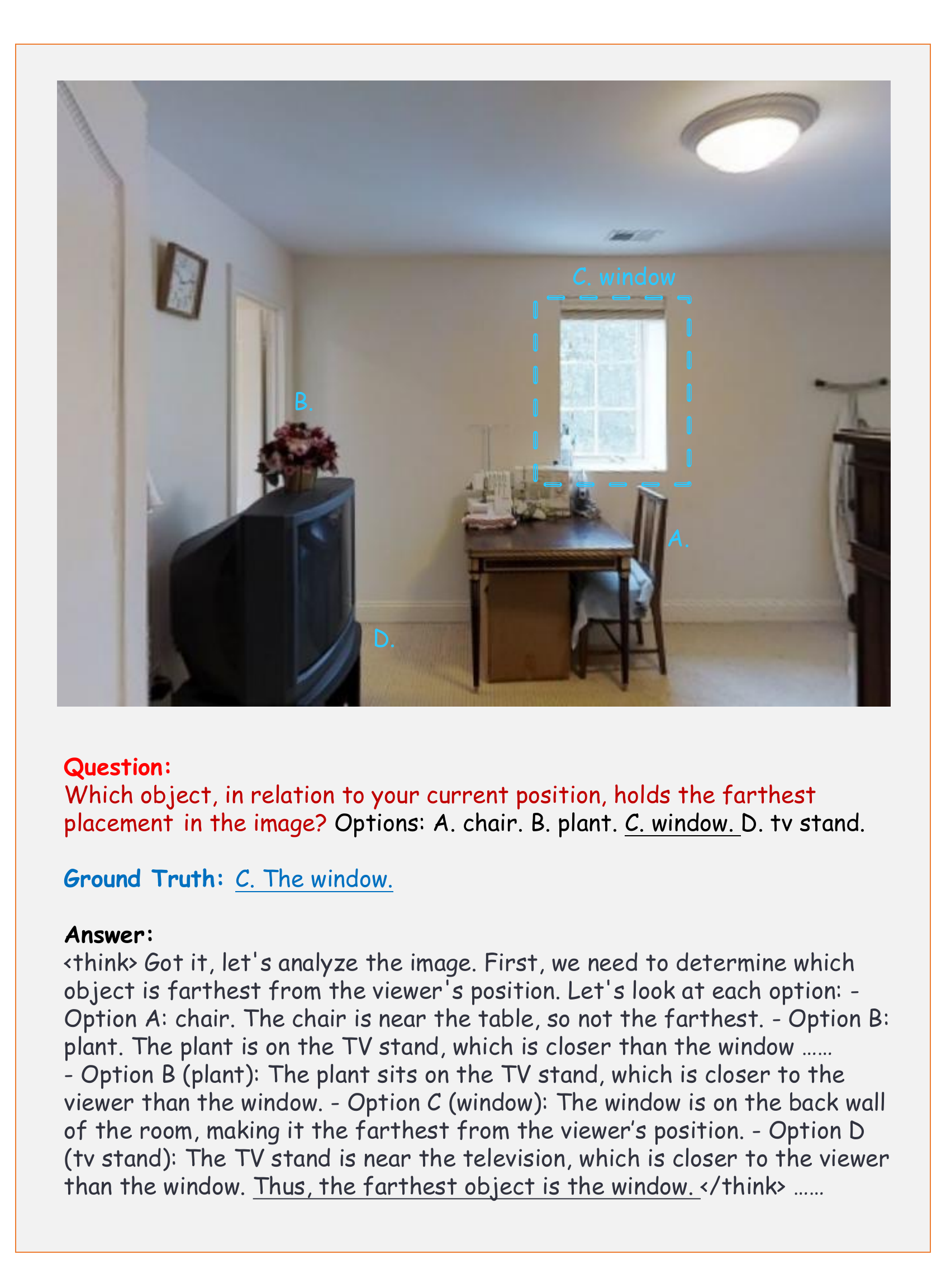}
    \caption{Embodied spatial understanding example 4.}
    \label{fig:spatial-4}
\end{figure}

\begin{figure}[H]
    \centering
    \includegraphics[width=1.0\linewidth]{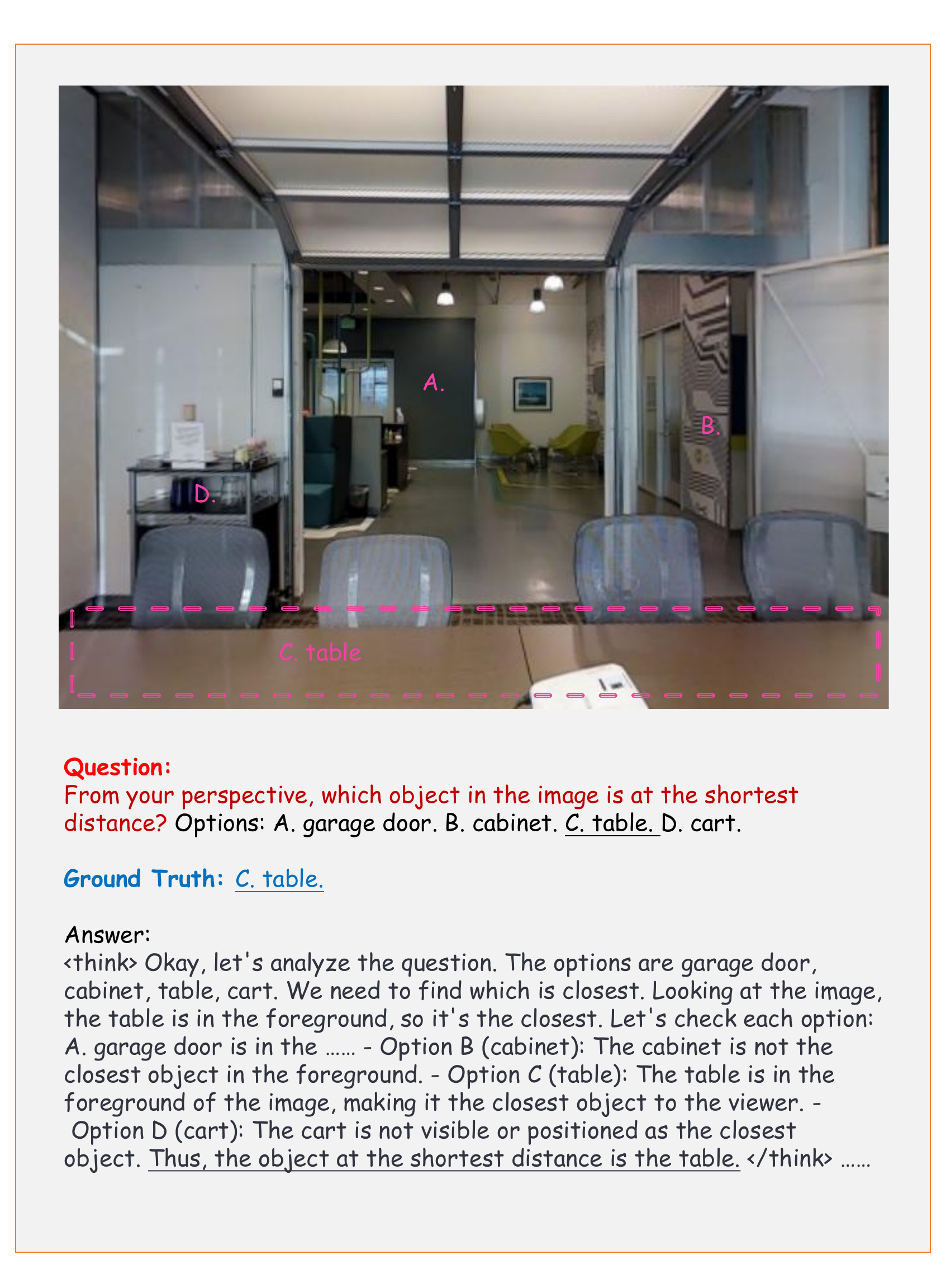}
    \caption{Embodied spatial understanding example 5.}
    \label{fig:spatial-5}
\end{figure}

\begin{figure}[H]
    \centering
    \includegraphics[width=1.0\linewidth]{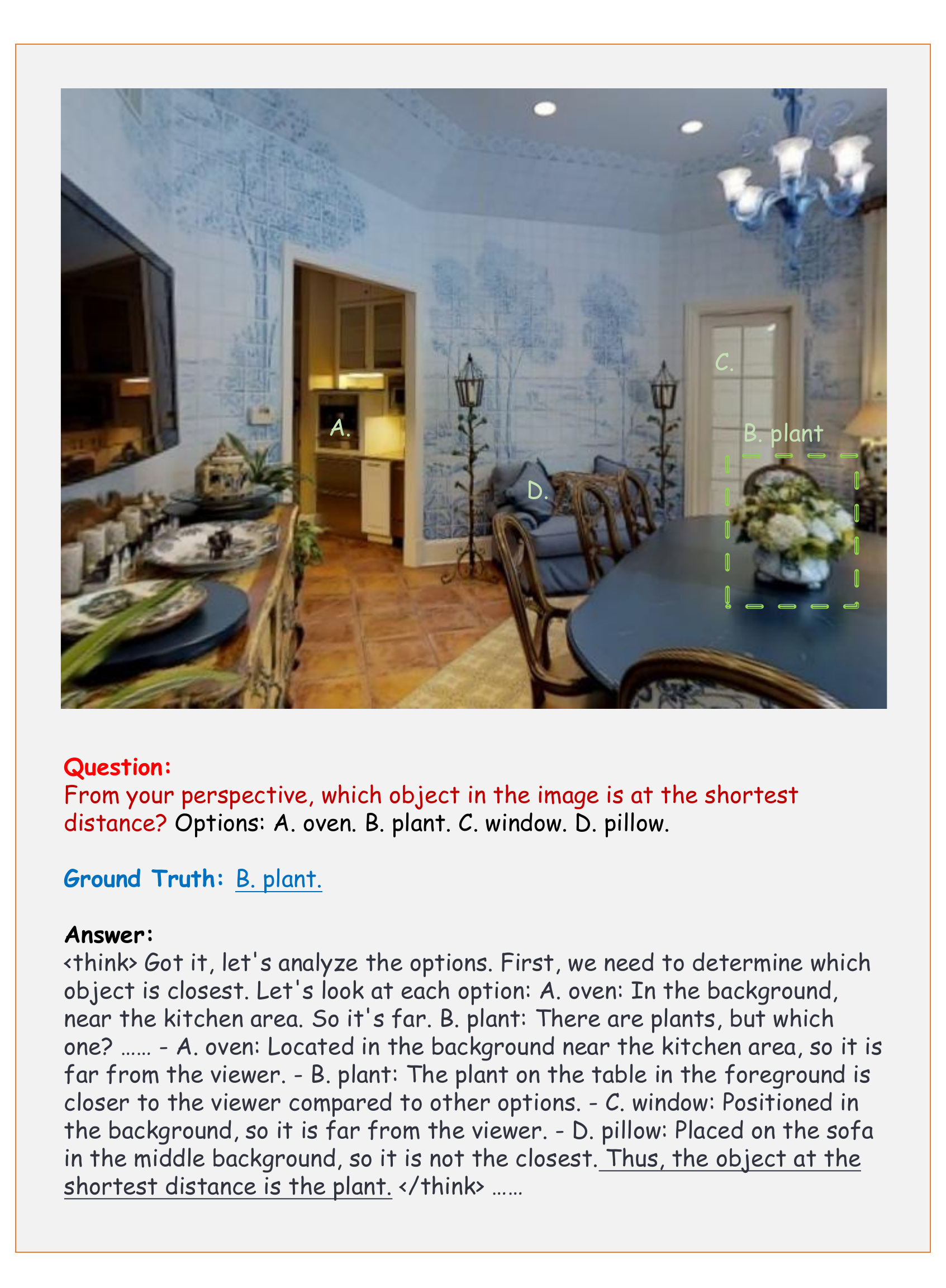}
    \caption{Embodied spatial understanding example 6.}
    \label{fig:spatial-6}
\end{figure}

\begin{figure}[H]
    \centering
    \includegraphics[width=1.0\linewidth]{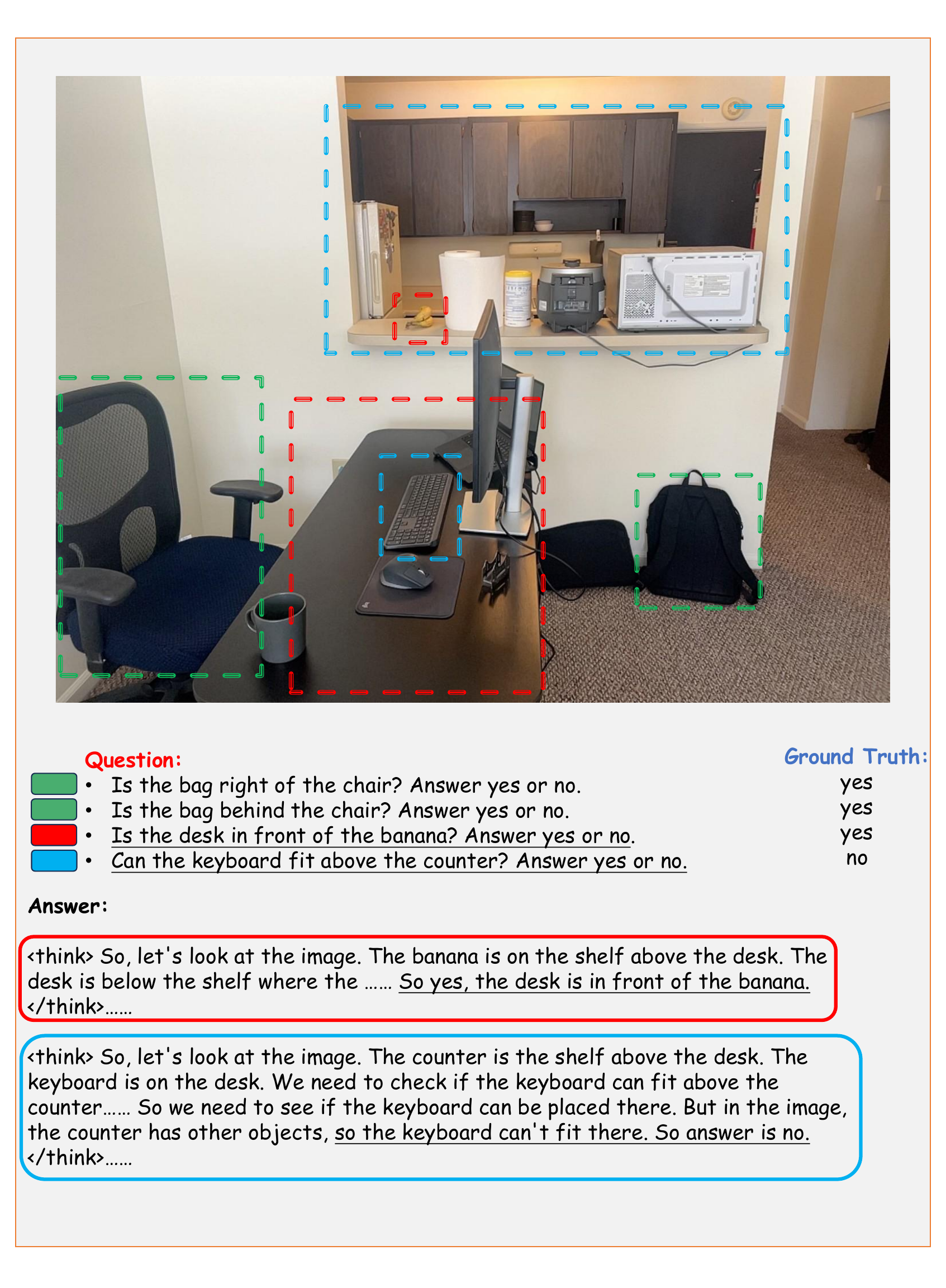}
    \caption{Embodied spatial understanding example 7.}
    \label{fig:spatial-7}
\end{figure}

\begin{figure}[H]
    \centering
    \includegraphics[width=1.0\linewidth]{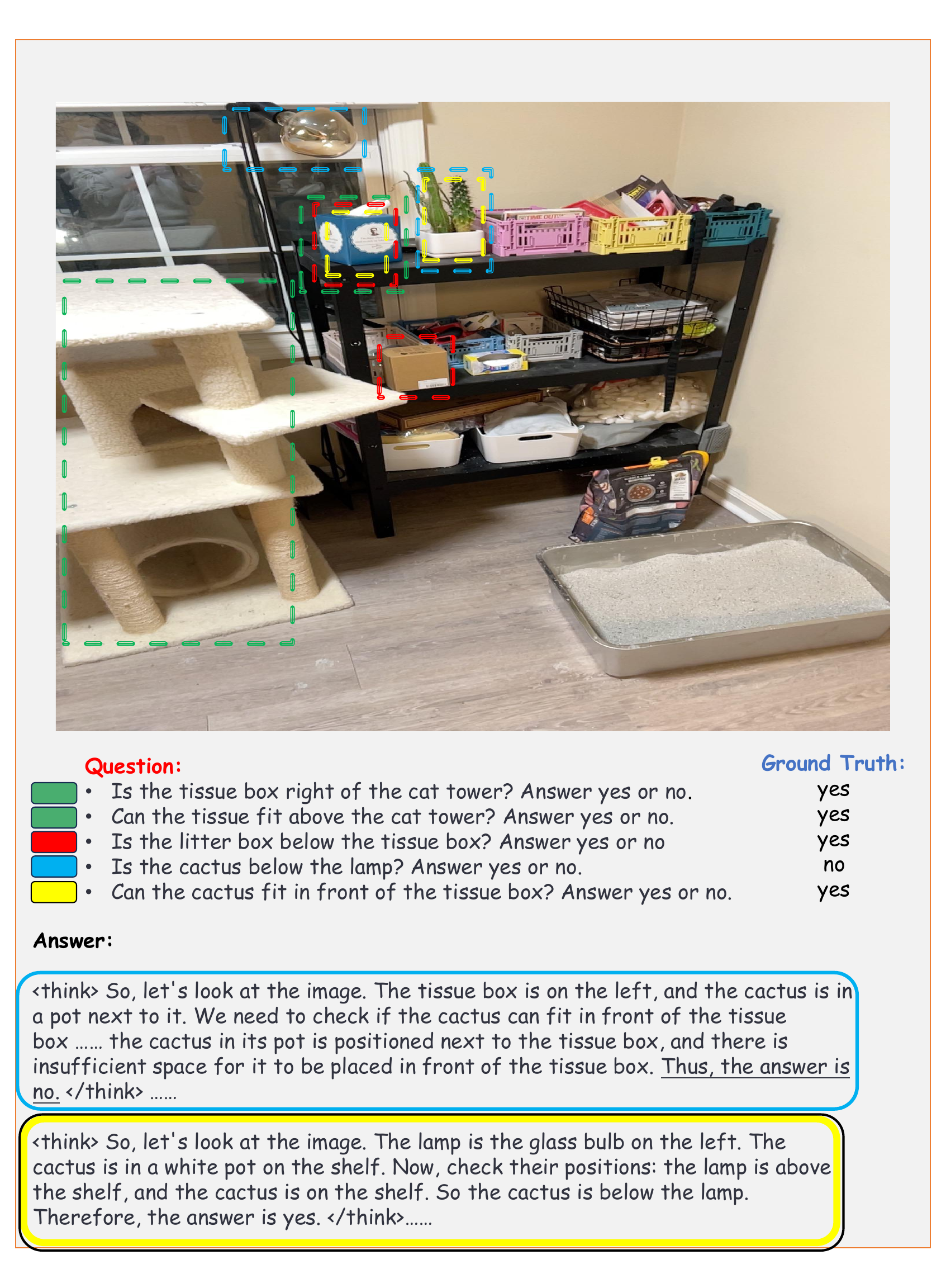}
    \caption{Embodied spatial understanding example 8.}
    \label{fig:spatial-8}
\end{figure}

\begin{figure}[H]
    \subsubsection{Affordance Prediction}
    \label{sssec:affordance_prediction}
    \centering
    \includegraphics[height=21.5cm]{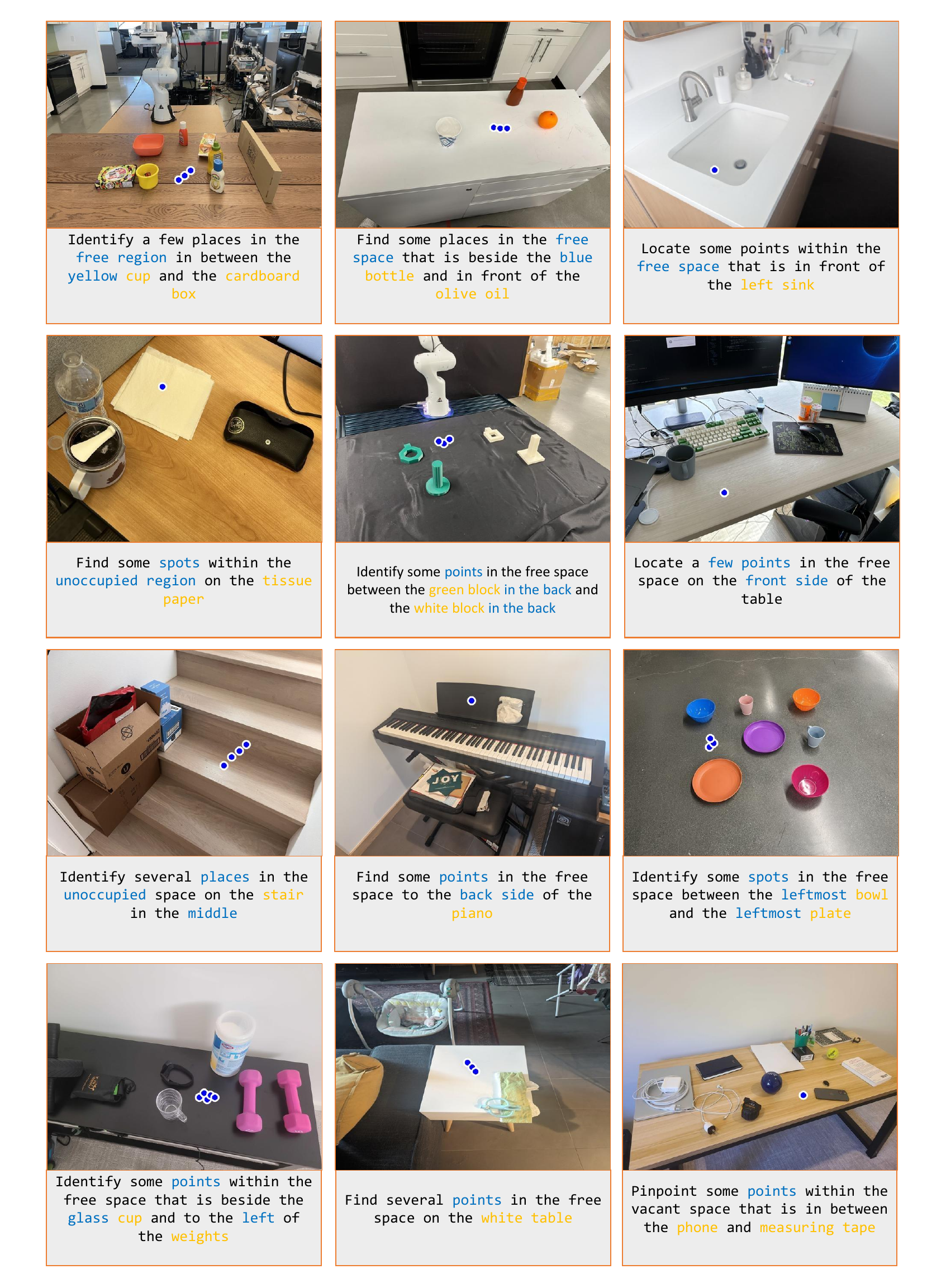}
    \caption{Embodied affordance prediction example 1.}
    \label{fig:afford-1}
\end{figure}

\begin{figure}[H]
    \centering
    \includegraphics[width=1.0\linewidth, height=22cm]{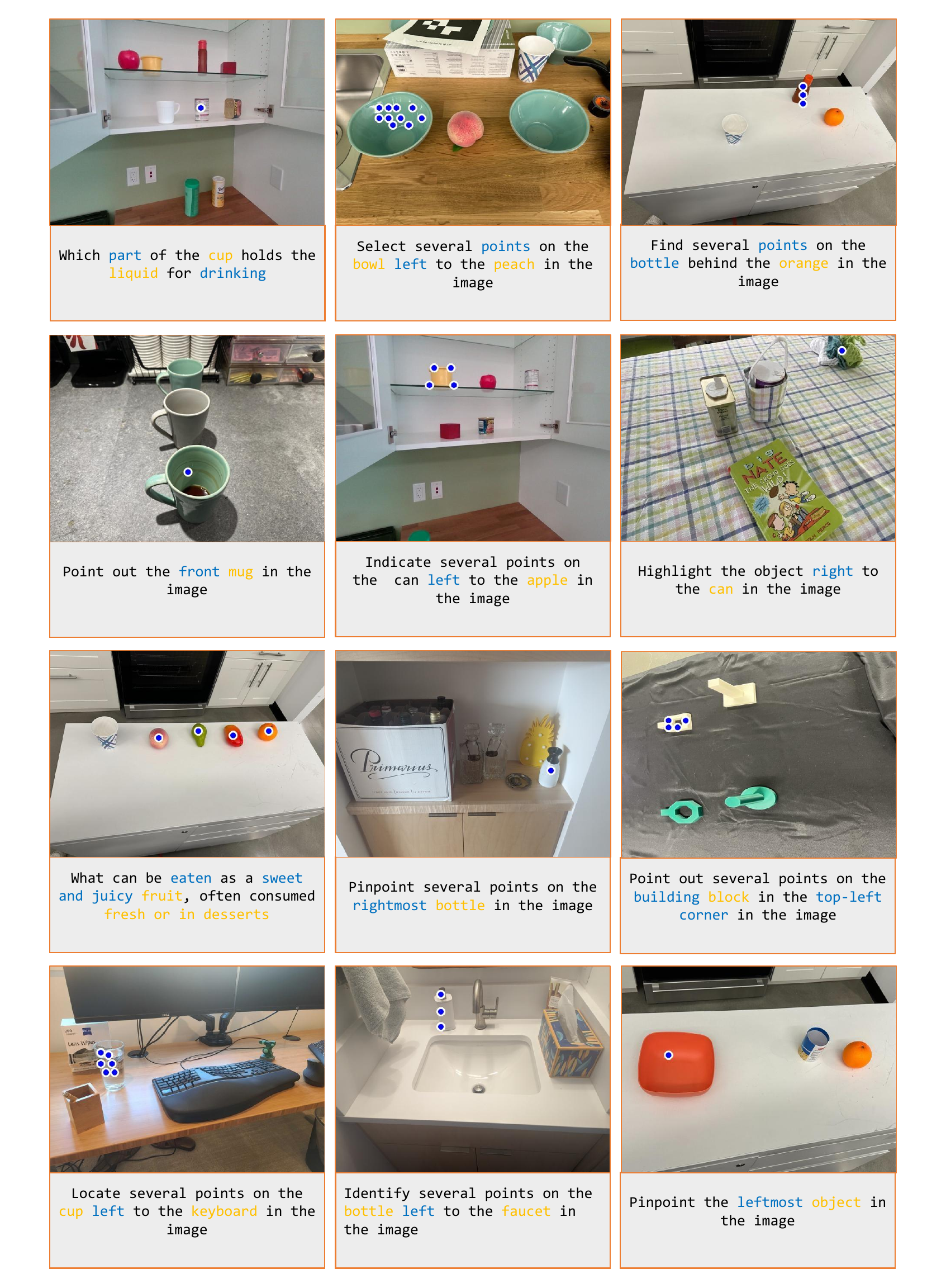}
    \caption{Embodied affordance prediction example 2.}
    \label{fig:afford-2}
\end{figure}

\begin{figure}[H]
    \centering
    \includegraphics[width=1.0\linewidth,height=22cm]{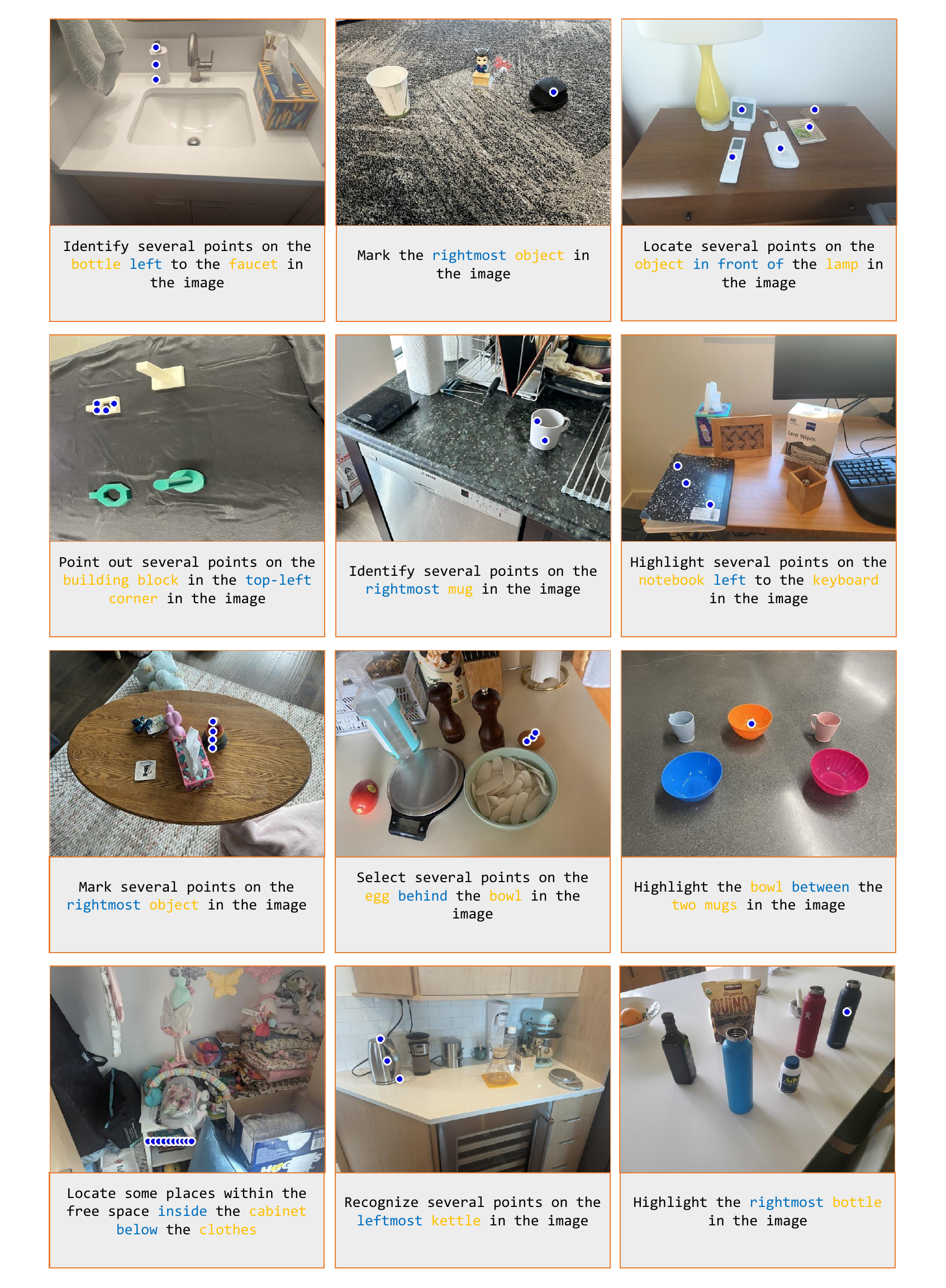}
    \caption{Embodied affordance prediction example 3.}
    \label{fig:afford-3}
\end{figure}

\begin{figure}[H]
    \centering
    \includegraphics[width=1.0\linewidth,height=22cm]{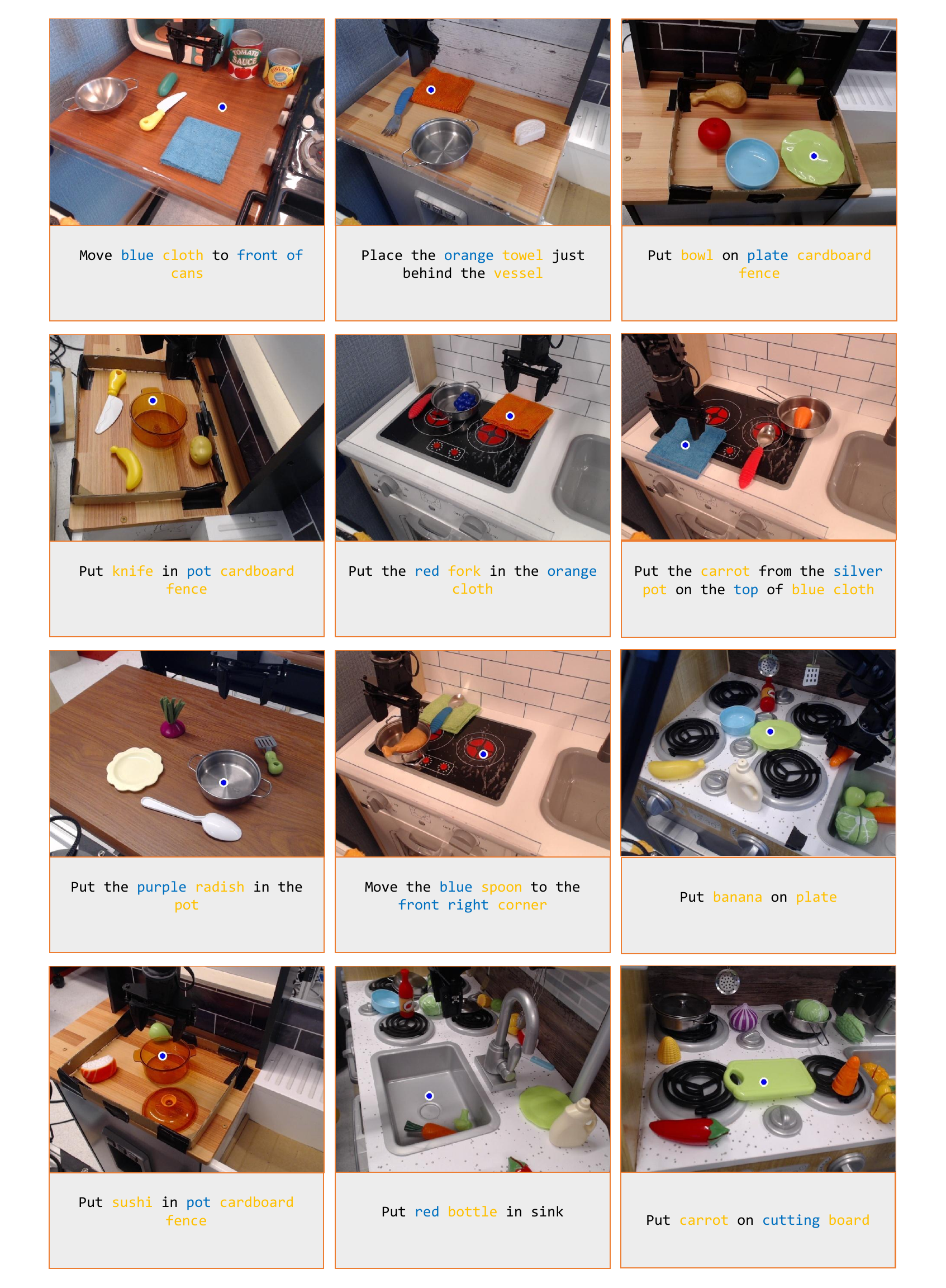}
    \caption{Embodied affordance prediction example 4.}
    \label{fig:afford-4}
\end{figure}

\begin{figure}[H]
    \centering
    \includegraphics[width=1.0\linewidth,height=22cm]{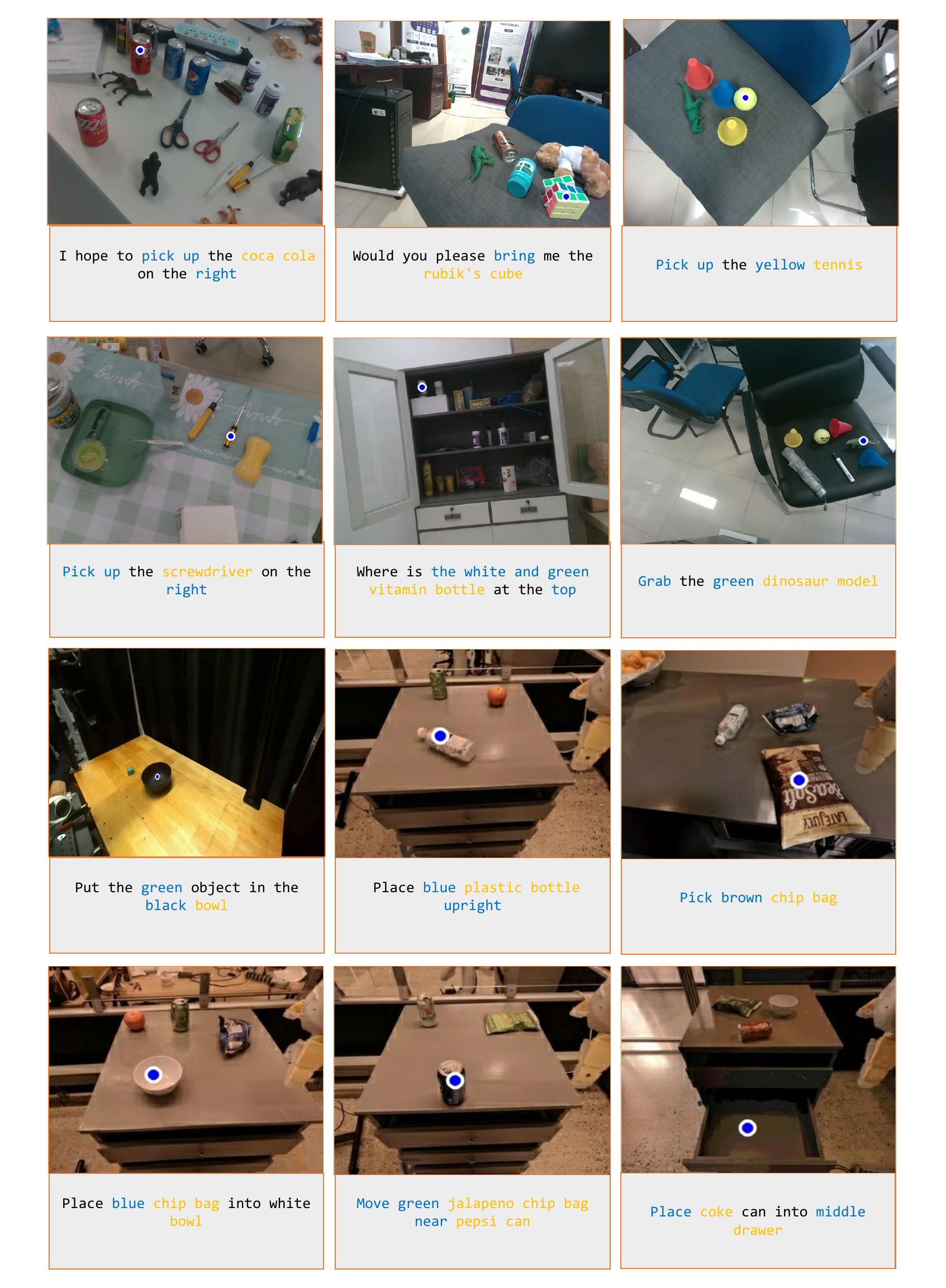}
    \caption{Embodied affordance prediction example 5.}
    \label{fig:afford-5}
\end{figure}

\begin{figure}[H]
    \centering
    \includegraphics[width=1.0\linewidth,height=22cm]{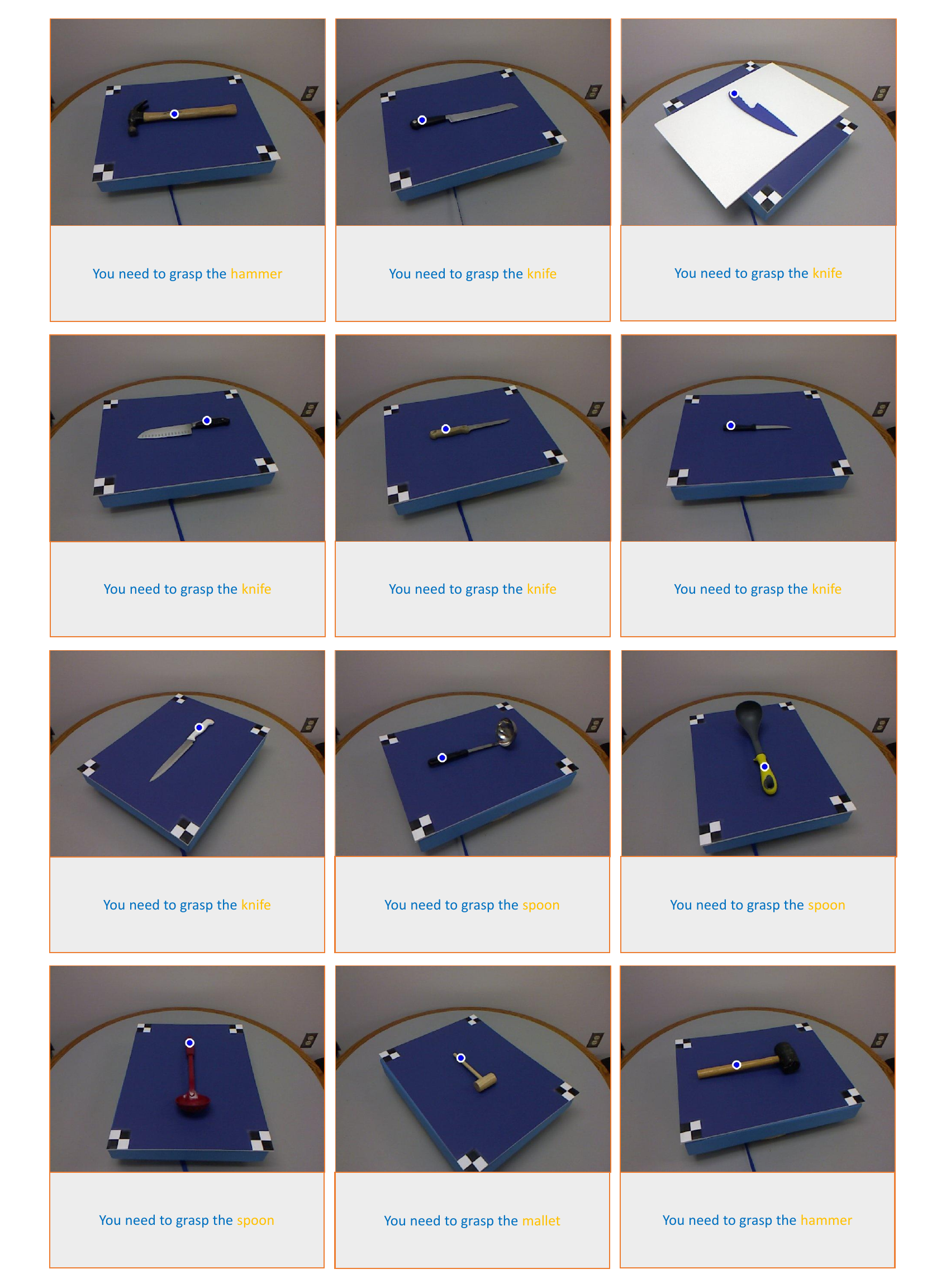}
    \caption{Embodied affordance prediction example 6.}
    \label{fig:afford-6}
\end{figure}

\begin{figure}[H]
    \centering
    \includegraphics[width=1.0\linewidth,height=22cm]{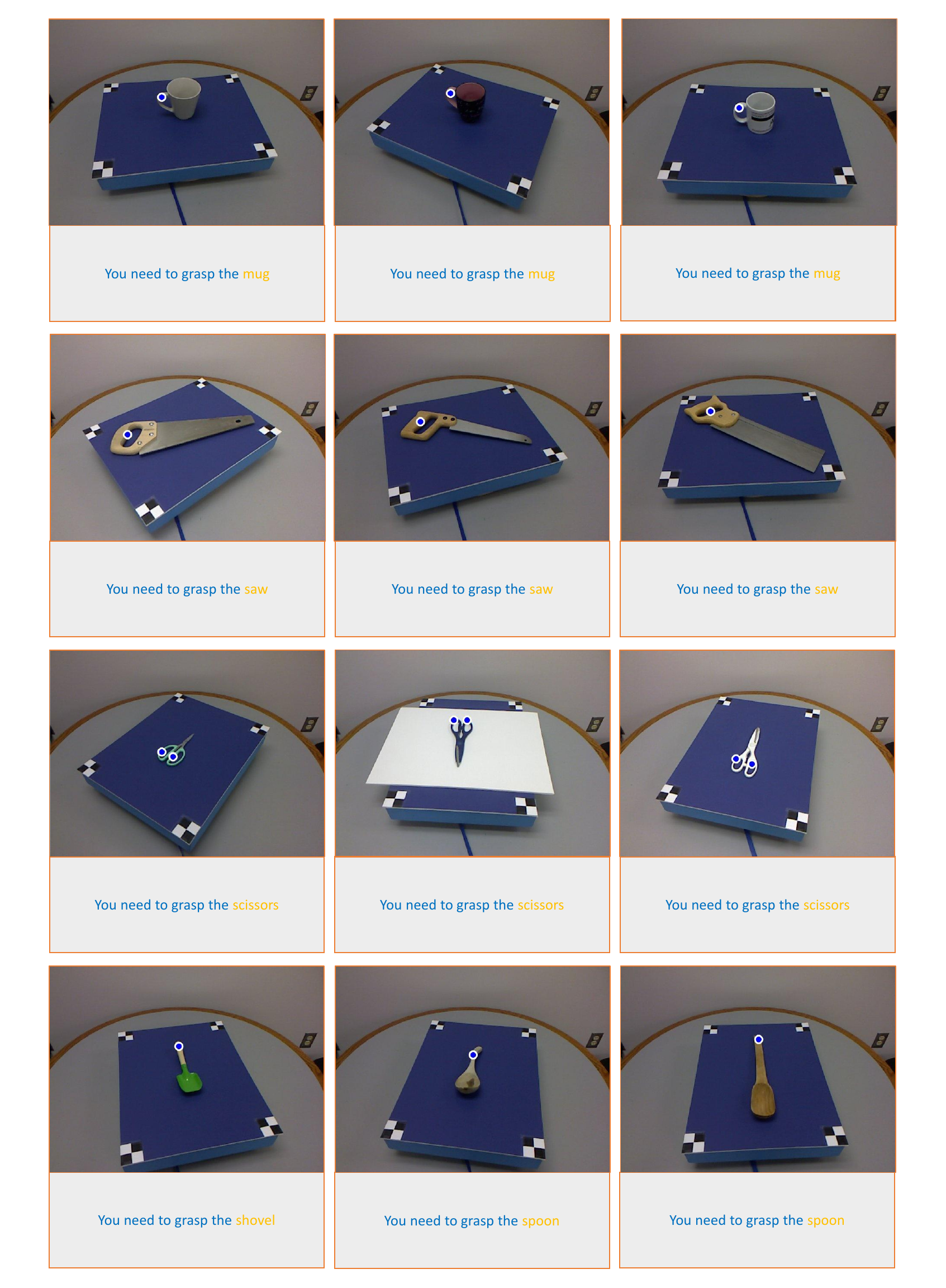}
    \caption{Embodied affordance prediction example 7.}
    \label{fig:afford-7}
\end{figure}

\begin{figure}[H]
    \centering
    \includegraphics[width=1.0\linewidth,height=22cm]{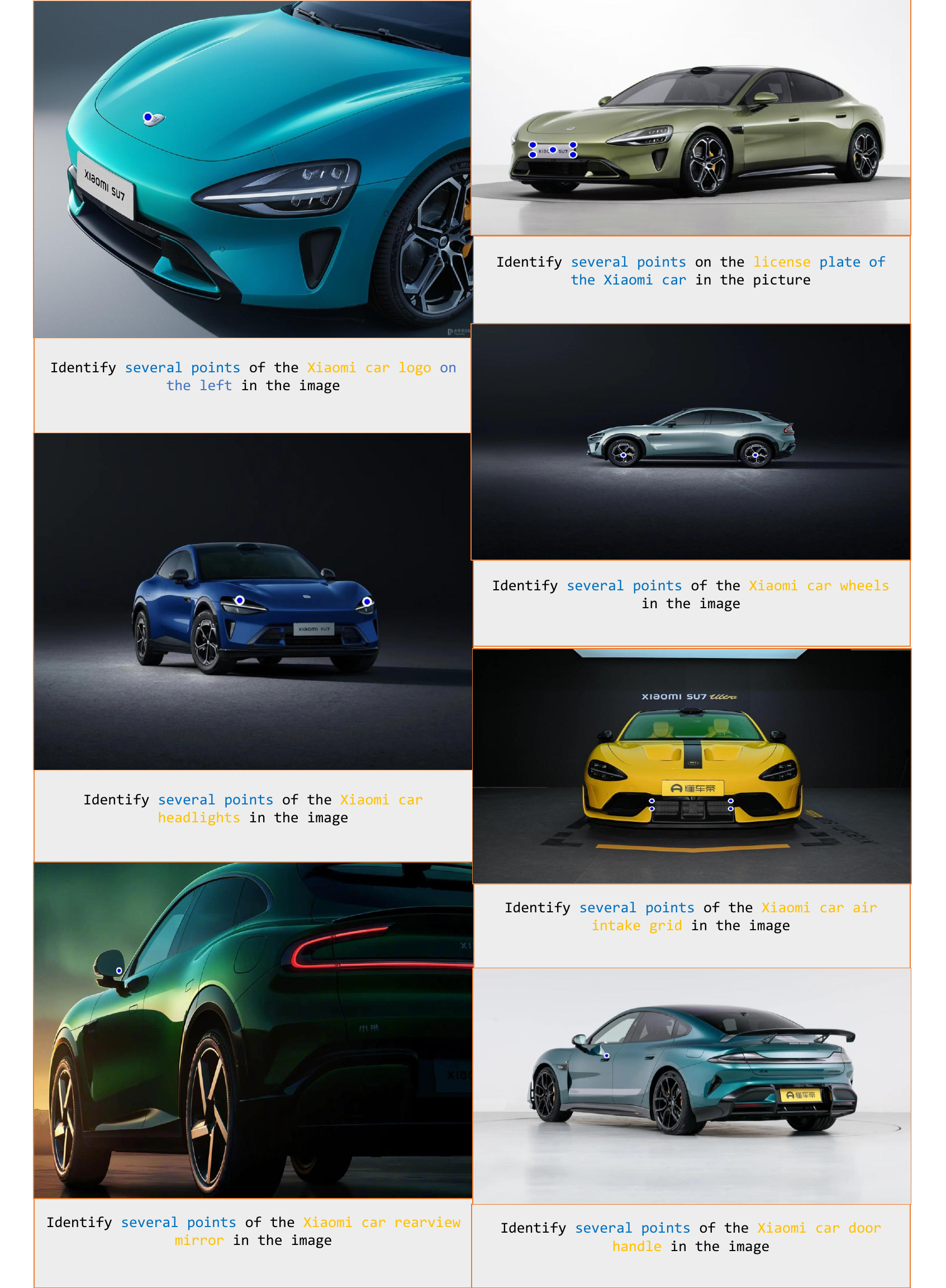}
    \caption{Embodied affordance prediction example 8.}
    \label{fig:afford-8}
\end{figure}

\begin{figure}[H]
    \centering
    \includegraphics[width=1.0\linewidth,height=22cm]{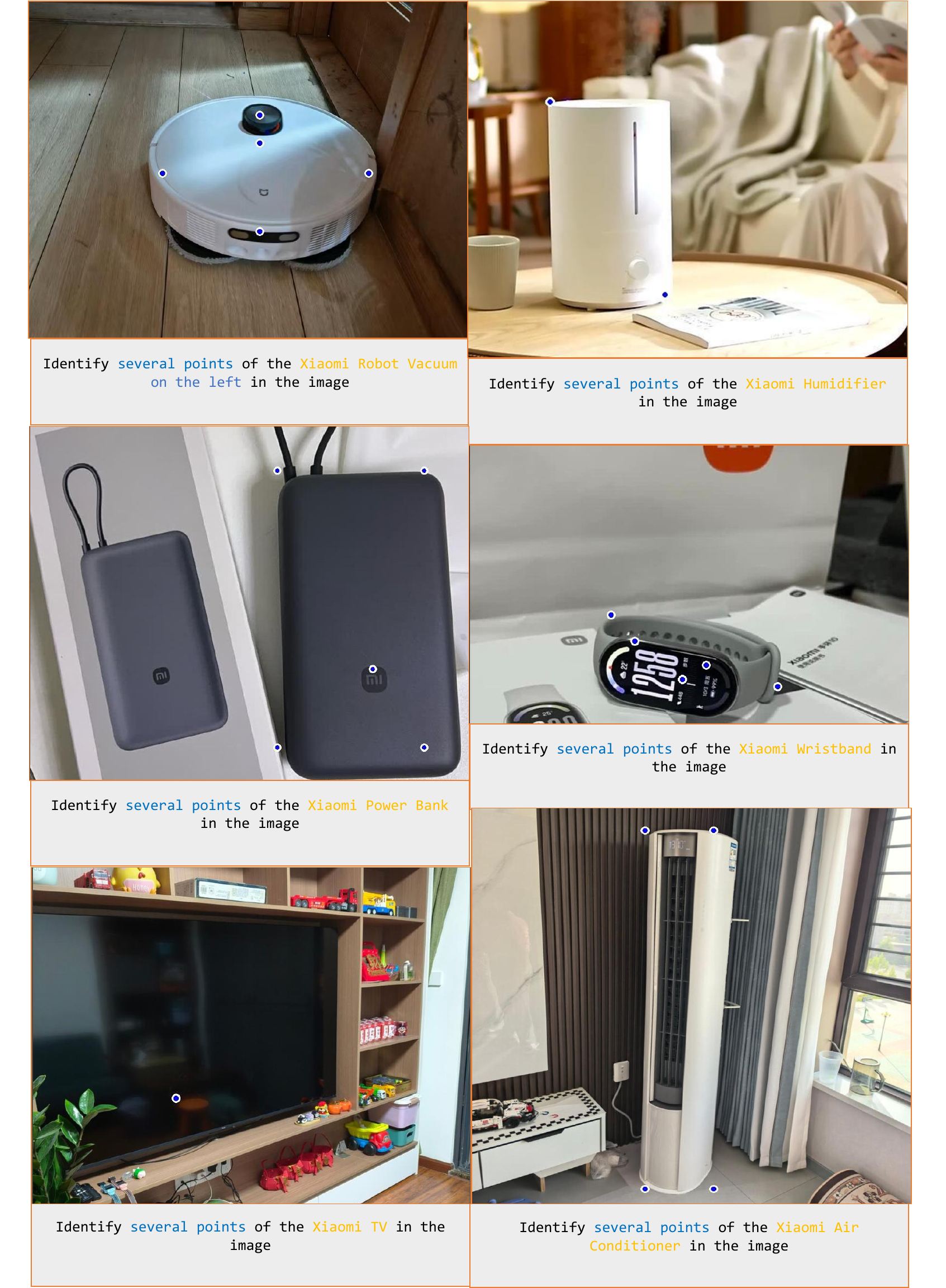}
    \caption{Embodied affordance prediction example 9.}
    \label{fig:afford-9}
\end{figure}

\begin{figure}[H]
    \centering
    \includegraphics[width=1.0\linewidth,height=22cm]{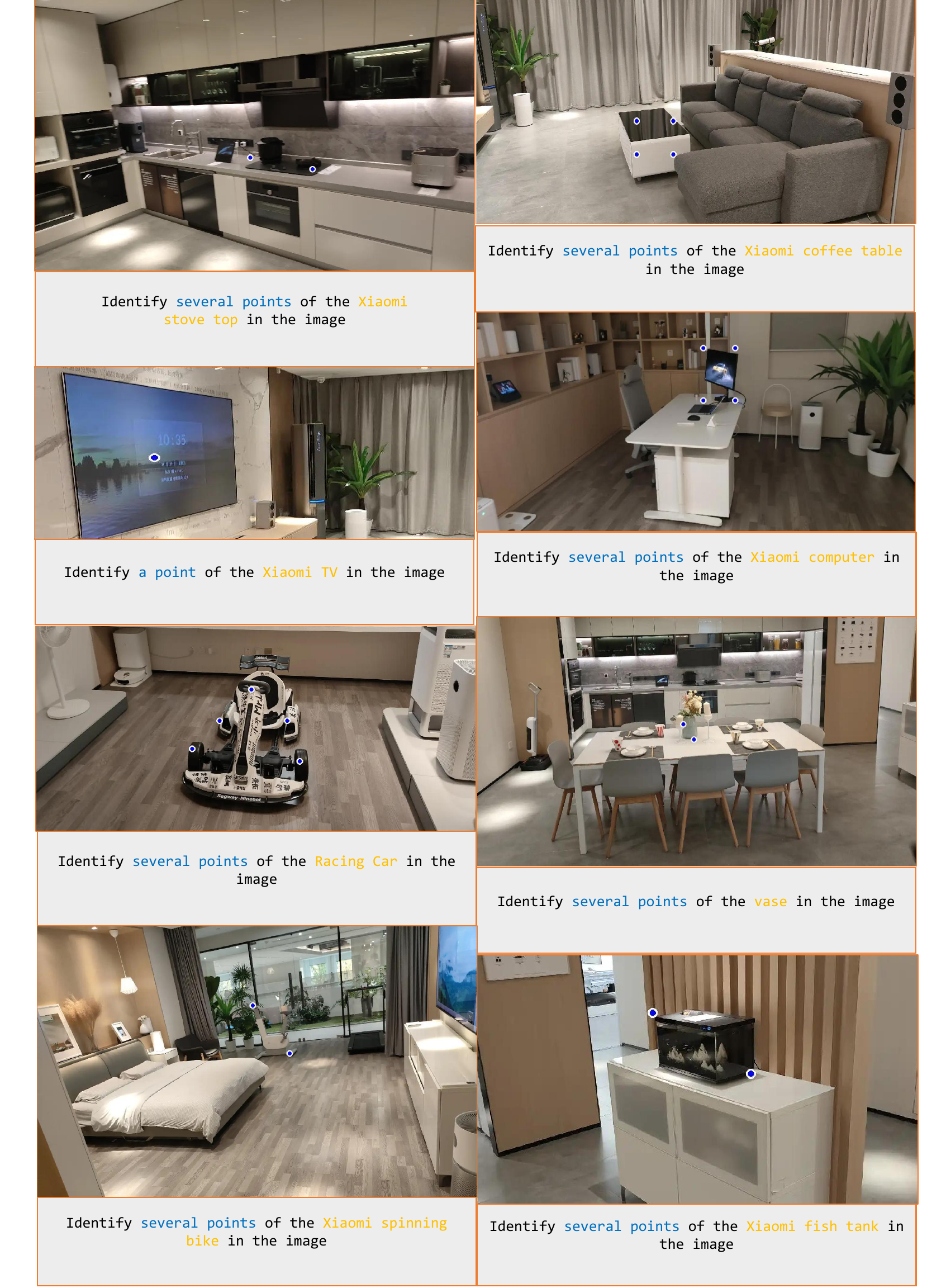}
    \caption{Embodied affordance prediction example 10.}
    \label{fig:afford-10}
\end{figure}

\begin{figure}[H]
    \subsubsection{Planning Task}
    \label{sssec:planning_task}
    \centering
    \includegraphics[height=21.5cm]{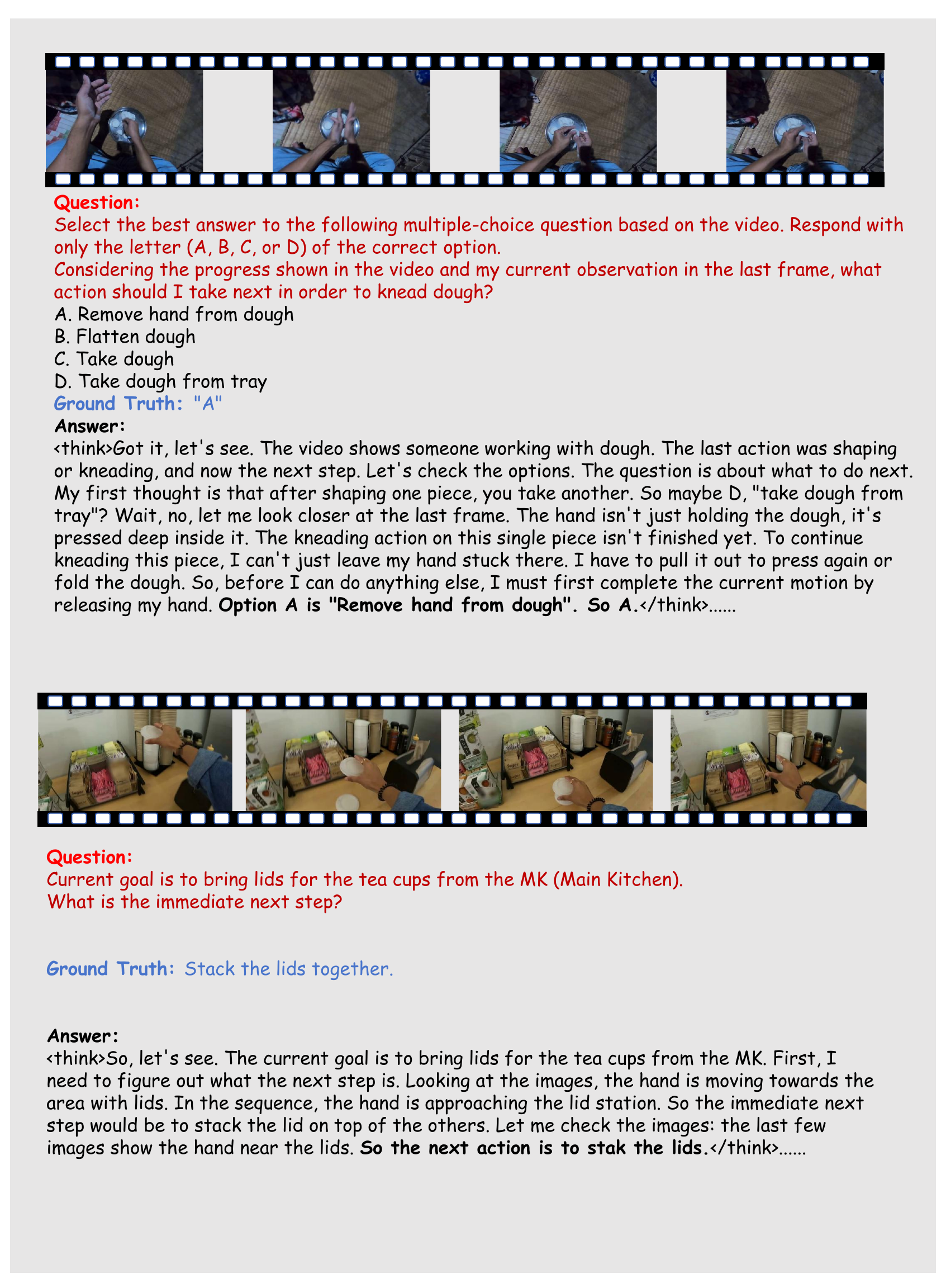}
    \caption{Embodied planning task example 1.}
    \label{fig:planning-1}
\end{figure}

\begin{figure}[H]
    \centering
    \includegraphics[width=1.0\linewidth]{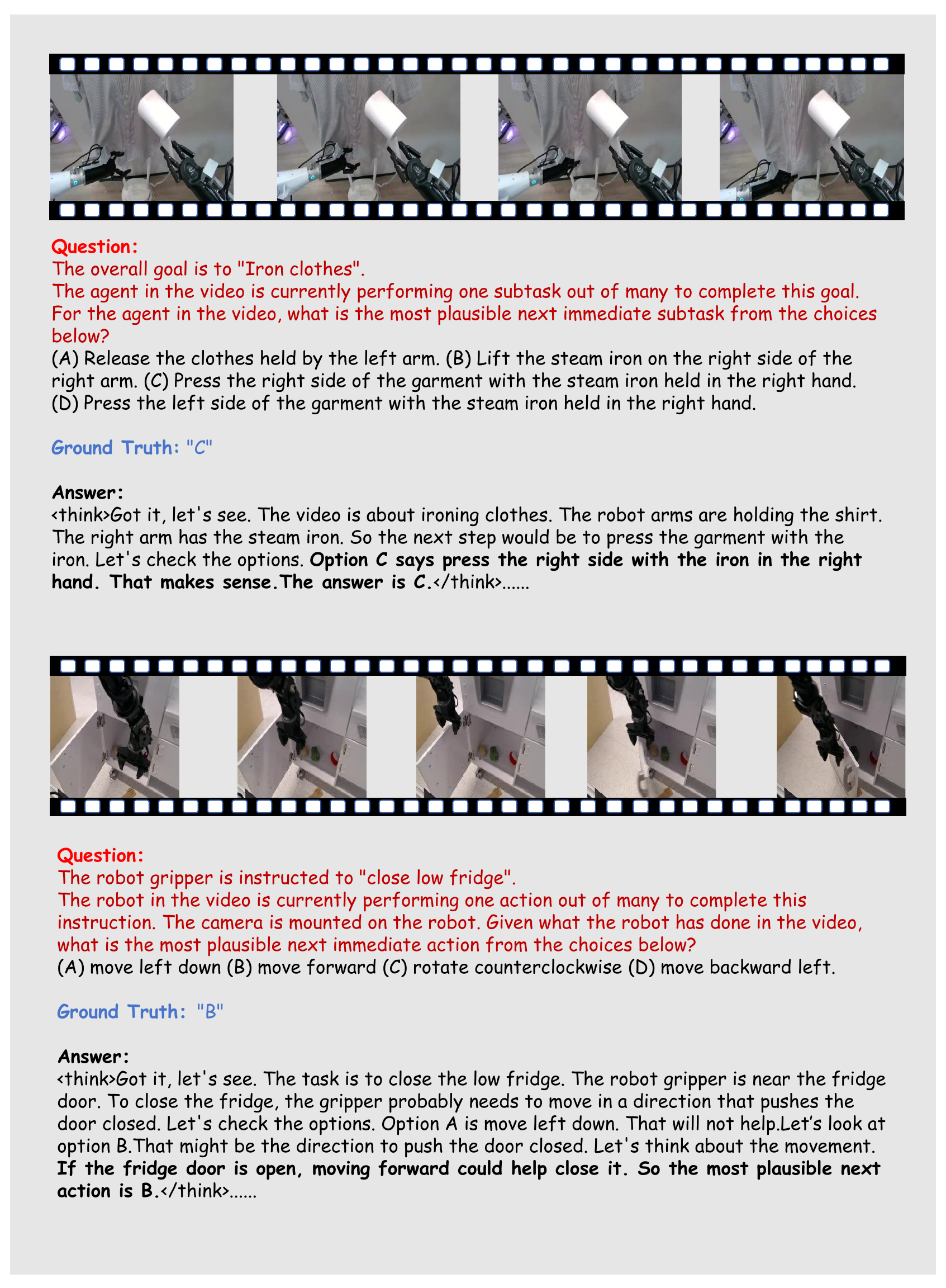}
    \caption{Embodied planning task example 2.}
    \label{fig:planning-2}
\end{figure}

\begin{figure}[H]
    \centering
    \includegraphics[width=1.0\linewidth]{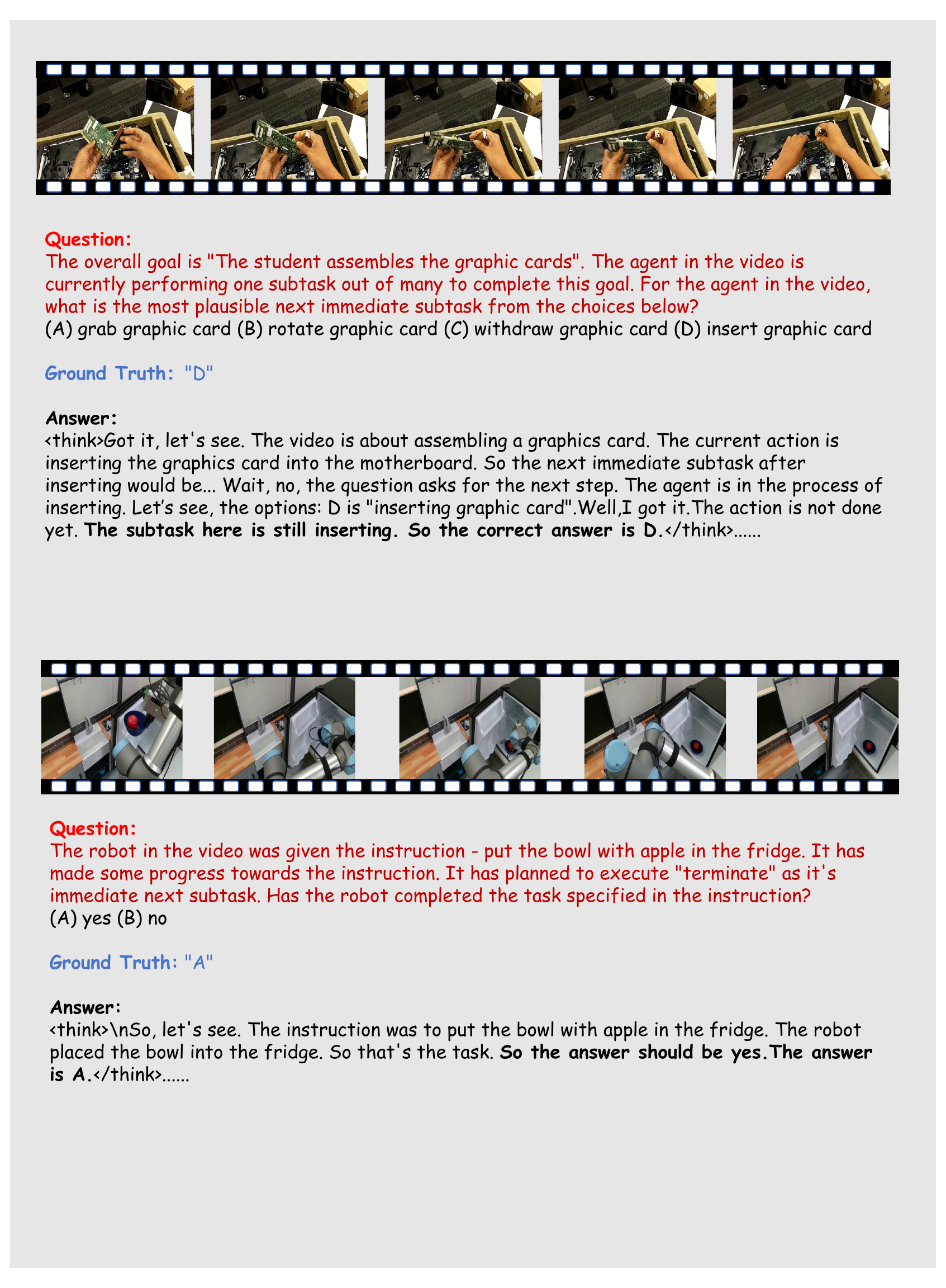}
    \caption{Embodied planning task example 3.}
    \label{fig:planning-3}
\end{figure}

\begin{figure}[H]
    \centering
    \includegraphics[width=1.0\linewidth, height=21.2cm]{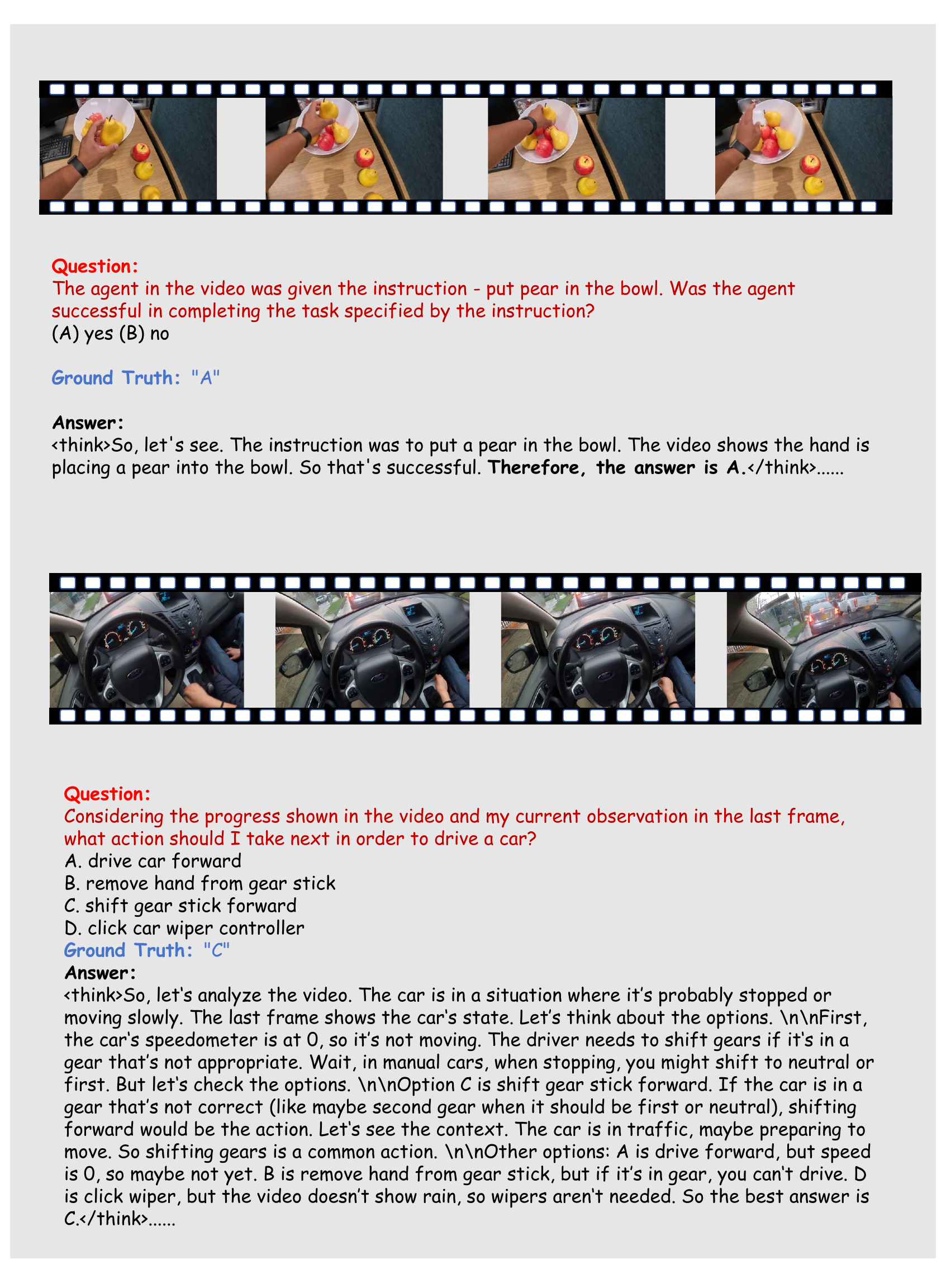}
    \caption{Embodied planning task example 4.}
    \label{fig:planning-4}
\end{figure}

\subsection{Autonomous Driving Visualization Examples}
\label{subsec:appendix_autodriving}

\begin{figure}[H]
    \subsubsection{Scene perception}
    \label{sssec:av_perception}
    \centering
    \includegraphics[width=1.0\linewidth,height=20cm]{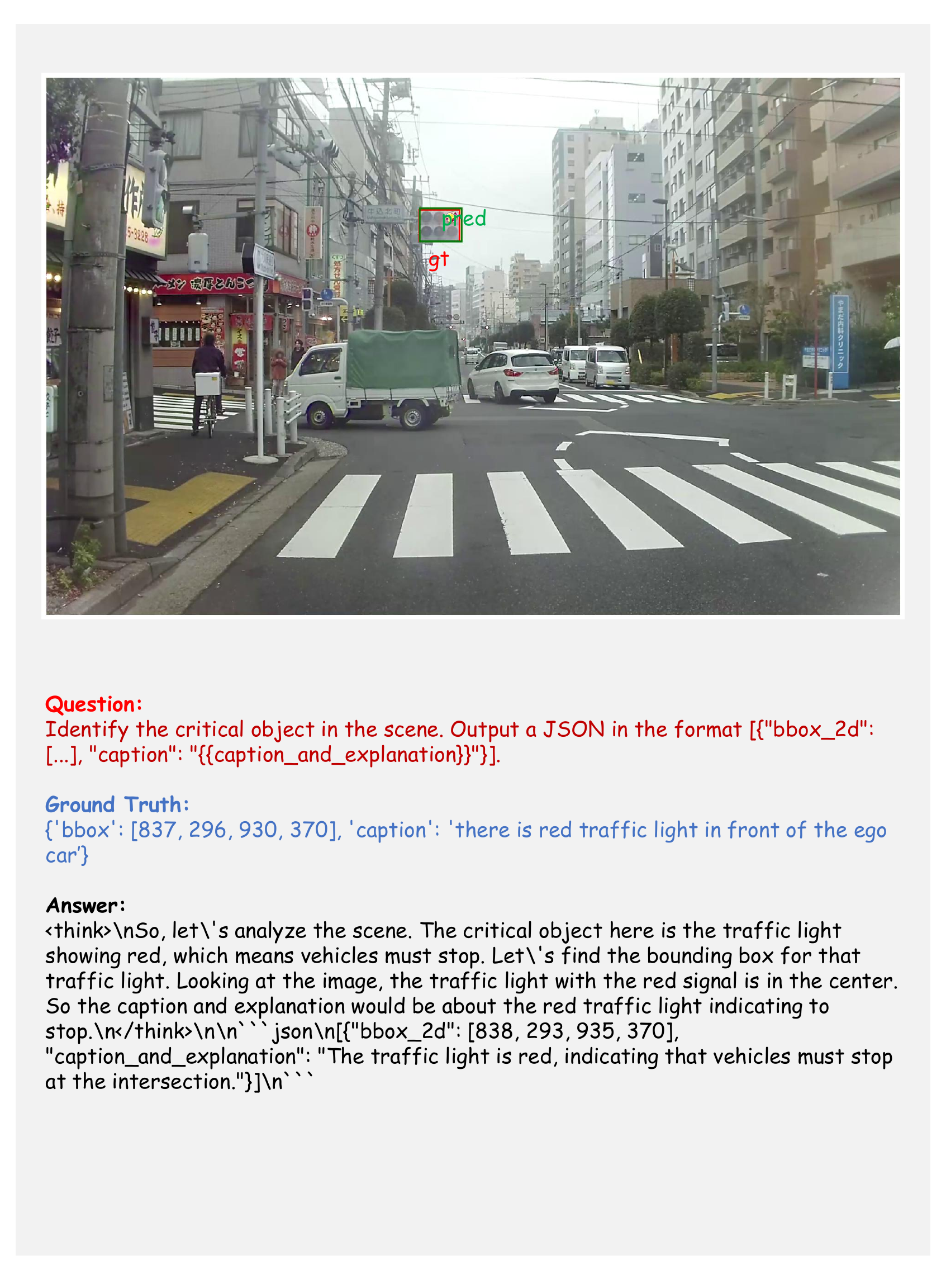}
    \caption{Autonomous driving scene perception example 1.}
    \label{fig:av-perception-1}
\end{figure}

\begin{figure}[H]
    \centering
    \includegraphics[width=1.0\linewidth]{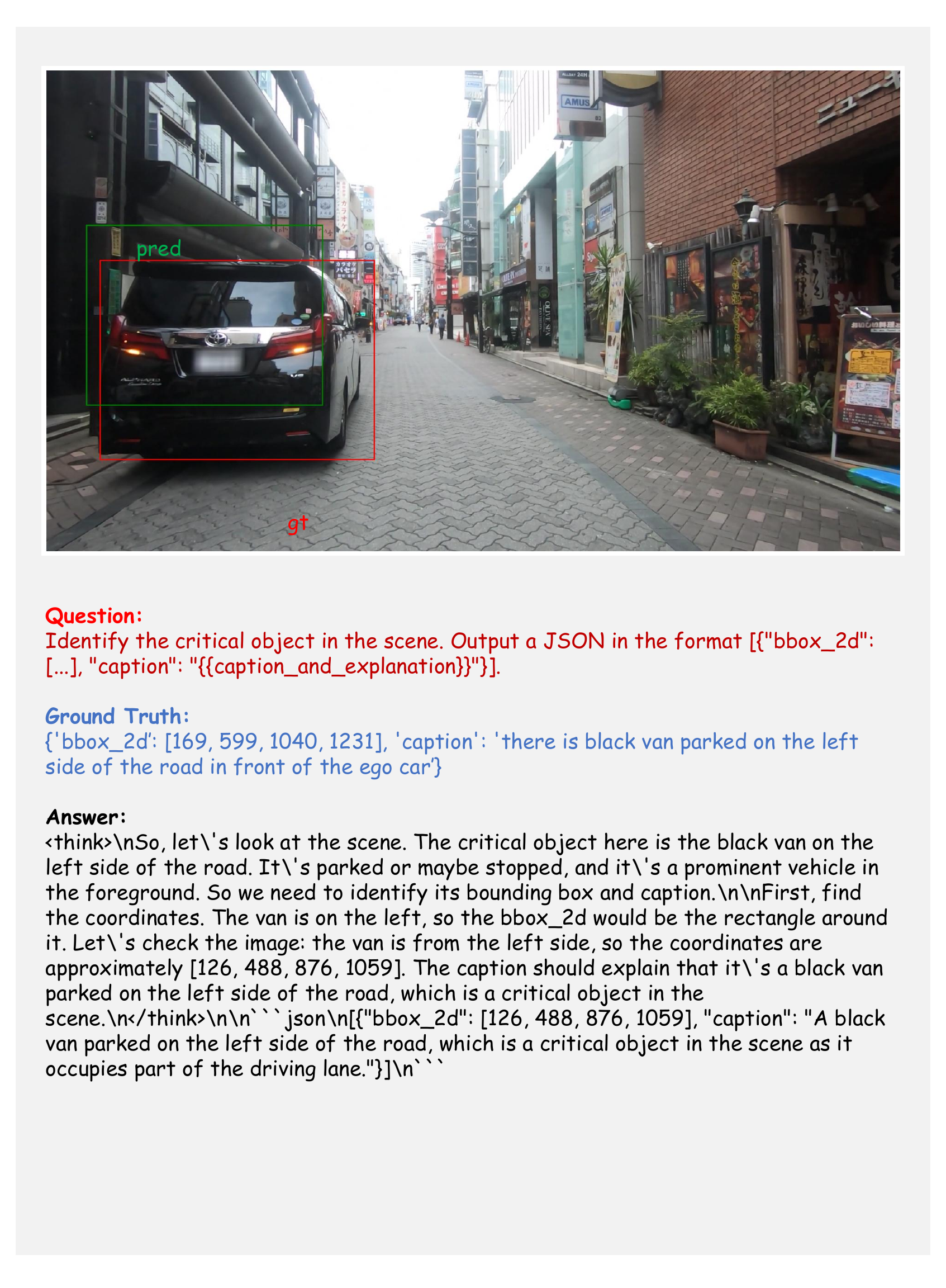}
    \caption{Autonomous driving scene perception example 2.}
    \label{fig:av-perception-2}
\end{figure}

\begin{figure}[H]
    \centering
    \includegraphics[width=1.0\linewidth, height=21.5cm]{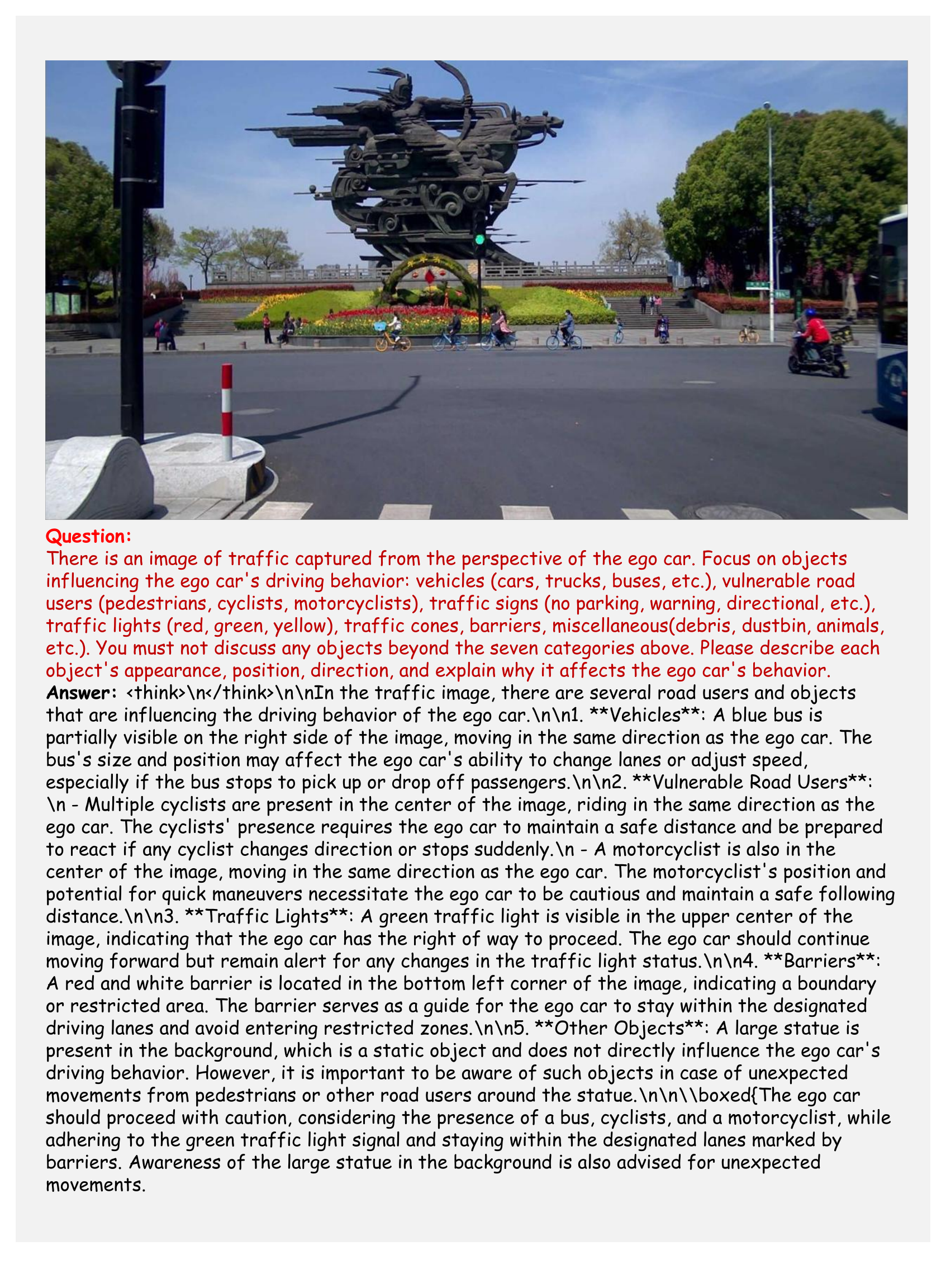}
    \caption{Autonomous driving scene perception example 3.}
    \label{fig:av-perception-3}
\end{figure}

\begin{figure}[H]
    \centering
    \includegraphics[width=1.0\linewidth]{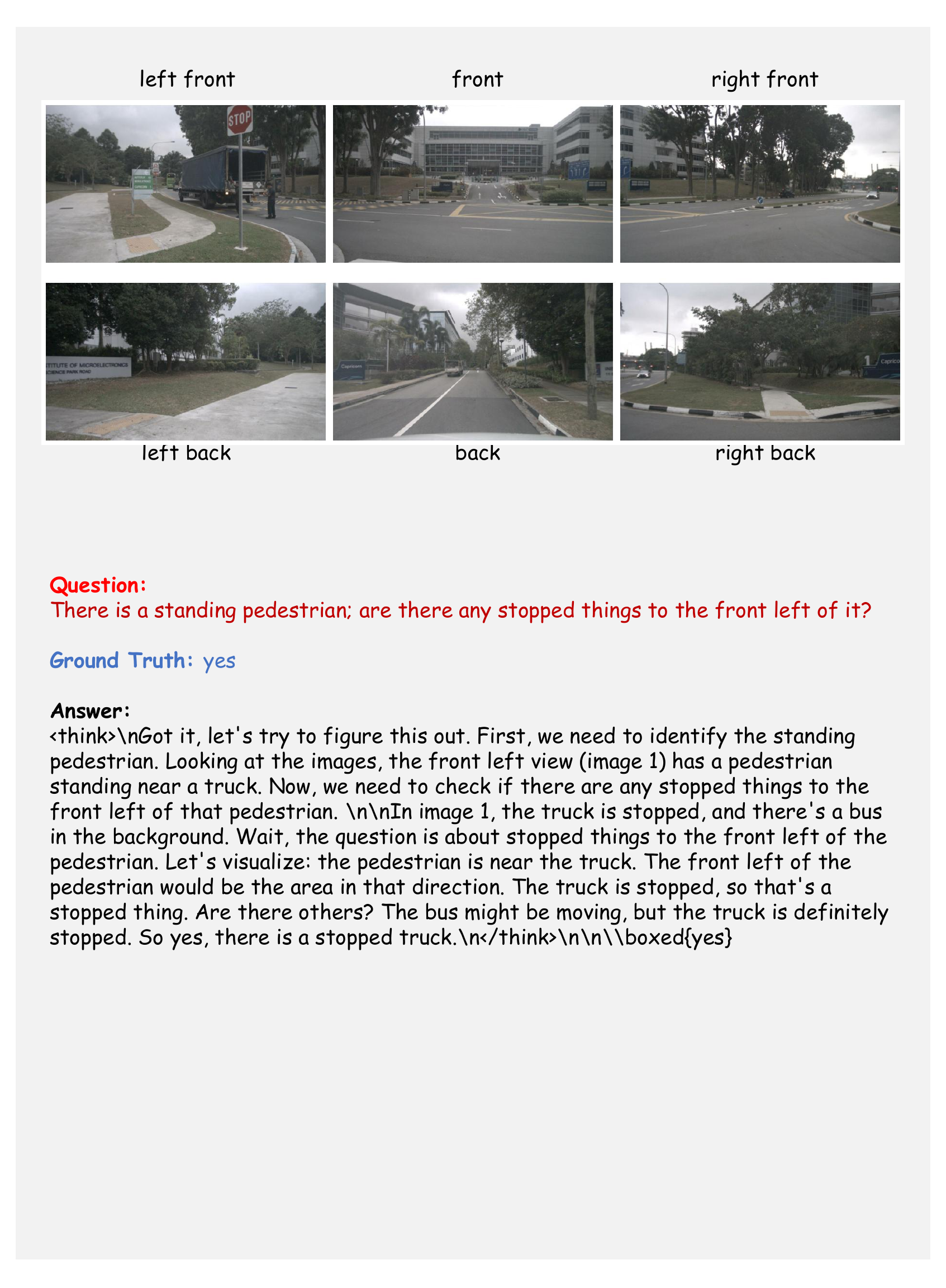}
    \caption{Autonomous driving scene perception example 4.}
    \label{fig:av-perception-4}
\end{figure}

\begin{figure}[H]
    \centering
    \includegraphics[width=1.0\linewidth]{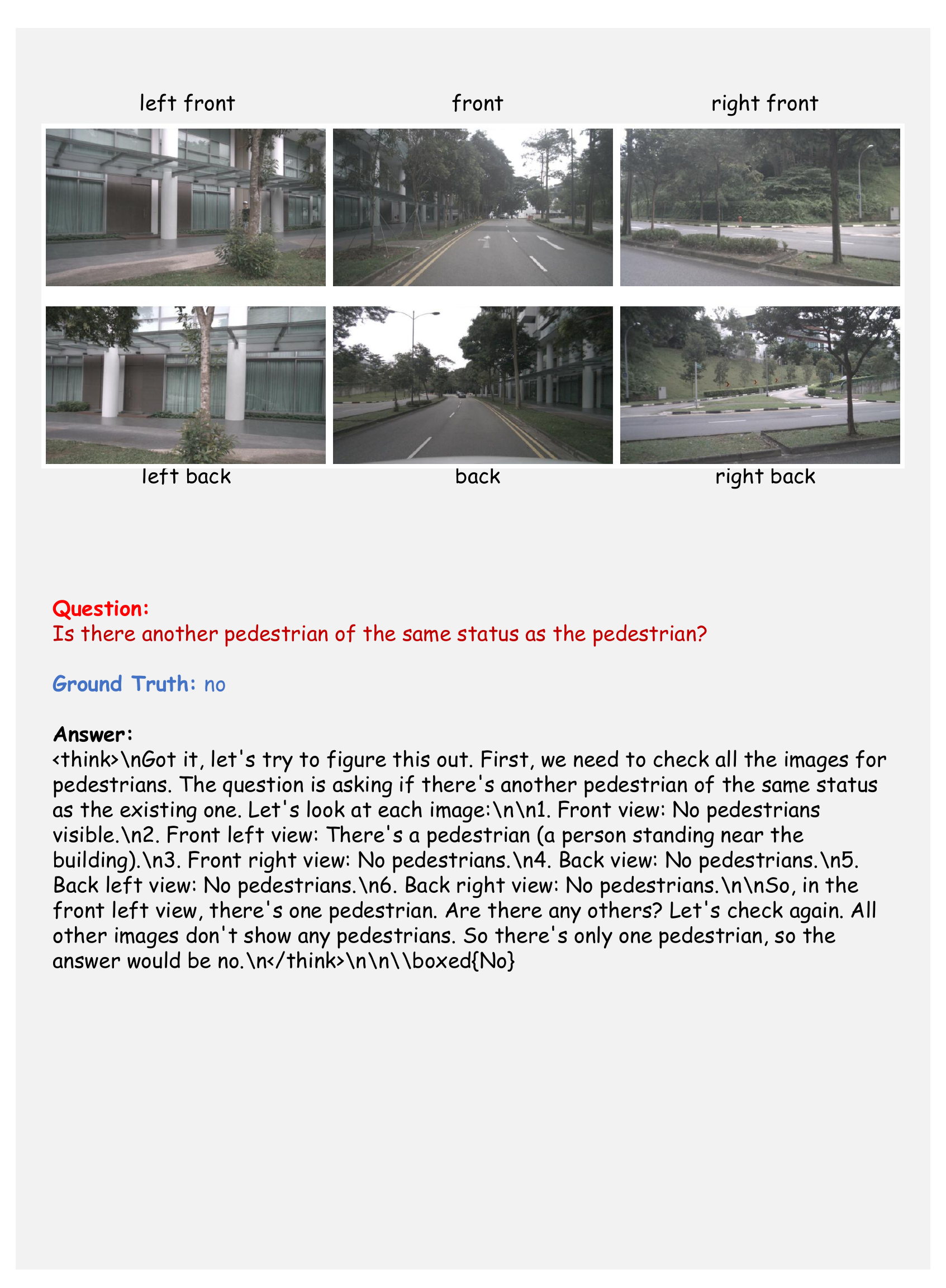}
    \caption{Autonomous driving scene perception example 5.}
    \label{fig:av-perception-5}
\end{figure}

\begin{figure}[H]
    \centering
    \includegraphics[width=1.0\linewidth]{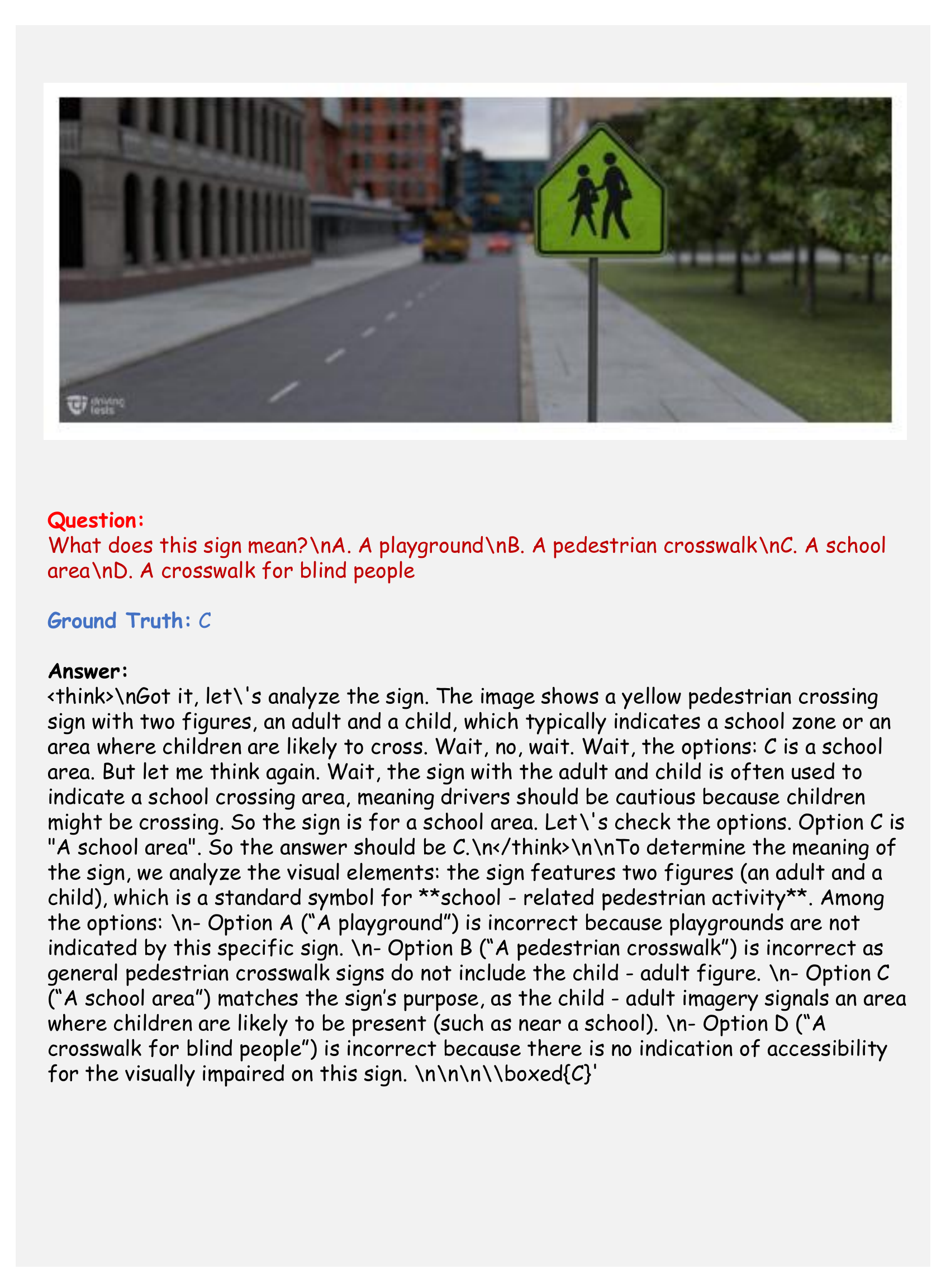}
    \caption{Autonomous driving scene perception example 6.}
    \label{fig:av-perception-6}
\end{figure}

\begin{figure}[H]
    \centering
    \includegraphics[width=1.0\linewidth]{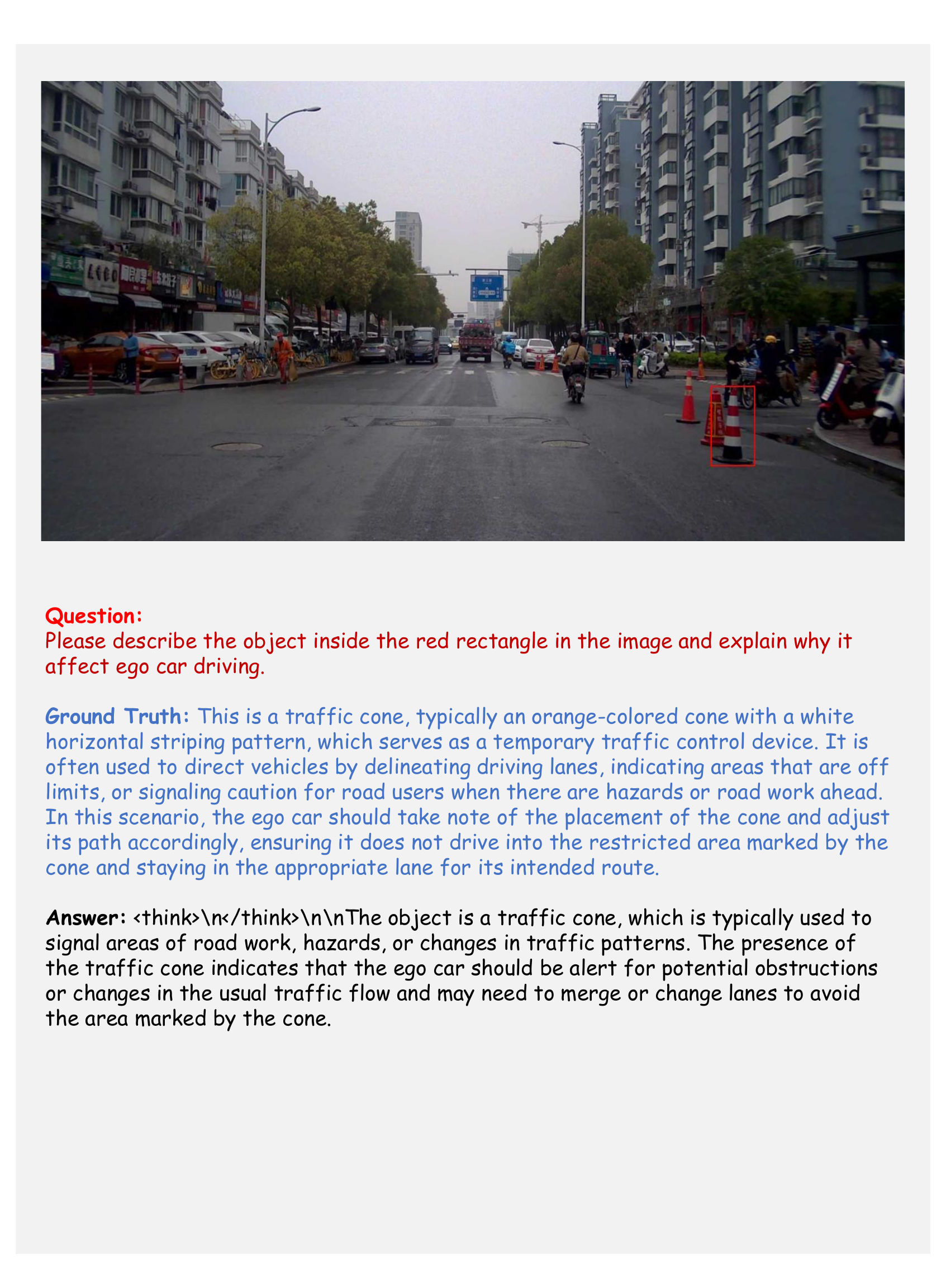}
    \caption{Autonomous driving scene perception example 7.}
    \label{fig:av-perception-7}
\end{figure}

\begin{figure}[H]
    \centering
    \includegraphics[width=1.0\linewidth]{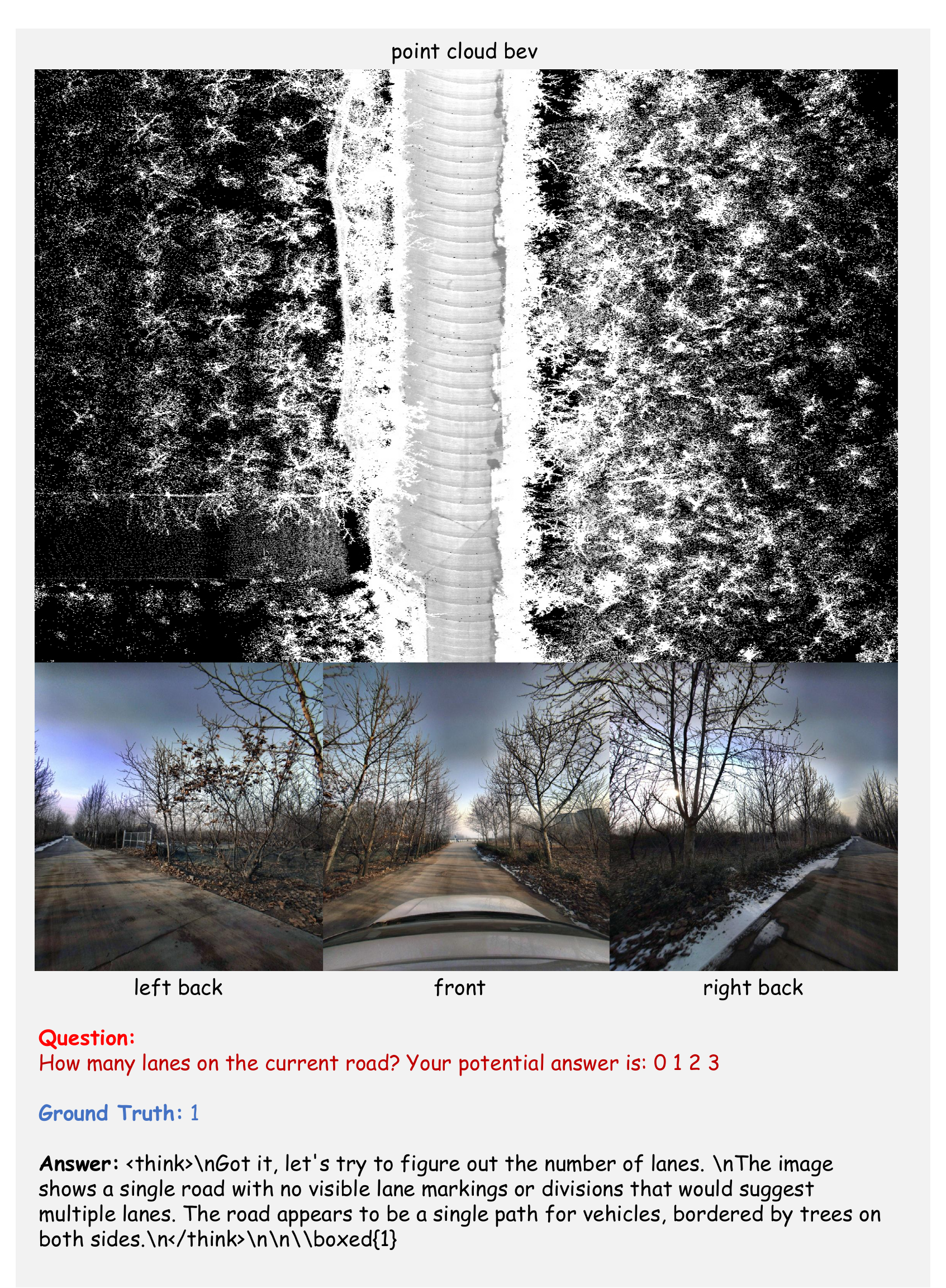}
    \caption{Autonomous driving scene perception example 8.}
    \label{fig:av-perception-8}
\end{figure}

\begin{figure}[H]
    \subsubsection{Prediction ability}
    \label{sssec:av_prediction}
    \centering
    \includegraphics[width=1.0\linewidth, height=18.5cm]{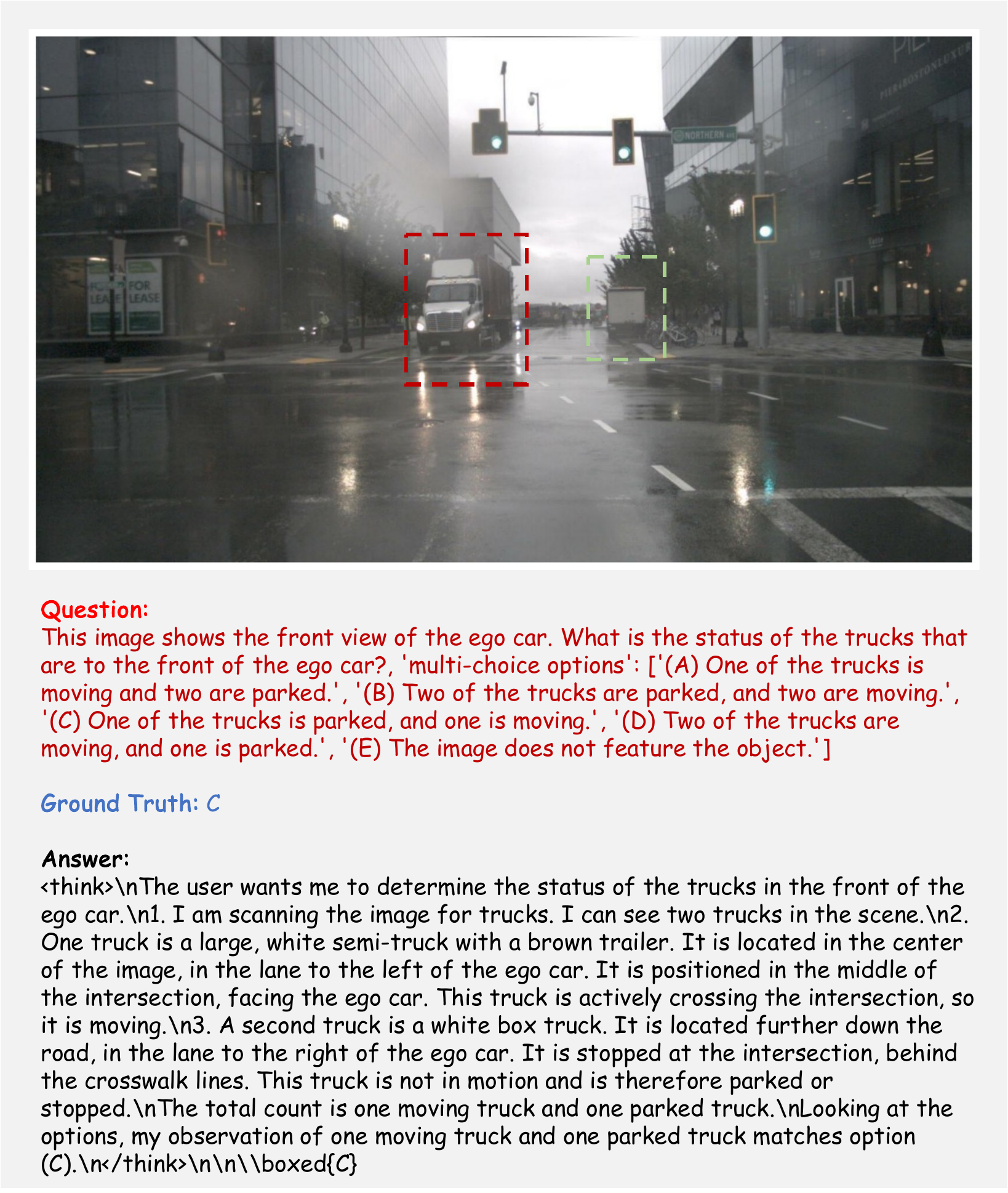}
    \caption{Autonomous driving prediction ability example 1.}
    \label{fig:av-prediction-1}
\end{figure}

\begin{figure}[H]
    \centering
    \includegraphics[width=1.0\linewidth]{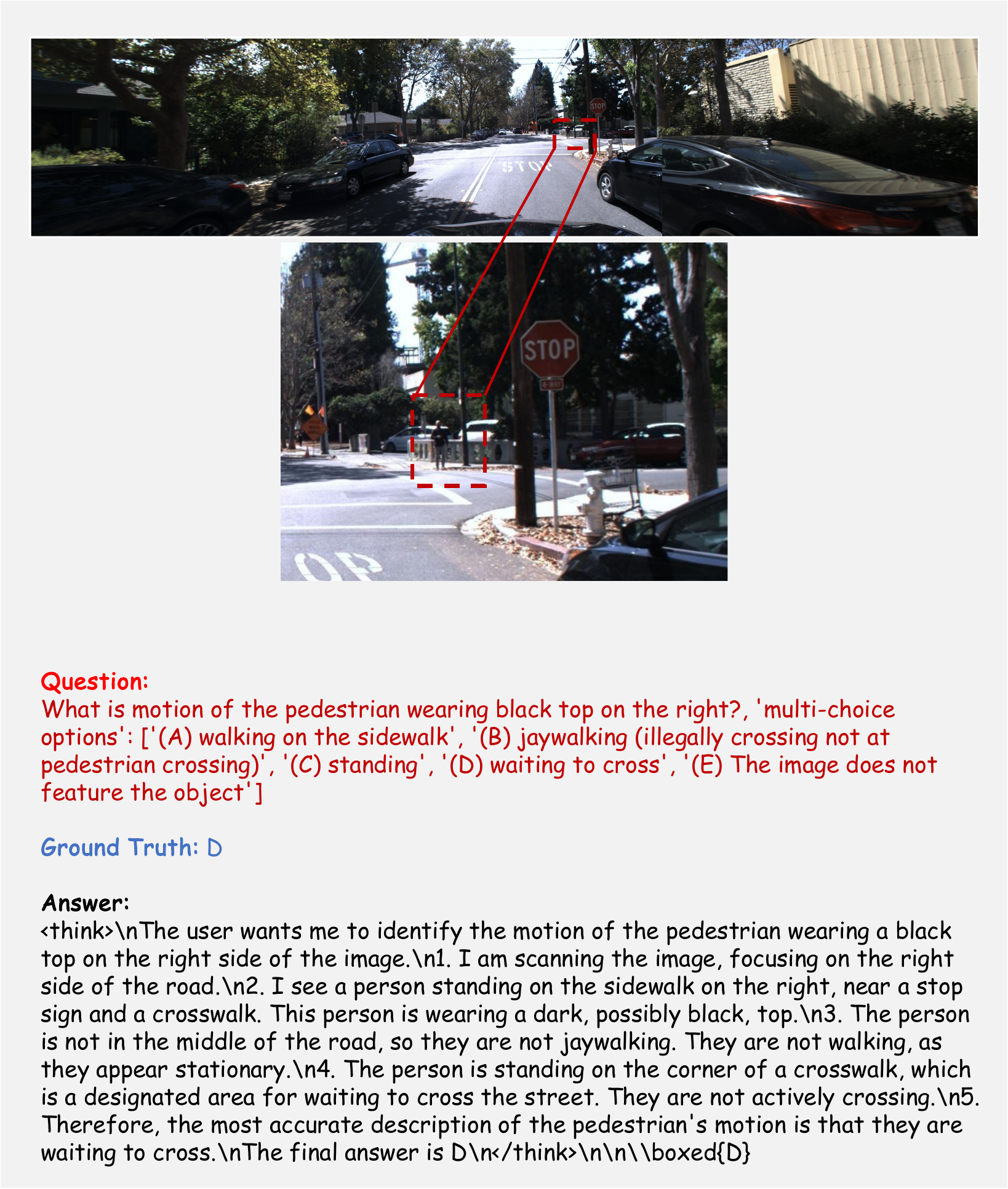}
    \caption{Autonomous driving prediction ability example 2.}
    \label{fig:av-prediction-2}
\end{figure}

\begin{figure}[H]
    \centering
    \includegraphics[width=1.0\linewidth]{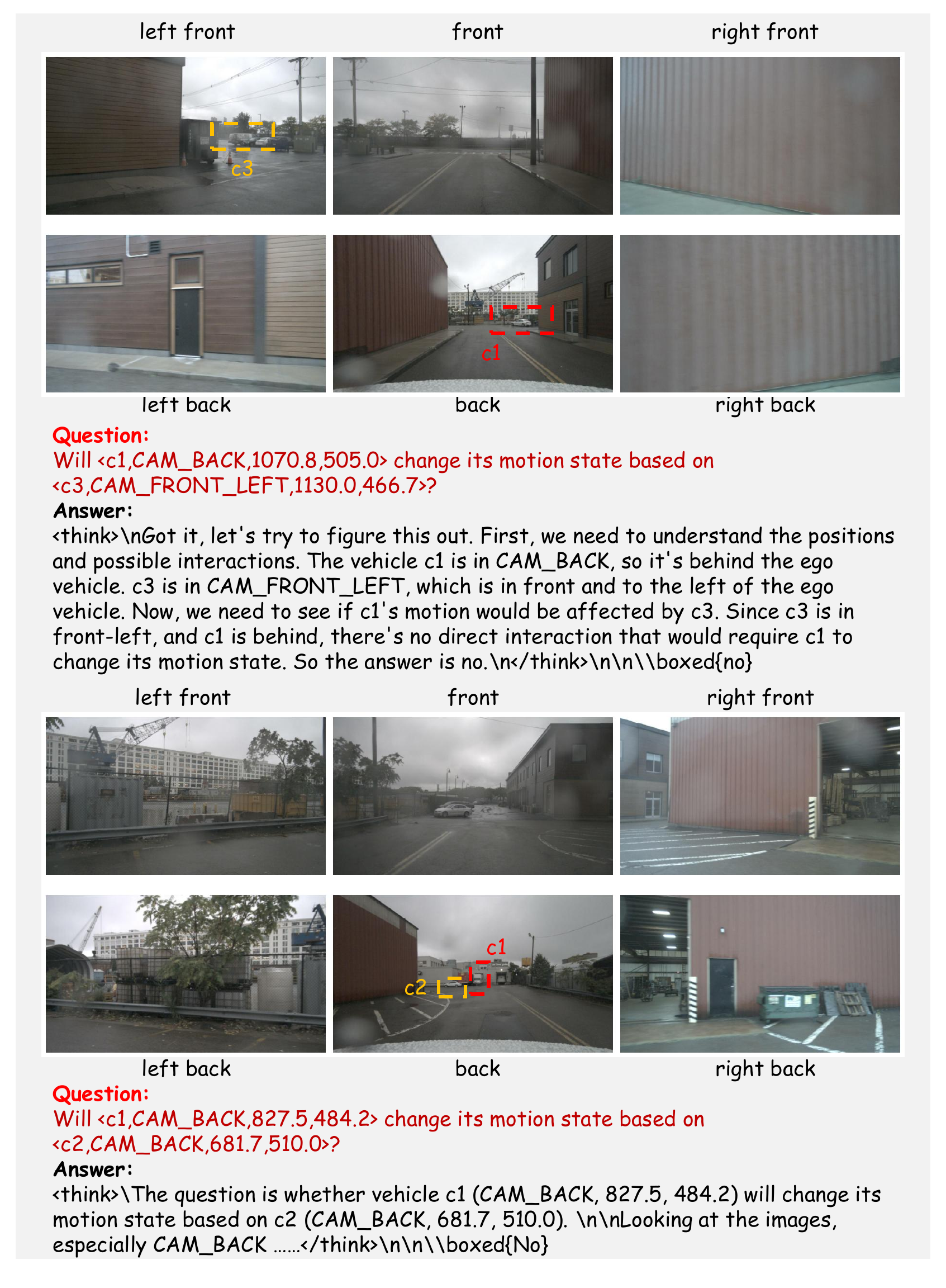}
    \caption{Autonomous driving prediction ability example 3.}
    \label{fig:av-prediction-3}
\end{figure}

\begin{figure}[H]
    \subsubsection{Planning Task}
    \label{sssec:av_planning}
    \centering
    \includegraphics[width=1.0\linewidth, height=21.5cm]{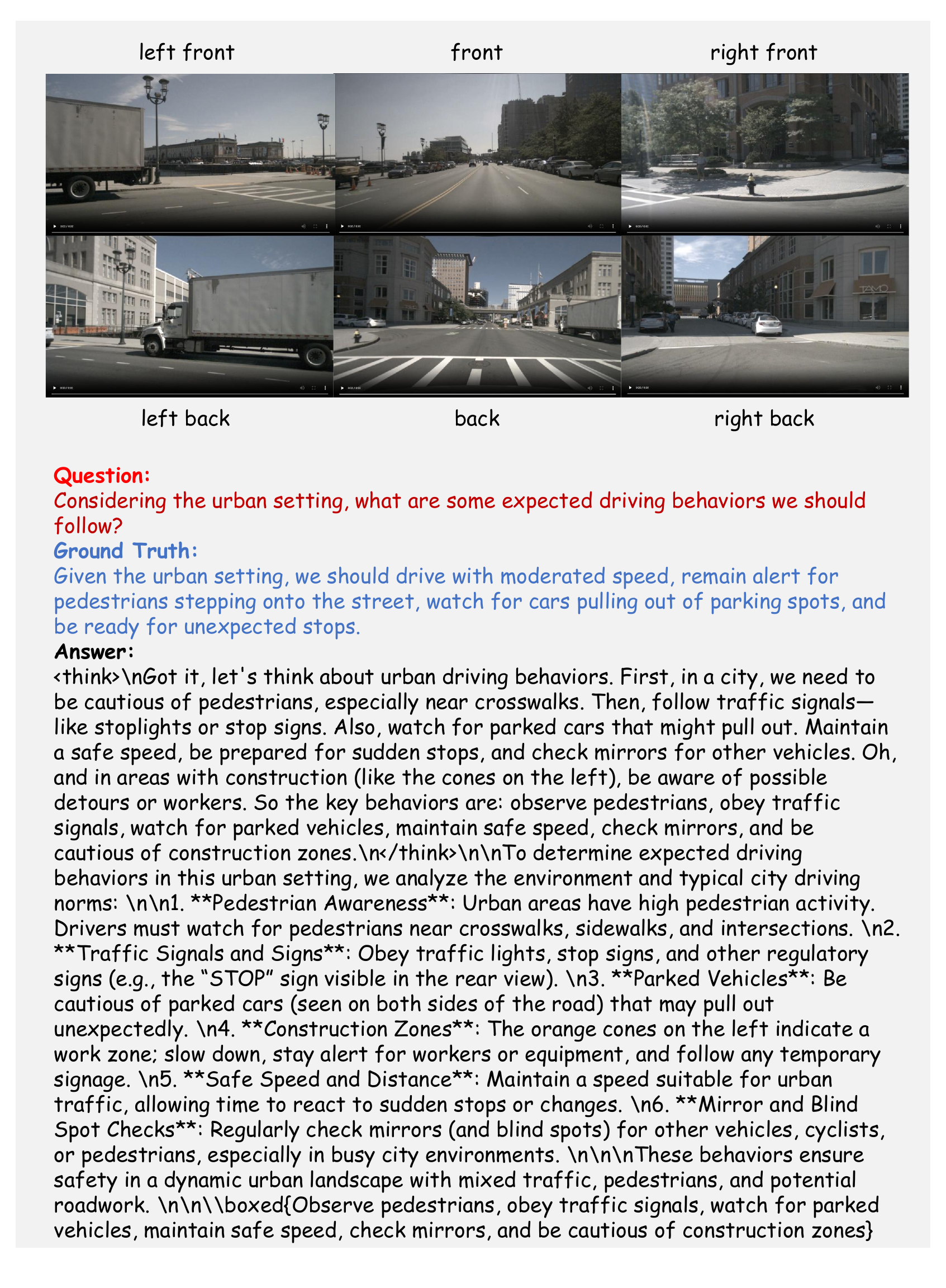}
    \caption{Autonomous driving planning task example 1.}
    \label{fig:av-planning-1}
\end{figure}

\begin{figure}[H]
    \centering
    \includegraphics[width=1.0\linewidth]{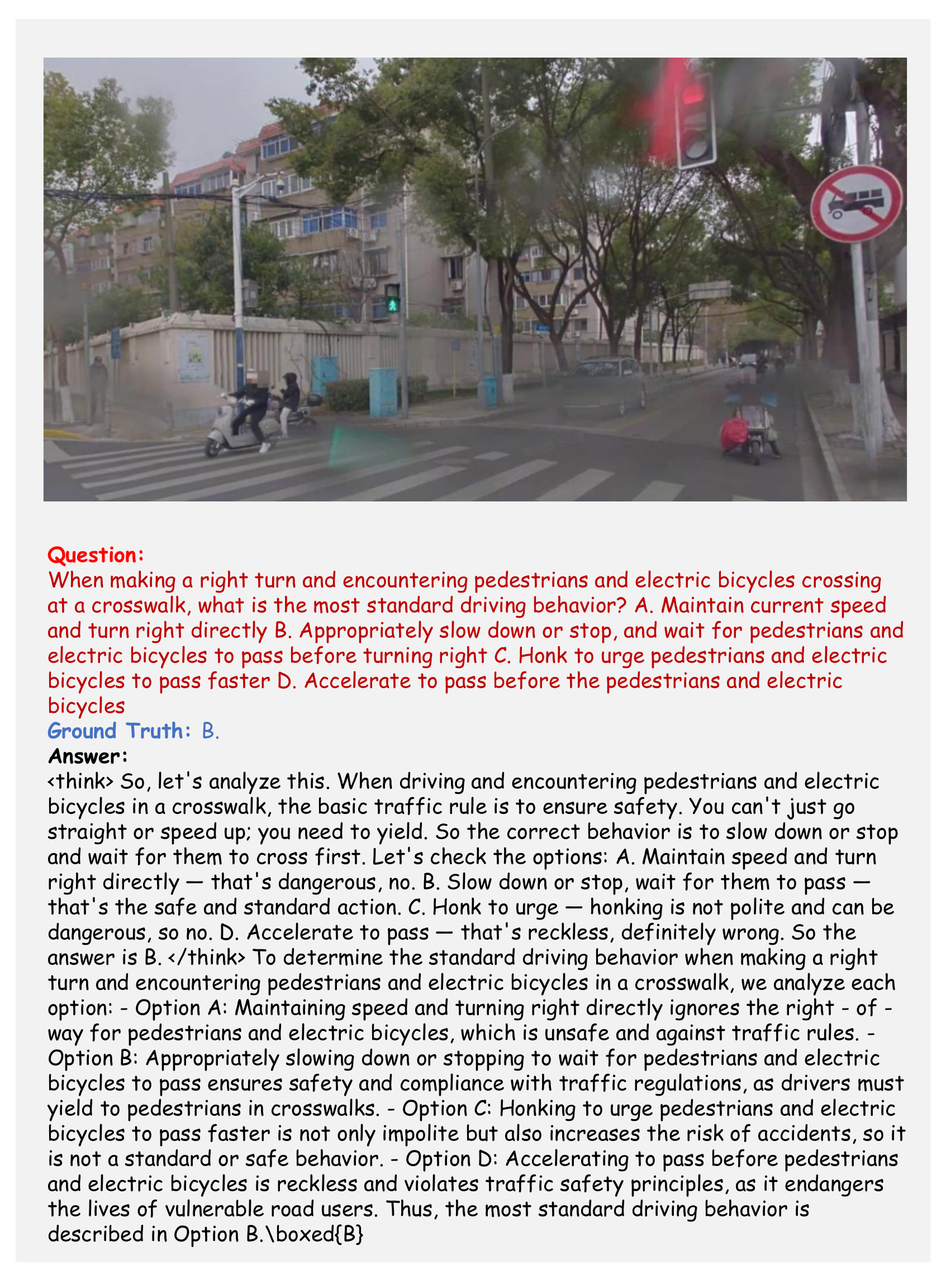}
    \caption{Autonomous driving planning task example 2.}
    \label{fig:av-planning-2}
\end{figure}

\begin{figure}[H]
    \centering
    \includegraphics[width=1.0\linewidth]{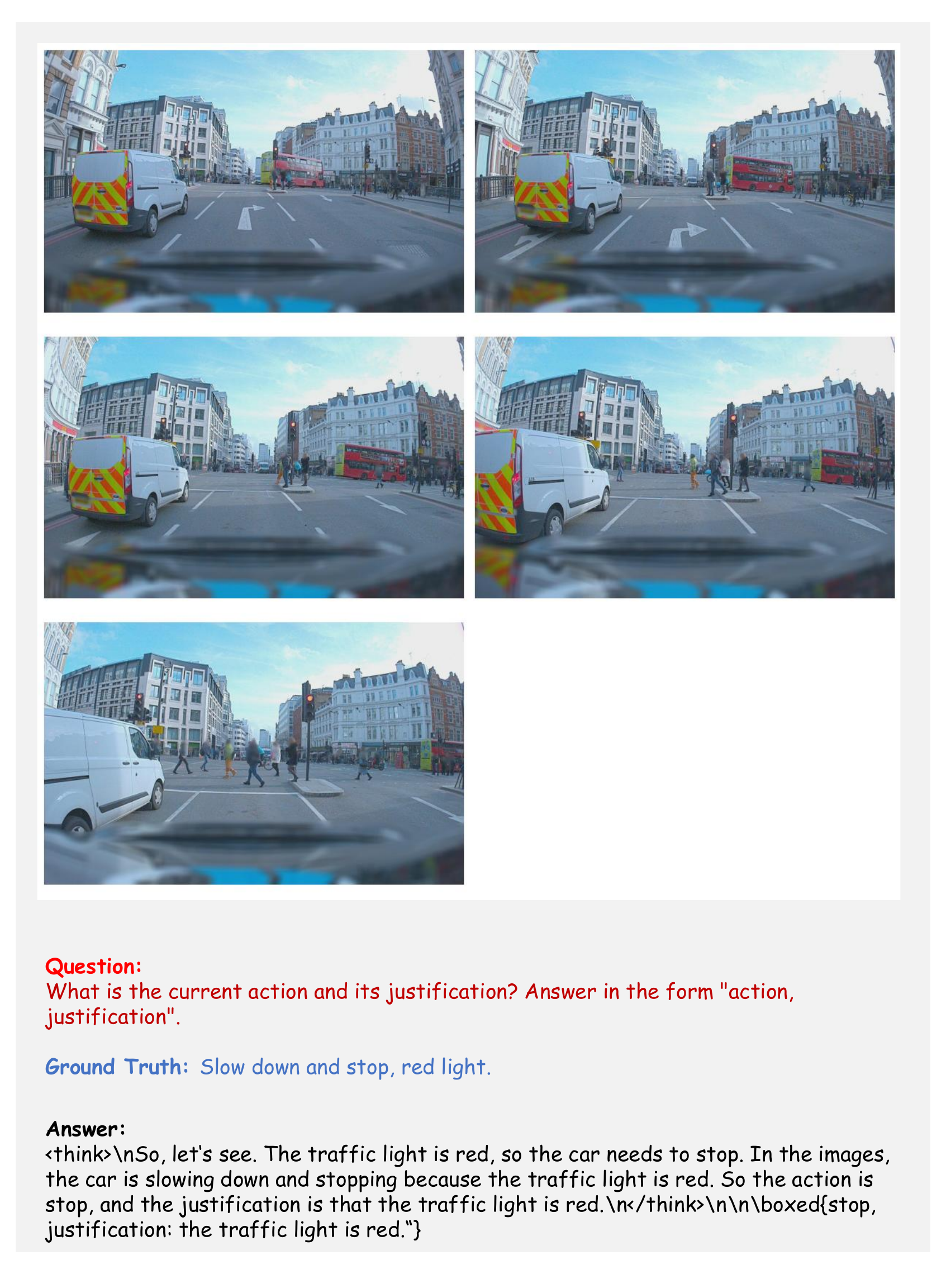}
    \caption{Autonomous driving planning task example 3.}
    \label{fig:av-planning-3}
\end{figure}

\begin{figure}[H]
    \centering
    \includegraphics[width=1.0\linewidth]{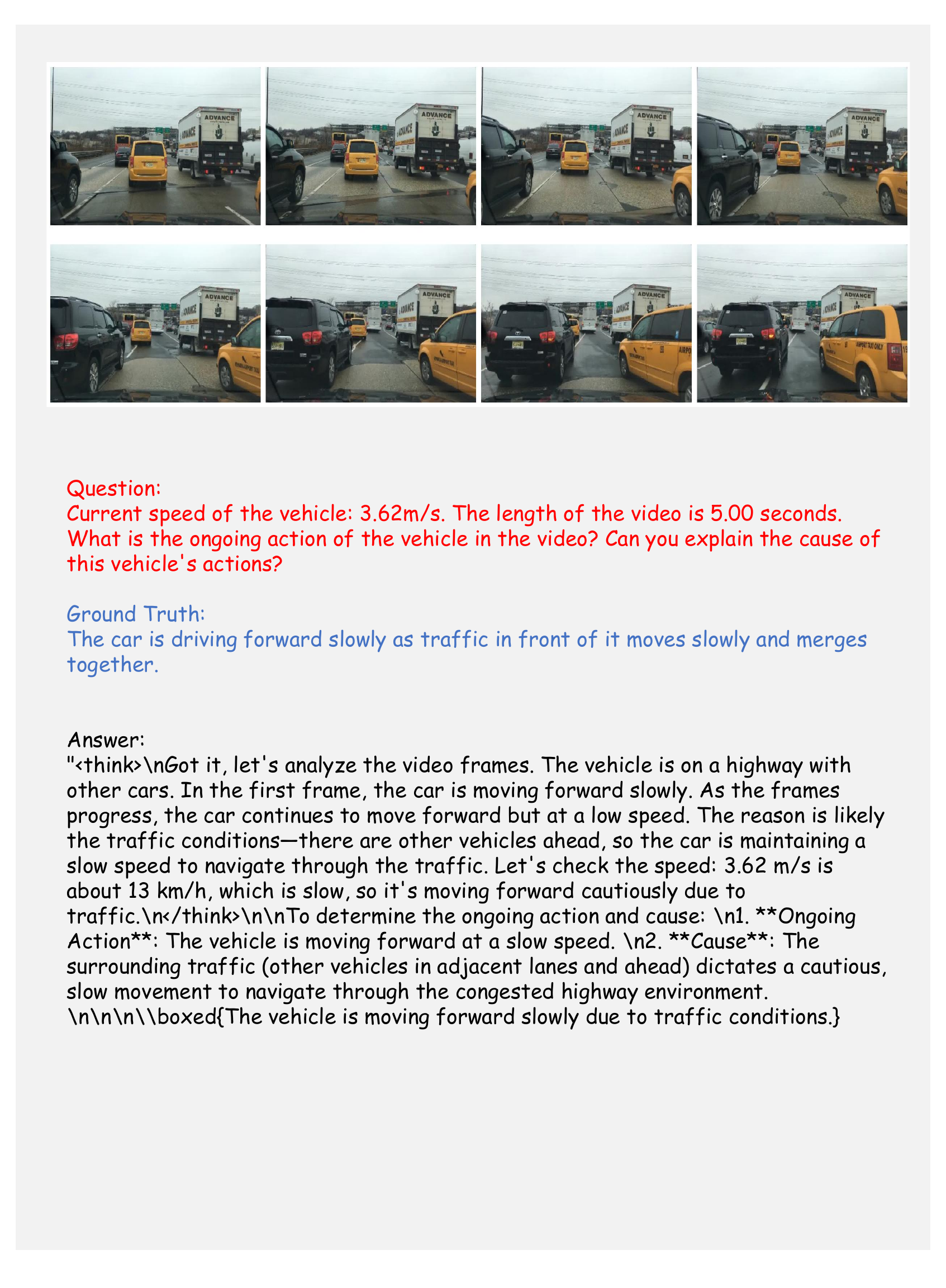}
    \caption{Autonomous driving planning task example 4.}
    \label{fig:av-planning-4}
\end{figure}

\begin{figure}[H]
    \centering
    \includegraphics[width=1.0\linewidth, height=21.2cm]{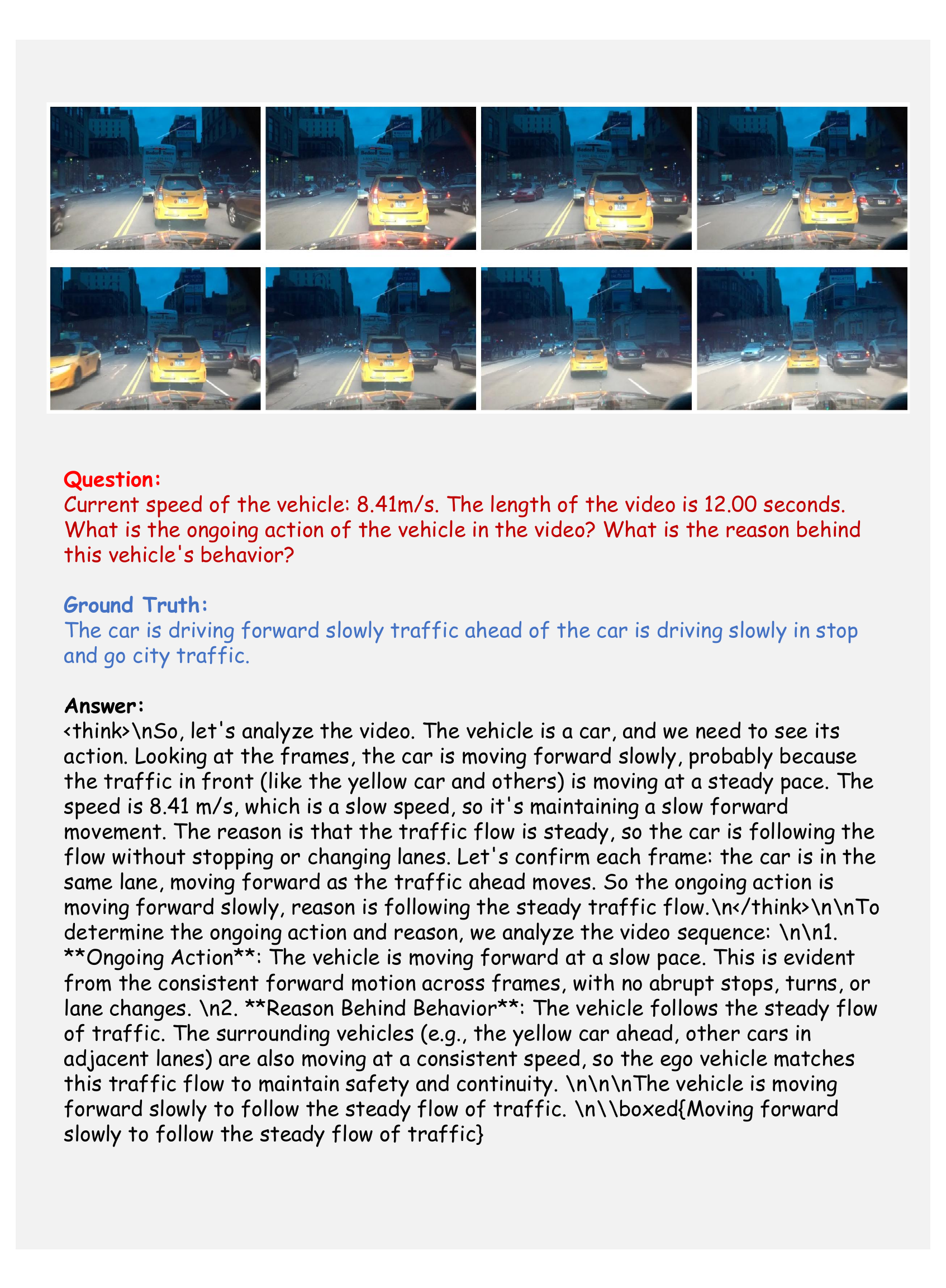}
    \caption{Autonomous driving planning task example 5.}
    \label{fig:av-planning-5}
\end{figure}

\begin{figure}[H]
    \centering
    \includegraphics[width=1.0\linewidth, height=21.2cm]{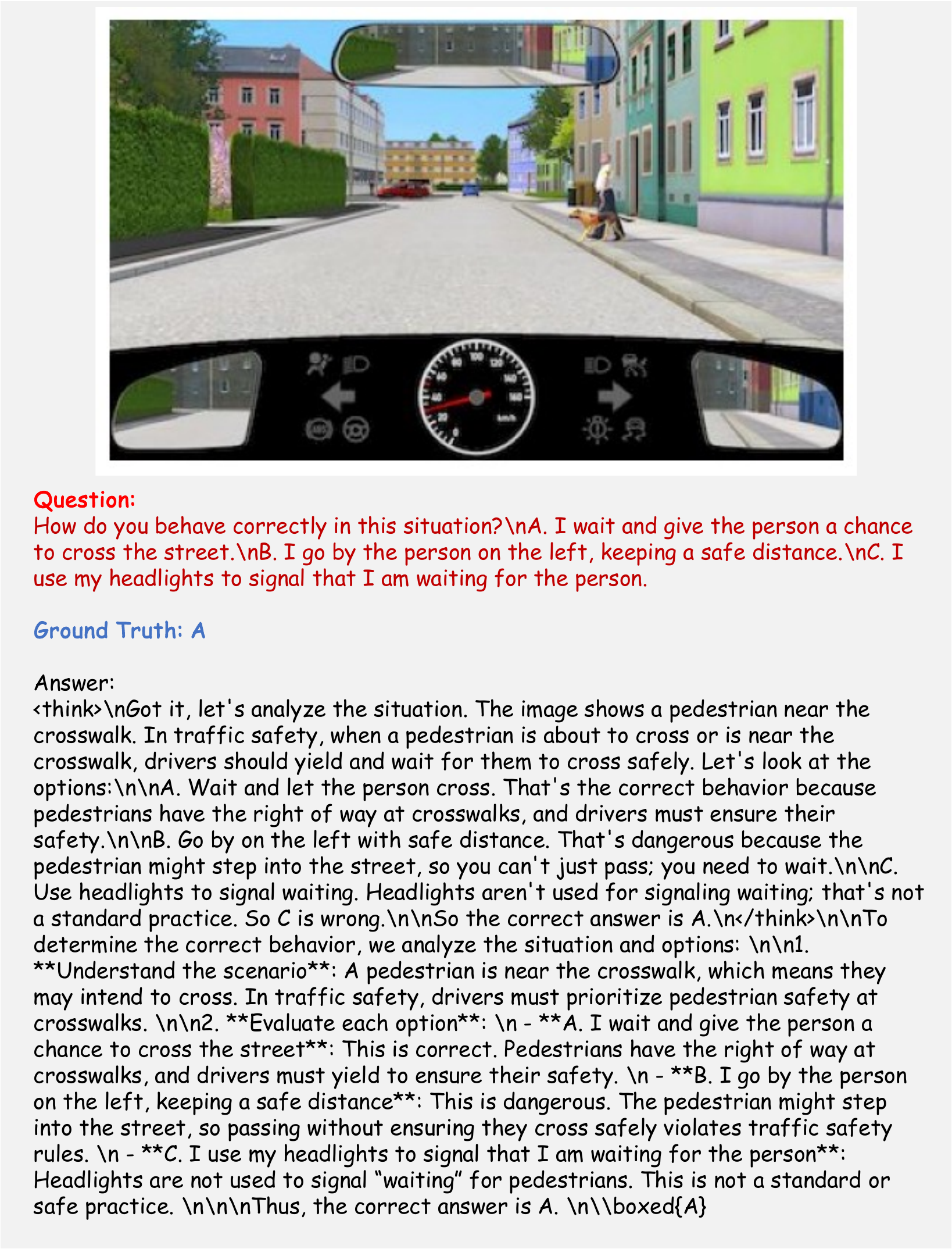}
    \caption{Autonomous driving planning task example 6.}
    \label{fig:av-planning-6}
\end{figure}
\let\clearpage\relax

\end{document}